\documentclass{article}

\usepackage{microtype}
\usepackage{graphicx}
\usepackage{booktabs} %

\usepackage{hyperref}

\usepackage[accepted]{icml2024}

\usepackage{amsmath}
\usepackage{amssymb}
\usepackage{mathtools}
\usepackage{amsthm}

\usepackage[capitalize,noabbrev]{cleveref}

\theoremstyle{plain}

\theoremstyle{definition}

\theoremstyle{remark}

\usepackage{mathtools}
\usepackage{amsmath}
\usepackage{amsthm}

\usepackage{graphicx}
\usepackage{subcaption}  %
\usepackage{cleveref}
\usepackage{wrapfig}
\usepackage{placeins}
\usepackage{float}
\usepackage{multicol}
\usepackage{multirow}
\usepackage{colortbl,xcolor}

\newtheorem*{hypothesis*}{Main Hypothesis}

\def\MH/{\textbf{MH}}

\icmltitlerunning{Meaning in Language Models Trained on Programs}

\begin{document}

\twocolumn[
\icmltitle{Emergent Representations of Program Semantics \\in Language Models Trained on Programs}

\begin{icmlauthorlist}
\icmlauthor{Charles Jin}{csail}
\icmlauthor{Martin Rinard}{csail}
\end{icmlauthorlist}

\icmlaffiliation{csail}{CSAIL, MIT, Cambridge, MA, USA}

\icmlcorrespondingauthor{Charles Jin}{ccj@csail.mit.edu}

\icmlkeywords{Machine Learning, ICML}

\vskip 0.3in
]

\printAffiliationsAndNotice{}  %

\begin{abstract}
We present evidence that language models (LMs) of code can learn to represent the formal semantics of programs, despite being trained only to perform next-token prediction.
Specifically, we train a Transformer model on a synthetic corpus of programs written in a domain-specific language for navigating 2D grid world environments.
Each program in the corpus is preceded by a (partial) specification in the form of several input-output grid world states.
Despite providing no further inductive biases, we find that a probing classifier is able to extract increasingly accurate representations of the \emph{unobserved, intermediate} grid world states from the LM hidden states over the course of training, suggesting the LM acquires an emergent ability to \emph{interpret} programs in the formal sense. We also develop a novel interventional baseline that enables us to disambiguate what is represented by the LM as opposed to learned by the probe. We anticipate that this technique may be generally applicable to a broad range of \emph{semantic} probing experiments. In summary, this paper does not propose any new techniques for training LMs of code, but develops an experimental framework for and provides insights into the acquisition and representation of formal semantics in statistical models of code.
\end{abstract}

\section{Introduction}

As LMs continue to improve on a range of downstream tasks, their capabilities in the domain of \emph{programming languages} have drawn increasing attention \citep{bommasani2023holistic,zan2023large}.
The advancement of recent models such as GPT-4 \citep{openai2023gpt4}, Code Llama \citep{rozière2023code}, Gemini \citep{geminiteam2023gemini}, and Claude 3 \citep{anthropic2024claude} has also spurred the widespread adoption of LMs in developer workflows via mainstream commercial products, offering functionality such as code completion, debugging assistance, generating documentation and commit messages, and writing test cases \citep{fan2023large}.

Despite these results, however, a major open question is whether current LMs capture any information about the \emph{semantics} of the text that they consume and generate~\citep{Mitchell_2023}. Indeed, one hypothesis---which takes a unified view of both natural and programming language domains---is that LMs trained purely on form (e.g., to model the conditional distribution of tokens in a training corpus) produce text only according to surface statistical correlations gleaned from the training data~\citep{bender2020climbing}, with any apparently sophisticated behavior attributable to the scale of the model and training data.

This work studies whether LMs of code learn aspects of semantics when trained using standard textual pretraining. We empirically evaluate the following hypothesis (\MH/):

\vspace{.25em}
\begin{hypothesis*}
\label{main_hypothesis}
LMs of code trained only to perform next token prediction on text do not model the formal semantics of the underlying programming language.
\end{hypothesis*}

To investigate \MH/, we apply language modeling to the task of \emph{program synthesis}, or generating a program that implements a given (partial) specification, which we take to be a set of input-output examples.
Specifically, we explore whether an LM trained on text that encodes only the \emph{input-output behavior of programs} also learns to model the \emph{intermediate program states} specified by the \emph{small-step operational semantics} of the synthesized program \citep{plotkin1981structural}.
We train an LM on example programs preceded by several input-output examples, then use small classifiers to \emph{probe} the LM's hidden states for (abstractions of) the intermediate states in the program execution.
Despite the text of the training corpus encoding only input-output behavior, we find the probe's ability to extract intermediate states undergoes a phase transition during training, with the phase transition strongly correlated with the LM's ability to generate a correct program in response to previously unseen specifications.
We also present results from a novel interventional experiment, which indicate that the semantics are represented by the LM (rather than learned by the probe).

\begin{figure*}[tb]
\centering
  \centering
  \includegraphics[scale=1.]{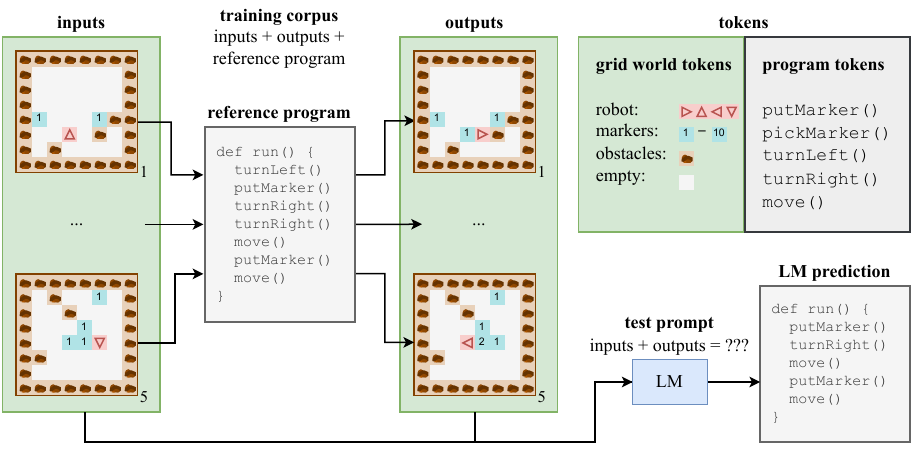}
  \caption{
  An overview of the experimental setting. We construct training examples by sampling a random reference program, then sampling 5 random inputs and executing the program to obtain the corresponding 5 outputs. 
  The LM is trained for next-token prediction on a corpus of examples consisting of the interleaved inputs and outputs, then the reference program. 
  At test time, we provide an unseen input-output specification to the LM, and use greedy decoding to predict a program.
  }
  \label{fig:task_overview}
\end{figure*}

We summarize our main contributions as follows:

\vspace{-.5em}
\paragraph{Emergence of meaning} We present results that are consistent with the emergence of representations of formal semantics in LMs trained to perform next token prediction (\Cref{sec:meaning}). In particular, we use the trained LM to generate programs given input-output examples, then train small probing classifiers to extract information about the intermediate program states from the hidden states of the LM. We find that the LM states encode (1) an abstract semantics---specifically, an \emph{abstract interpretation}---that tracks the intermediate states of the program through its execution and (2) predictions of \emph{future} program states corresponding to program tokens that have yet to be generated. During training, these representations of semantics emerge in lockstep with the LM's ability to generate correct programs.

\vspace{-.5em}
\paragraph{Semantic probing interventions} We present a novel interventional technique for disentangling the contributions of the LM and the probe when probing for semantics (\Cref{sec:semantic_intervention}). Specifically, one possible explanation for the results in \Cref{sec:meaning} is that the LM states contain a (purely syntactic) record of the inputs and generated program, from which the probe learns to generate the abstract interpretation.
Our key insight is that, if this were true, we should be able to supervise a new probe to interpret the (hypothetical) syntactic record according to an appropriately chosen set of \emph{alternative} semantics and achieve accuracies similar to the original semantics. However, we find that probes trained on alternative semantics achieve lower accuracies, which is consistent with the proposition that LM states are aligned with the original semantics and inconsistent with the proposition that the LM states simply encode a syntactic record.

Taken together, \Cref{sec:meaning,sec:semantic_intervention} present evidence that rejects \MH/: we find that, contrary to \MH/, representations of formal semantics emerge via next token prediction in our setting.
We therefore conclude that training LMs of code solely to predict the next token does not imply that they cannot develop accurate models of the underlying domain's semantics. 
More broadly, we see programs and their precise formal semantics as a promising direction for working toward a deeper understanding of the behavior of LMs, such as whether or how LMs acquire and use semantic representations of the underlying domain more generally.

\section{Background and setting}
\label{sec:background}

This section provides brief background of the \emph{program trace} as our chosen model of formal program semantics, introduces the language modeling task and setting, and presents qualitative results from our training run. %

\subsection{Program tracing as meaning}

A foundational topic in the theory of programming languages, formal semantics \citep{winskel1993formal} is the study of how to formally specify the meaning of programs. In this work, we use the {\em small step semantics} \citep{plotkin1981structural} to generate \emph{program traces} \citep{cousot2002constructive}: given an input (i.e, an assignment of values to input variables), the trace is the sequence of intermediate program states traversed by the program as it executes over the input. 
A (syntactic) program can then be formally assigned a (semantic) meaning, given by the collection of all of its traces.

Beyond its amenability to formal analysis, tracing is attractive as a model of program semantics for several reasons. 
In novice programmers, the ability to accurately trace a piece a code has been directly linked to the ability to explain the code \citep{lopez2008relationships, lister2009further}, and computer science education has emphasized tracing as a method of developing program understanding \citep{hertz2013trace} and localizing reasoning errors \citep{10.1145/2483710.2483713}. Expert programmers also rely on tracing, both as a mental process \citep{letovsky1987cognitive} and via trace-based debuggers. %

\vspace{-.5em}
\paragraph{Abstract interpretation}

\emph{Abstract interpretation} \citep{cousot1977abstract} is a technique for producing sound approximations of concrete program semantics. For example, given the multiplication operator $\times$ over the integers $\mathbb{Z}$, we could define an abstract interpretation $\alpha$ by abstracting each integer to its sign $\alpha : \mathbb{Z} \mapsto \{-, 0, +\}$. 
In this paper, we use abstract interpretation to establish a precise, formal connection between the concrete program states and the abstract program states measured in our experiments.

\subsection{Language modeling task and training}
\label{sec:experimental_procedure}

\paragraph{Karel domain}

Karel is an educational programming language \citep{pattis1994karel} developed at Stanford in the 1970s, which is still in use in their introductory programming course today \citep{karelreader,stanford_intro_cs}. The domain features a robot (named Karel) navigating a 2D grid world with obstacles while leaving and picking up markers. Since being introduced by \citet{devlin2017neural}, Karel has been adopted by the program synthesis community as a standard benchmark  \citep{bunel2018leveraging,shin2018improving,sun2018neural,chen2019execution,chen2021latent}, in which input-output examples are provided, and the task is to produce a program which maps each input grid to its corresponding output grid.

\Cref{fig:task_overview} gives an overview of our domain. Each 8x8 grid world contains 4 types of tokens: the robot controlled by the program, which we represent graphically with an arrow in the direction that the robot currently faces (red); markers (blue); obstacles (brown); or an empty space (gray). We use a subset of the language consisting of the following 5 operations: \texttt{move} advances the robot by one space in the facing direction if there is not an obstacle ahead (otherwise, the robot does not move); \texttt{turnRight} and \texttt{turnLeft} turn the robot right and left, respectively; \texttt{putMarker} and \texttt{pickMarker} increment and decrement the number of markers on the space occupied by the robot (with no effect if there are 10 and 0 markers), respectively. The robot also obscures the number of markers on the space it currently occupies; the obscured markers have no effect on the correctness of the program.
Note that there is no control flow and all programs consist of straight line programs, so that each operation produces exactly one program state when the program is traced.

\vspace{-.5em}
\paragraph{Synthetic dataset construction}

Our training set consists of 500,000 randomly sampled Karel programs of lengths between 6 and 10, inclusive. For each program, we randomly sample 5 grid worlds as input, then evaluate the program to obtain 5 output grids. We create textual representations for Karel grid worlds by scanning the grid in row order, with one token per grid space. Each training sample consists of the concatenation of the 5 input-output grid world states (the \emph{specification}), followed by the \emph{reference program}.
The language modeling task thus consists of predicting a program from a (partial) specification in the form of input-output grid world states. Note that (1) the training set consists only of programs which correctly implement the preceding specification and (2) the intermediate states of the trace are not observed in the training data. We also generate a test set of 10,000 specifications in the same manner, except that we sample reference programs of length between 1 and 10.

\vspace{-.5em}
\paragraph{Training an LM to synthesize programs}
We train an off-the-shelf Transformer \citep{vaswani2017attention} to perform next token prediction on our dataset. Specifically, we train a 350M parameter variant of the CodeGen architecture \citep{nijkamp2023codegen} in the HuggingFace Transformers library \citep{huggingface} from initialization for approximately 2.5 billion tokens. \Cref{appendix:LM_details} contains further details.

\begin{figure}[tb]
  \begin{center}
    \includegraphics[height=5.5cm]{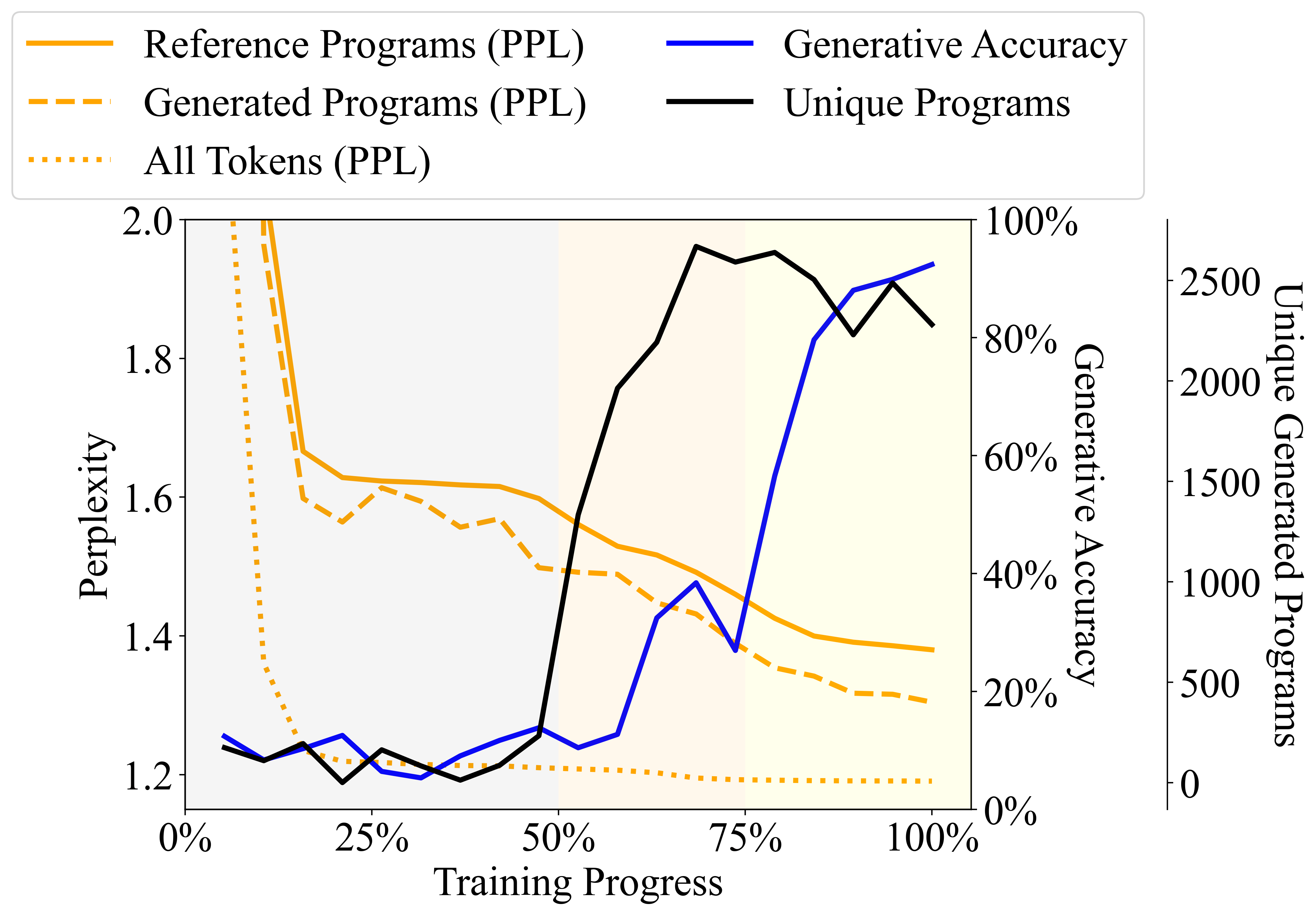}
  \end{center}
  \caption{Three distinct phases during training: babbling (gray), syntax acquisition (orange), and semantics acquisition (yellow), based on qualitative differences in the evolution of perplexity (orange), generative accuracy (blue), and diversity of output (black). The number of unique programs is measured over the test set, which contains 10,000 specifications and 6,473 unique reference programs.}
  \label{fig:perplexity_main}
\end{figure}

\subsection{Results}

\begin{figure*}[bt]
  \centering
  \centering
  \includegraphics[scale=1.,trim={0 0 1.15cm 0},clip]{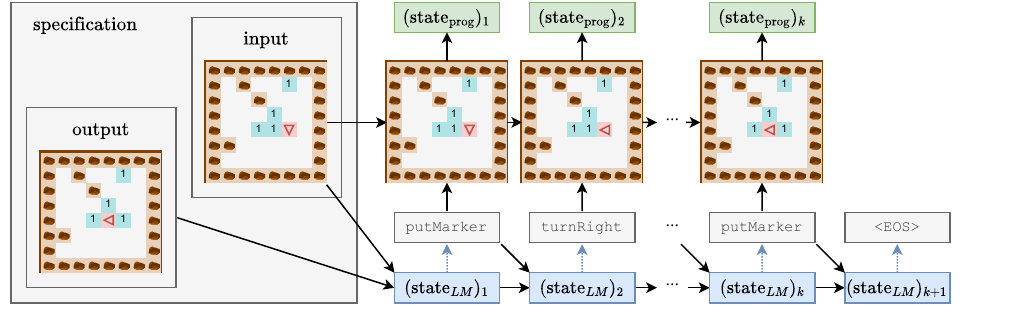}
  \caption{An overview of the trace dataset construction for the probe task. Given a specification consisting of $\text{input}$ and $\text{output}$ for some (unobserved) reference program, we use the trained LM to generate a program using next-token prediction (dotted blue arrows), yielding a sequence of $(\text{state}_{LM})_i$. At the same time, each token is an operation that induces a transition in the program state to $(\text{state}_\text{prog})_i$.
  The probe is trained to predict $(\text{state}_\text{prog})_i$ given $(\text{state}_{LM})_i$.
  Note that, while the depicted generation is correct as the final $(\text{state}_\text{prog})_k$ is equal to the specified output state, this need not be the case in general (i.e., the LM may generate incorrect programs).
  For clarity, we depict the specification as a single input-output example (rather than 5); autoregressive edges are also hidden.
  }
  \label{fig:traces}
\end{figure*}

\Cref{fig:perplexity_main} plots the main results from our training run.
To measure the ability of the LM to synthesize programs, we use the LM to generate text starting from a specification using greedy decoding constrained to program tokens, i.e., the generated text is guaranteed to be a syntactically well-formed program. The program is correct if it maps each input in the specification to its corresponding output; we define the \textbf{generative accuracy} as the percentage of correct programs over the test set. The LM reaches 92.4\% generative accuracy at the end of training. %

We also track two additional metrics related to the \emph{syntax} (or form) of the LM outputs: the number of unique programs generated by the LM over the test set (black) and the perplexity of the LM over different subsets of tokens in the test set (orange). In particular, the solid orange line plots the perplexity on the reference programs, the dashed orange line plots the perplexity on the programs generated by the LM, and the dotted orange line plots the perplexity over all tokens. Note that overall perplexity improves over the entire run, indicating that the training dynamics remain stable.

We observe 3 distinct phases during training:
in the \textbf{babbling} phase (up to 50\% of training, gray background), the generated programs are often highly repetitive, with a plateau in perplexity on the reference programs starting around 20\%. The generative accuracy stays flat at around 10\%.
The next phase (50-75\% of training, orange background) exhibits a sharp increase in the diversity of the generated outputs with a corresponding decrease in perplexity on the reference programs---i.e., the LM begins to model the program tokens---with a modest increase in generative accuracy (from 10\% to 25\%).
In the final phase (from 75\% to the end of training, yellow background), the diversity of the generated programs stays roughly constant, and the perplexity on the reference programs continues to improve at the previous rate. Conversely, the generative accuracy of the LM increases rapidly, from 25\% to over 90\%. As such, the middle phase sees the most significant change in the syntactic properties of the LM's generation, while the final phase is characterized by a rapid improvement in the LM's ability to generate semantically correct output. We hence identify these two phases primarily with \textbf{syntax acquisition} and \textbf{semantics acquisition}, respectively.

Finally, while it is natural for the perplexity of the generated programs to be lower than the reference programs (due to the use of greedy decoding), the constant margin between the two perplexities suggests that the LM consistently \emph{underfits} the program tokens in the training data, despite producing increasingly correct programs; we refer to \cref{appendix:additional_results} for further analyses.
These dynamics suggest that the increase in generative accuracy over the course of LM training cannot be attributed entirely to the LM's ability to model the surface distribution of program tokens in the training corpus.

\section{Emerging representations of semantics}
\label{sec:meaning}

We train small probing classifiers to extract information about the program state from the hidden states of the LM.
The idea is to prompt the LM to generate a program given some inputs, and check whether the LM states contain a representation of the \emph{intermediate program states as it generates the program}.
A positive result is consistent with the LM having learned to model the underlying semantics of the programs it generates, constituting evidence against \MH/.

\subsection{Probing for representations of the program trace}

\paragraph{Trace dataset construction}
Every 4000 steps (or roughly 5\%) of training, we take a snapshot of (1) the hidden states of the LM as it generates programs using next-token prediction and (2) the corresponding program states after evaluating the partial program on each of the 5 specified inputs. Specifically, starting from an input-output specification, we generate according to the standard autoregressive loop:
\begin{align}
\label{eq:trace_dataset}
(\text{state}_{LM})_i = {LM}(\text{input}, \text{output}, \{(\text{state}_{LM})_j\}_{j=1}^{i-1}) \\
\label{eq:greedy_decode}
\text{token}_i = \texttt{greedy}({LM}_{\text{head}}((\text{state}_{LM})_i)) \\
(\text{state}_{\text{prog}})_i = \texttt{exec}(\text{token}_i, (\text{state}_{\text{prog}})_{i-1})
\end{align}
where each $(\text{state}_{LM})_i$ is a hidden state,  $(\text{state}_{\text{prog}})_0$ is the inputs from the specification, \texttt{greedy} performs greedy decoding (constrained to program tokens), and \texttt{exec} executes a program token on the program state. We generate up to 14 tokens, or until the model outputs a special end-of-sequence token ($\texttt{<EOS>}$). \Cref{fig:traces} illustrates this process.

We average the hidden state over the layer dimension, so that the snapshot is a 1-dimensional tensor of size 1024, and call this the \textbf{model state}. We repeat this process for each of the training and test sets, producing two \emph{trace datasets} consisting of aligned pairs of $(\text{state}_{LM})_i$, $(\text{state}_{\text{prog}})_i$ from all the programs generated from the specifications in the training and test sets, respectively

\vspace{-.5em}
\paragraph{Probe training}
For each training trace dataset, we train a set of \textbf{probes} (ranging from linear to 2-layer MLPs) to predict features of the program state given the model state, using standard supervised learning. The features consist of (1) the facing direction of the robot, (2) the position of the robot as an offset from its starting position, and (3) whether the robot is current facing an obstacle, i.e.:
\begin{align}
\alpha: \text{state}_{\text{prog}} \mapsto (\text{position}, \text{direction}, \text{obstacle})
\end{align}
As the features are an abstraction of the full program state, we refer to them collectively as the \textbf{abstract state}.\footnote{To avoid introducing more notation, we use $\text{state}_{LM}$ and $\text{state}_{\text{prog}}$ to refer to both (1) the full hidden states of the LM and program state during trace dataset generation, and (2) the averaged model state and abstract program state during probing, respectively, whenever the distinction is clear from context.}
We then evaluate the accuracy of the probe over the corresponding test trace dataset, and define the \textbf{semantic content} as the geometric mean (over the 3 features). As tracing the abstract state is formally equivalent to performing an abstract interpretation of the program, the semantic content measures, in a precise sense, the extent to which the model states encode an abstract interpretation of the formal semantics.

\subsection{Results}

This section presents a summary of the main results; \Cref{appendix:additional_results} contains additional results, including results for individual features and all 3 probes. We also evaluate several additional hypothesis, including that the semantic content is due to a retrieval process (i.e., the LM is simply recalling the abstract states from previously seen data), which can be viewed as a variation on \MH/. The main idea is that the training corpus contains only programs of length 6 or greater, so the LM cannot use retrieval for shorter programs and must learn to infer (rather than retrieve) the semantics.

\begin{figure}[tb]
  \begin{center}
    \includegraphics[height=5.5cm]{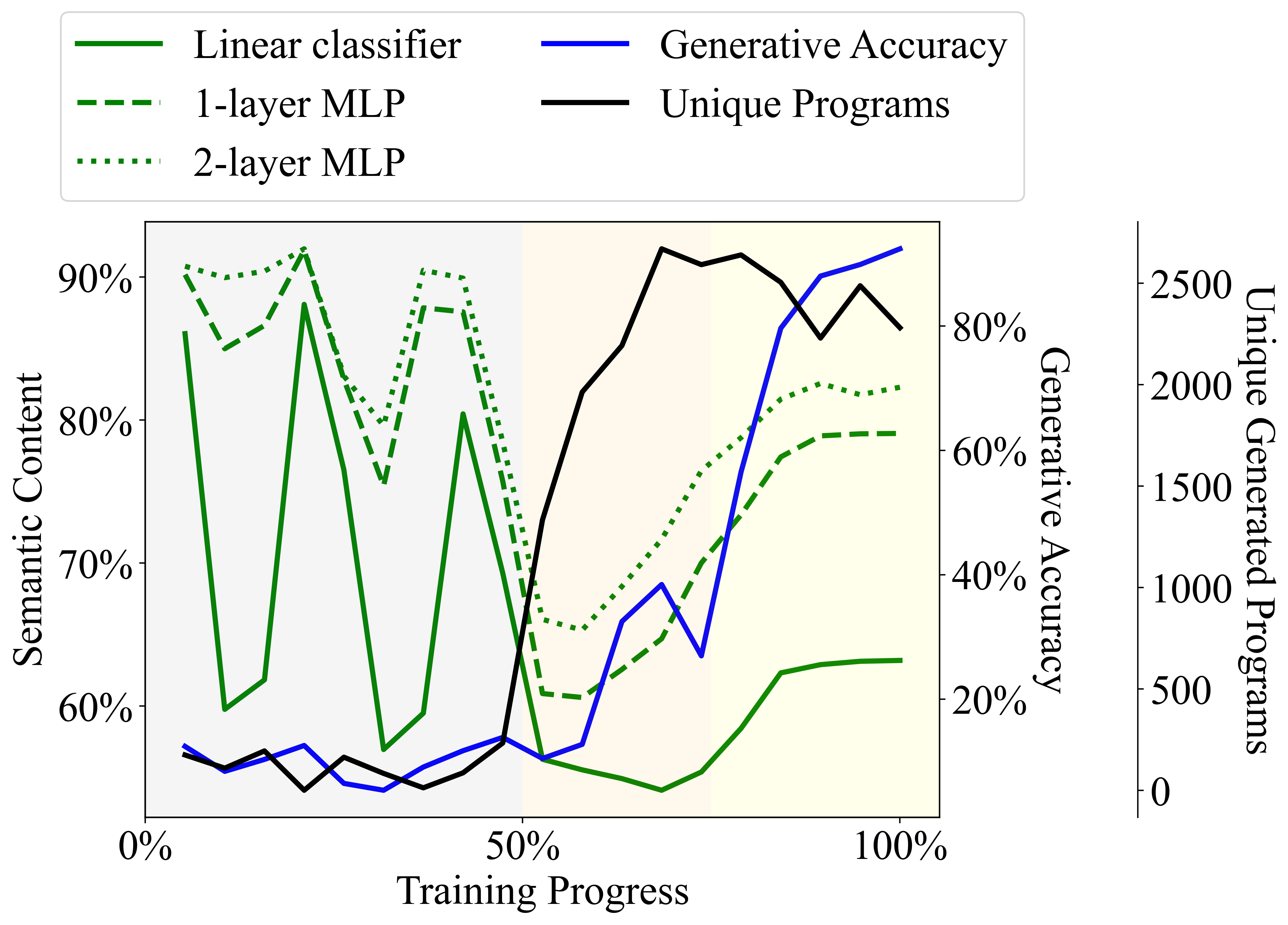}
  \end{center}
  \caption{The semantic content (green) measured by different probing classifiers.}
  \label{fig:gen_vs_sem}
\end{figure}

\vspace{-.5em}
\paragraph{Emergence of semantics is correlated with generative accuracy} 

\Cref{fig:gen_vs_sem} plots the results of our probing experiments. Our first observation is that the semantic content during the babbling phase is extremely noisy, which can be attributed to a lack of diversity in the outputs of the LM, so that the probe only needs to fit a trivial set of semantics. For instance, about 20\% of the way through training, the LM degenerates to generating a single program of 9 \textsc{pickMarker} tokens, regardless of the specification.
Conversely, all 3 probes reach a minimum during the syntax acquisition phase (where the LM outputs grow more diverse), and steadily increase over the semantics acquisition phase of training. This result is consistent with the proposition that the hidden states of the LM do in fact contain (relatively) shallow encodings of the abstract state, and crucially these representations emerge within an LM trained purely to perform next token prediction on text.
Regressing generative accuracy against semantic content during the second half of training also yields strong, statistically significant linear correlations ($R^2=0.904, 0.886, 0.821$ with $p<0.001$ for the linear, 1-layer MLP, and 2-layer MLP probes, respectively).

\vspace{-.5em}
\paragraph{Representations are predictive of future program states}

\begin{figure}[tb]
  \begin{center}
    \includegraphics[height=5.5cm]{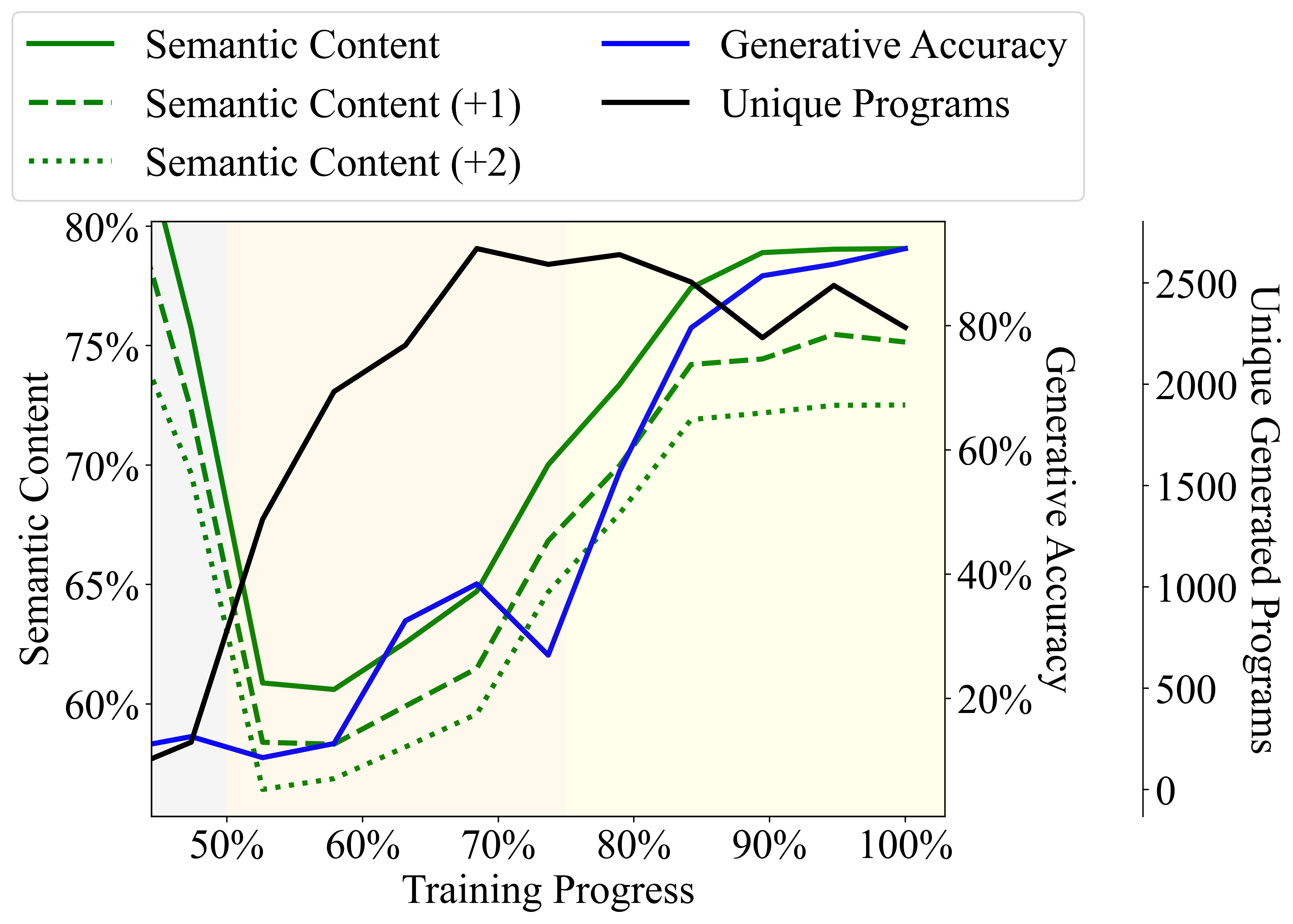}
  \end{center}
  \caption{The semantic content (1-layer MLP) for the current and next two abstract states over the second half of training.}
  \label{fig:intent_over_time}
\end{figure}

We next explore whether the trained LM encodes the semantics of text that has yet to be generated.
Specifically, we train probes to predict \emph{future} abstract states from model states. %
\Cref{fig:intent_over_time} displays how well a 1-layer MLP is able to predict abstract states 1 and 2 steps into the future. As with the previous results, the probe's performance reaches a minimum during syntax acquisition, then increases for the remainder of training. We also find a strong correlation between the semantic content of future states and the generative accuracy in the second half of training;
regressing the semantic content against the generative accuracy yields an $R^2$ of $0.878$ and $0.874$ ($p<.001$) for 1 and 2 abstract states into the future, respectively.

\begin{table}
\centering
\begin{tabular}{ l | c | c | c }
\toprule
 & 0 & +1 & +2 \\
 \midrule
 linear classifier & 63.2 & 61.1 & 60.1 \\
 1-layer MLP & 79.1 & 75.1 & 72.5 \\
 2-layer MLP & 82.3 & 79.2 & 77.3 \\
 \midrule
 baseline (current state) & 100.0 & 83.9 & 75.6 \\
 \bottomrule
\end{tabular}
\caption{Probing accuracy for current and future abstract states at the end of training vs. a baseline of predicting the current abstract state for future abstract states.}
\label{table:future}
\end{table}

\Cref{table:future} compares the probing results at the end of training against a baseline which simply predicts the current abstract state for all future abstract states (which is the Bayes-optimal predictor absent any information about future states). We observe that (1) the accuracy of using the baseline degrades more rapidly than the probe, which suggests that the probes are not simply using encodings of the current state to predict future states, and (2) the absolute accuracy at 2 states into the future is greater using the 2-layer MLP probe than the baseline. These results suggest that the LM encodes information about what it \emph{intends} to say ahead of its generation.

\section{Semantic probing interventions}
\label{sec:semantic_intervention}

We next evaluate the possibility that semantics are learned by the probe instead of the LM.
For example, because the probe is explicitly supervised on intermediate states, the model states may encode the inputs and a record of the generated program tokens, with the probe learning to interpret the tokens one-by-one.
More generally, the semantic content could be attributed to (1) the LM encoding only lexical and syntactic structure while (2) the probe learns to derive the semantics from the lexical and syntactic structure encoded in the LM state (because it is explicitly supervised to predict the semantics from the LM state). We refer to this as the \textbf{syntactic record} hypothesis, which offers an explanation for the results in \Cref{sec:meaning} consistent with \MH/.

To test this hypothesis, we design a novel interventional experiment that preserves the lexical and syntactic structure of the language and intervenes only on the semantics.
Then, we re-execute the program with the alternative semantics to obtain a \emph{new} trace with \emph{new} abstract states, and train a \emph{new} probe to predict the new abstract states using the \emph{original} model states. Our key insight is that, if the model states encode only syntactic information, then the new probe should be capable of extracting the new semantics from the original syntactic record equally well, leaving the semantic content unchanged.

\subsection{Intervening on semantics}

\begin{figure*}[tbp]
\centering
  \centering
  \includegraphics[width=\linewidth]{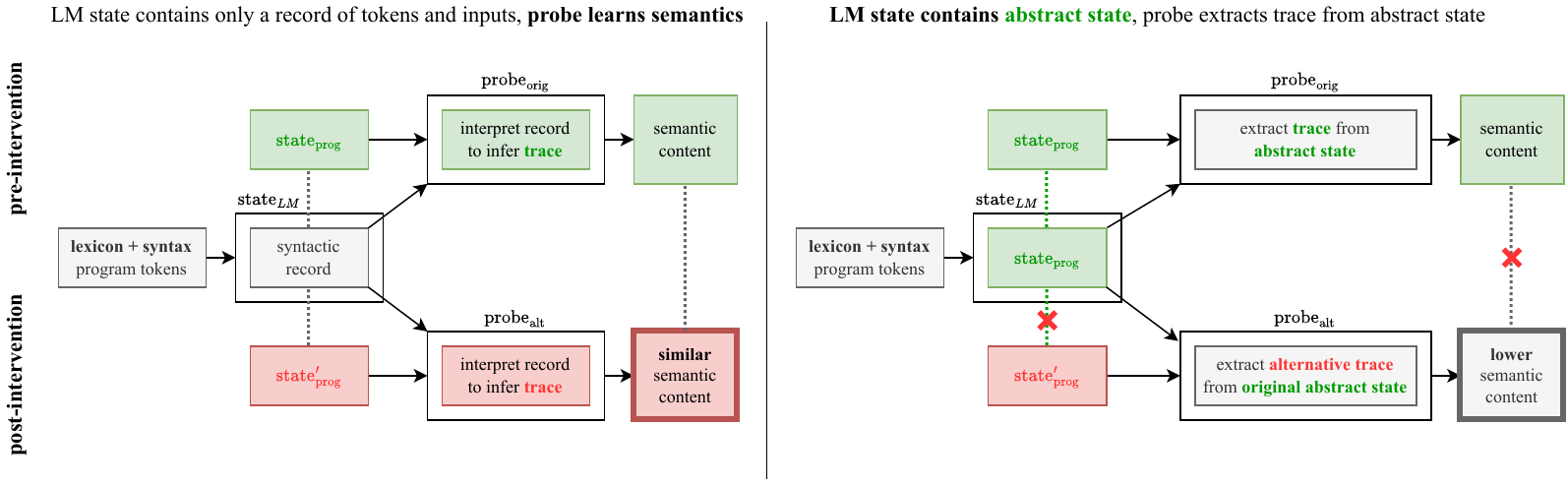}
  \caption{
  The proposed semantic intervention baseline. We use green for the original semantics, red for the alternative semantics, and gray for non-semantic components (such as syntax).
  We aim to distinguish between two hypotheses: (left) the LM only maintains a syntactic record (e.g., a list of the inputs and program tokens generated thus far), while $\text{probe}_\text{orig}$ learns to infer semantics from the record; and (right) the LM learns to represent $\text{state}_\text{prog}$, while $\text{probe}_\text{orig}$ just extracts the semantics. We mark the emergent connection between the original semantics and the LM representations in the latter case by a dashed green line. The top half depicts how, pre-intervention, both cases can lead to the high semantic content measured in \Cref{sec:meaning}.
  The bottom half displays why intervening on the semantics while preserving the form of programs separates the two hypotheses: if the LM representations contain only syntax (bottom left), then it should be possible to train $\text{probe}_\text{alt}$ to learn to interpret the record according to the alternative $\text{state}'_\text{prog}$ (bold red outcome); however, if the LM representations encode the original abstract state (bottom right), then $\text{probe}_\text{alt}$ needs to extract the alternative $\text{state}'_\text{prog}$ from the original $\text{state}_\text{prog}$, yielding a lower semantic content (bold gray outcome).
  }
  \label{fig:interventional_design}
\end{figure*}

\begin{table*}[tbp]
\caption{The results of our semantic intervention experiments. For each of the original, flip, and adversarial semantics, we report the semantic content (SC) at the end of training for 2 abstract states into the past (-2, -1), the current state (0), and 2 abstract states into the future (+1, +2), using linear, 1-layer MLP, and 2-layer MLP probes. We also regress the SC against the generative accuracy over the second half of training ($R^2 (p)$). For each of the alternative semantics, we additionally compute the difference with respect to the original semantics ($\Delta$) and regress the difference against the generative accuracy over the second half of training as ($R^2 (p)$ of $\Delta$). Highlighted cells are statistically significant at a level of $p<0.05$ with an $R^2$ of at least 50\%; all such correlations are positive. We reject the syntactic record hypothesis due to the magnitude ($\Delta$ and $R^2$ of $\Delta$) and statistical significance ($p$ of $\Delta$) of the drop in semantic content when probing for the alternative semantics.}
\label{table:main}
\begin{center}
\tabcolsep=0.15cm
\begin{tabular}{ c c | c c | c c | c c | c c | c c }
\toprule
 & & \multicolumn{2}{c|}{original} & \multicolumn{4}{c|}{flip} &  \multicolumn{4}{c}{adversarial} \\
 & & SC & $R^2 (p)$ & SC & \multicolumn{1}{c}{$R^2 (p)$} & $\Delta$ & $R^2 (p)$ of $\Delta$ & SC & \multicolumn{1}{c}{$R^2 (p)$} & $\Delta$ & $R^2 (p)$ of $\Delta$ \\
 \midrule

\multirow{5}{*}{\rotatebox[origin=c]{90}{linear}} & -2 & \cellcolor{yellow!25} 64.2 & \cellcolor{yellow!25} 86.0 (\textless.001) & \cellcolor{yellow!25} 60.5 & \cellcolor{yellow!25} 55.5 (0.021) & \cellcolor{yellow!25} 3.7 & \cellcolor{yellow!25} 72.2 (0.004) & 55.8 & 7.4 (0.478) & \cellcolor{yellow!25} 8.4 & \cellcolor{yellow!25} 60.5 (0.014) \\
 & -1 & \cellcolor{yellow!25} 64.4 & \cellcolor{yellow!25} 87.1 (\textless.001) & \cellcolor{yellow!25} 59.9 & \cellcolor{yellow!25} 66.1 (0.008) & \cellcolor{yellow!25} 4.5 & \cellcolor{yellow!25} 81.4 (\textless.001) & 54.8 & 2.6 (0.680) & \cellcolor{yellow!25} 9.7 & \cellcolor{yellow!25} 70.7 (0.005) \\
 & 0 & \cellcolor{yellow!25} 63.2 & \cellcolor{yellow!25} 90.4 (\textless.001) & \cellcolor{yellow!25} 57.3 & \cellcolor{yellow!25} 52.7 (0.027) & \cellcolor{yellow!25} 5.9 & \cellcolor{yellow!25} 86.9 (\textless.001) & 53.1 & 0.4 (0.878) & \cellcolor{yellow!25} 10.1 & \cellcolor{yellow!25} 71.9 (0.004) \\
 & 1 & \cellcolor{yellow!25} 61.1 & \cellcolor{yellow!25} 91.3 (\textless.001) & 55.2 & 38.4 (0.075) & \cellcolor{yellow!25} 5.9 & \cellcolor{yellow!25} 89.3 (\textless.001) & 52.6 & 0.6 (0.846) & \cellcolor{yellow!25} 8.5 & \cellcolor{yellow!25} 59.4 (0.015) \\
 & 2 & \cellcolor{yellow!25} 60.1 & \cellcolor{yellow!25} 92.3 (\textless.001) & 54.3 & 23.2 (0.189) & \cellcolor{yellow!25} 5.7 & \cellcolor{yellow!25} 90.7 (\textless.001) & 52.6 & 0.7 (0.827) & 7.5 & 49.4 (0.035) \\
\midrule
\multirow{5}{*}{\rotatebox[origin=c]{90}{MLP-1}} & -2 & \cellcolor{yellow!25} 82.8 & \cellcolor{yellow!25} 83.8 (\textless.001) & \cellcolor{yellow!25} 81.2 & \cellcolor{yellow!25} 87.1 (\textless.001) & 1.6 & 23.2 (0.190) & \cellcolor{yellow!25} 73.8 & \cellcolor{yellow!25} 66.3 (0.008) & \cellcolor{yellow!25} 9.0 & \cellcolor{yellow!25} 66.1 (0.008) \\
 & -1 & \cellcolor{yellow!25} 83.6 & \cellcolor{yellow!25} 83.9 (\textless.001) & \cellcolor{yellow!25} 81.6 & \cellcolor{yellow!25} 86.9 (\textless.001) & 1.9 & 30.3 (0.125) & \cellcolor{yellow!25} 73.4 & \cellcolor{yellow!25} 65.4 (0.008) & \cellcolor{yellow!25} 10.1 & \cellcolor{yellow!25} 72.9 (0.003) \\
 & 0 & \cellcolor{yellow!25} 79.1 & \cellcolor{yellow!25} 88.6 (\textless.001) & \cellcolor{yellow!25} 76.7 & \cellcolor{yellow!25} 90.3 (\textless.001) & \cellcolor{yellow!25} 2.3 & \cellcolor{yellow!25} 61.4 (0.012) & 66.7 & 44.7 (0.049) & \cellcolor{yellow!25} 12.4 & \cellcolor{yellow!25} 76.5 (0.002) \\
 & 1 & \cellcolor{yellow!25} 75.1 & \cellcolor{yellow!25} 87.8 (\textless.001) & \cellcolor{yellow!25} 72.2 & \cellcolor{yellow!25} 87.2 (\textless.001) & \cellcolor{yellow!25} 2.9 & \cellcolor{yellow!25} 83.3 (\textless.001) & 61.8 & 29.0 (0.135) & \cellcolor{yellow!25} 13.4 & \cellcolor{yellow!25} 75.0 (0.003) \\
 & 2 & \cellcolor{yellow!25} 72.5 & \cellcolor{yellow!25} 87.4 (\textless.001) & \cellcolor{yellow!25} 69.1 & \cellcolor{yellow!25} 86.9 (\textless.001) & \cellcolor{yellow!25} 3.4 & \cellcolor{yellow!25} 84.7 (\textless.001) & 59.1 & 12.2 (0.356) & \cellcolor{yellow!25} 13.4 & \cellcolor{yellow!25} 75.1 (0.002) \\
\midrule
\multirow{5}{*}{\rotatebox[origin=c]{90}{MLP-2}} & -2 & \cellcolor{yellow!25} 85.4 & \cellcolor{yellow!25} 75.5 (0.002) & \cellcolor{yellow!25} 84.0 & \cellcolor{yellow!25} 76.7 (0.002) & 1.4 & 48.5 (0.037) & \cellcolor{yellow!25} 83.1 & \cellcolor{yellow!25} 74.5 (0.003) & 2.3 & 5.2 (0.556) \\
 & -1 & \cellcolor{yellow!25} 85.6 & \cellcolor{yellow!25} 78.0 (0.002) & \cellcolor{yellow!25} 83.9 & \cellcolor{yellow!25} 79.3 (0.001) & \cellcolor{yellow!25} 1.7 & \cellcolor{yellow!25} 64.3 (0.009) & \cellcolor{yellow!25} 82.5 & \cellcolor{yellow!25} 72.6 (0.004) & 3.2 & 20.7 (0.218) \\
 & 0 & \cellcolor{yellow!25} 82.3 & \cellcolor{yellow!25} 82.1 (\textless.001) & \cellcolor{yellow!25} 80.6 & \cellcolor{yellow!25} 84.3 (\textless.001) & \cellcolor{yellow!25} 1.7 & \cellcolor{yellow!25} 63.0 (0.011) & \cellcolor{yellow!25} 76.1 & \cellcolor{yellow!25} 68.1 (0.006) & 6.2 & 25.2 (0.169) \\
 & 1 & \cellcolor{yellow!25} 79.2 & \cellcolor{yellow!25} 82.5 (\textless.001) & \cellcolor{yellow!25} 77.6 & \cellcolor{yellow!25} 84.5 (\textless.001) & \cellcolor{yellow!25} 1.7 & \cellcolor{yellow!25} 62.0 (0.012) & \cellcolor{yellow!25} 69.8 & \cellcolor{yellow!25} 51.7 (0.029) & 9.5 & 48.0 (0.039) \\
 & 2 & \cellcolor{yellow!25} 77.3 & \cellcolor{yellow!25} 83.7 (\textless.001) & \cellcolor{yellow!25} 75.6 & \cellcolor{yellow!25} 86.0 (\textless.001) & \cellcolor{yellow!25} 1.6 & \cellcolor{yellow!25} 58.2 (0.017) & 65.6 & 48.0 (0.039) & \cellcolor{yellow!25} 11.6 & \cellcolor{yellow!25} 59.3 (0.015) \\

 \bottomrule
\end{tabular}
\end{center}
\end{table*}

Concretely, we define two \textbf{alternative semantics} by reassigning the semantics of individual operations as follows:
\begin{center}
\begin{tabular}{ c | c | c }
\toprule
 original & flip & adversarial \\
 \midrule
 \texttt{pickMarker} & \texttt{pickMarker} & \texttt{turnRight} \\
 \texttt{putMarker} & \texttt{putMarker} & \texttt{turnLeft} \\
 \texttt{turnRight} & \texttt{turnLeft} & \texttt{move} \\
 \texttt{turnLeft} & \texttt{turnRight} & \texttt{turnRight} \\
 \texttt{move} & \texttt{move} & \texttt{turnLeft} \\ 
 \bottomrule
\end{tabular}
\end{center}
For instance, $\texttt{exec}(\texttt{turnRight}, \cdot)$ in the original semantics would have the robot turn 90 degrees clockwise, while $\texttt{exec}_\text{adversarial}(\texttt{turnRight}, \cdot)$ advances the robot by a space (i.e., according to the original semantics of \texttt{move}).

Next, for each sequence of program tokens in \Cref{eq:trace_dataset} from the construction of the original trace dataset, we use the \emph{same} tokens to define a corresponding alternative trace:
\begin{align}
\label{eq:alt_trace_dataset}
(\text{state}'_{\text{prog}})_i = \texttt{exec}_\text{alt}(\text{token}_i, (\text{state}'_{\text{prog}})_{i-1}),
\end{align}
where we also start from the \emph{same} initial program state: $(\text{state}_{\text{prog}})'_0 = (\text{state}_{\text{prog}})_0$ (i.e., the specification inputs).

Finally, with the \emph{new} traces from \Cref{eq:alt_trace_dataset} and the \emph{original} $\text{state}_{LM}$ from \Cref{eq:trace_dataset}, we train a new probe
\begin{align}
\text{probe}_\text{alt}: \text{state}_{LM} \mapsto \text{state}'_{\text{prog}},
\end{align}
and compare its accuracy against the original probe:
\begin{align}
\text{probe}_\text{orig}: \text{state}_{LM} \mapsto \text{state}_{\text{prog}}.
\end{align}
\Cref{fig:interventional_design} illustrates our setup. 
 
For an intervention to be a proper control, we identify two critical properties: (1) the alternative semantics should be limited to \emph{reassigning} the semantics of individual operations in the language (as opposed to inventing completely new semantics, e.g., ``jump three spaces diagonally'') and (2) the intervention must preserve the syntactic structure of programs (i.e., how the individual operations are composed when interpreting a full program). Then assuming the syntactic record hypothesis is true, $\text{probe}_\text{alt}$ should be able to interpret the record according to an analogous procedure as $\text{probe}_\text{orig}$, yielding comparable measurements of semantic content. Hence, \emph{rejecting the syntactic record hypothesis reduces to measuring a statistically significant degradation in the alternative semantic content} (relative to the original semantic content).

\subsection{Results}

\begin{figure}[tb]
  \begin{center}
    \includegraphics[height=5.5cm]{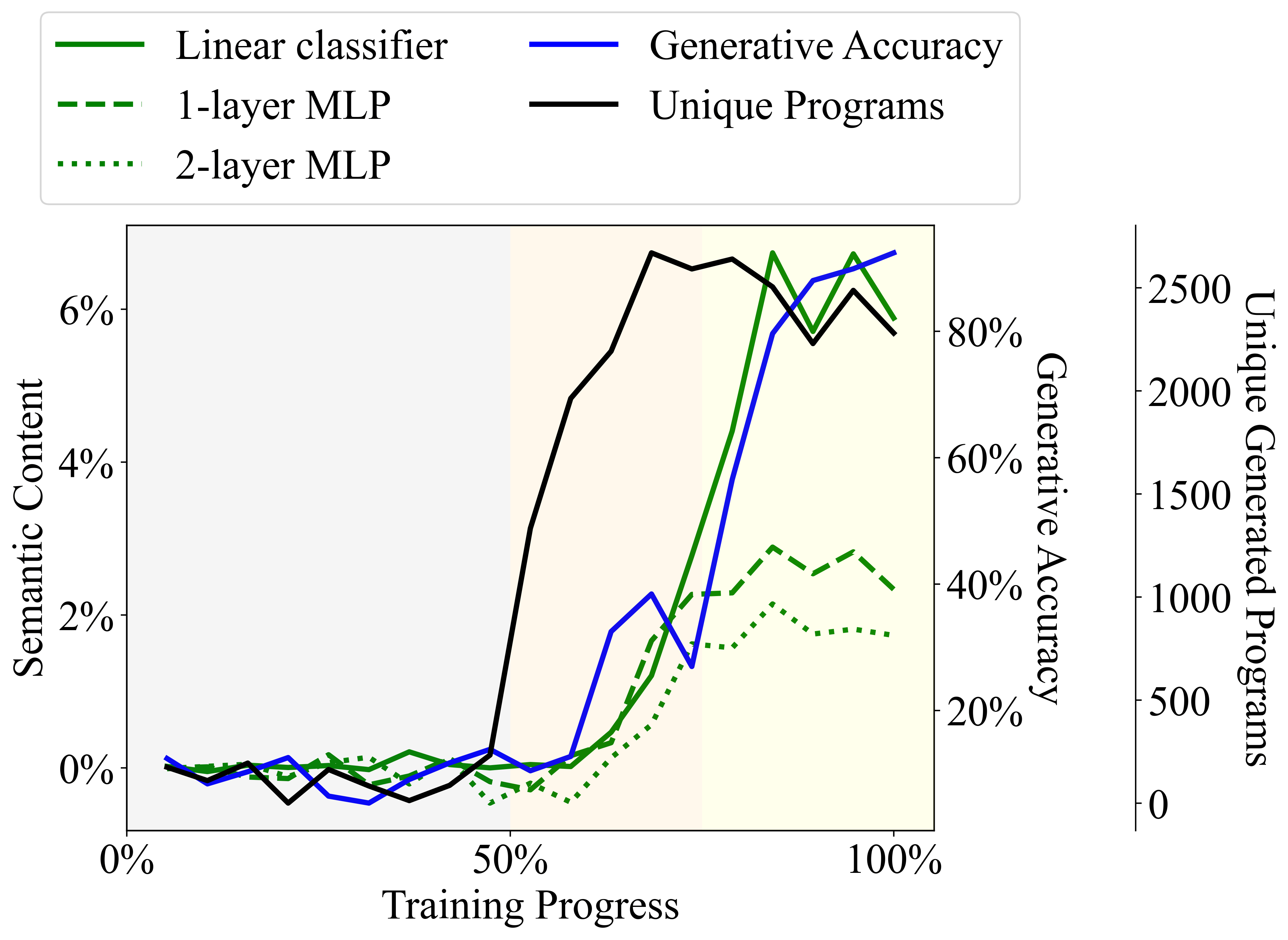}
  \end{center}
  \caption{Excess of original over flip semantic content using different probing classifiers.}
  \label{fig:alt_gen_vs_sem}
\end{figure}

\begin{figure}[tb]
  \begin{center}
    \includegraphics[height=5.5cm]{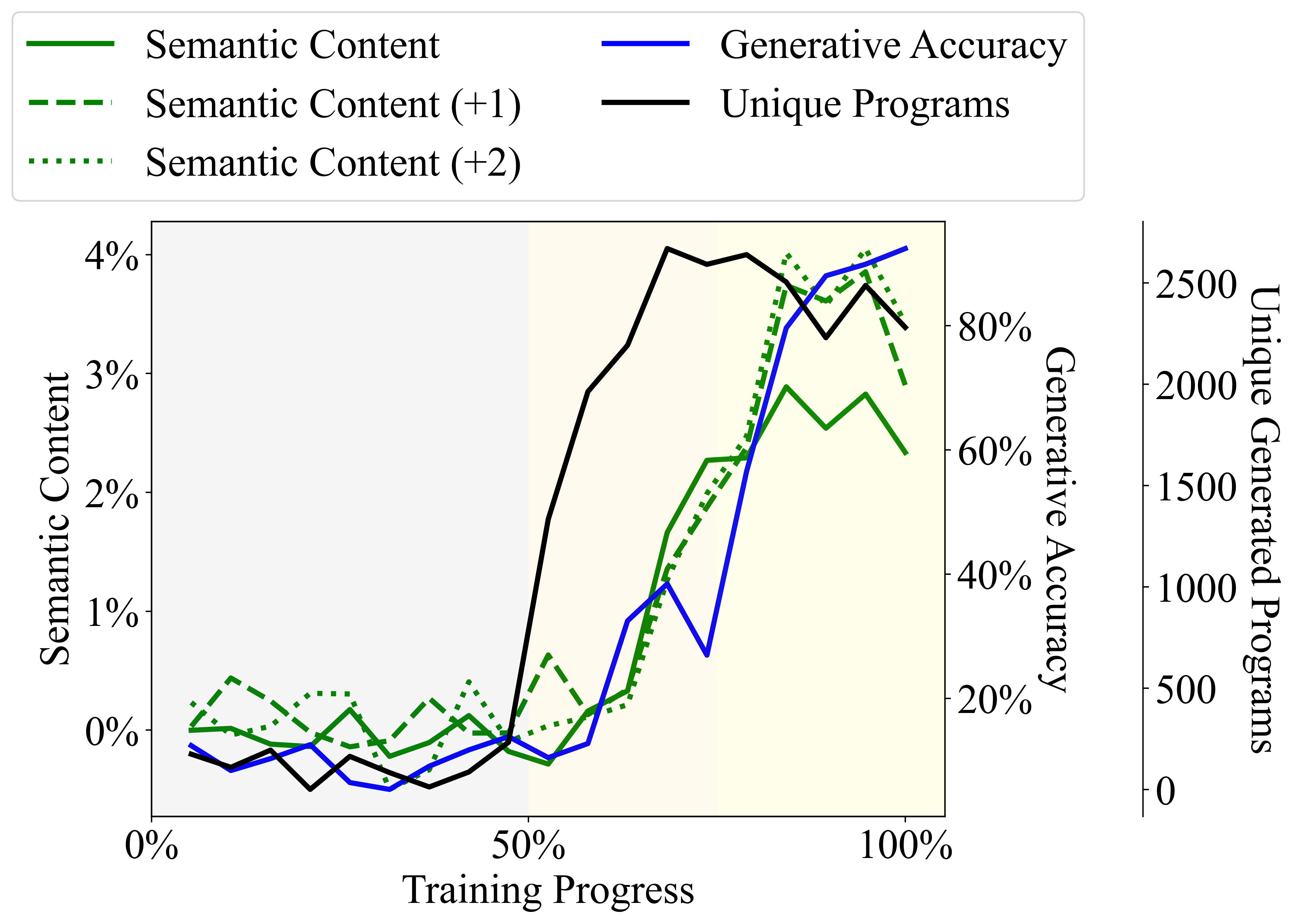}
  \end{center}
  \caption{Excess of original over flip semantic content for current and future abstract states (1-layer MLP) .}
  \label{fig:alt_intent_over_time}
\end{figure}

\Cref{table:main} displays the results of our semantic intervention baseline, where we trained probes to predict up to two abstract states into the past and future using the original and alternative semantics. In all 5 cases, the semantic content for the alternative semantics is significantly degraded when compared to the original semantics, which supports rejecting the hypothesis that the model states only encode a syntactic record (i.e., lexical and syntactic information only) while the probe learns to interpret the record (i.e., semantics).

Note that the flip semantics are strongly related to the original semantics: absent any obstacles in the robot's path, they only require reflecting the original path of the robot across an axis; in contrast, the adversarial semantics completely changes the semantics of each operator. Hence, if the syntactic record hypothesis is false, then we would expect the semantic content to be lower for adversarial vs. flip, since it should be more challenging to map from the original to the adversarial semantics; our results support this interpretation.

We also plot the excess of the original over the flip semantic contents in \Cref{fig:alt_gen_vs_sem,fig:alt_intent_over_time}. Note that the apparently high semantic content of the babbling phase---which was observed pre-intervention in \Cref{fig:gen_vs_sem,fig:intent_over_time}, respectively, and attributed to the probe being able to learn the semantics of a small number of unique generated programs (black)---disappears post-intervention. This is consistent with the probe learning the semantics equally well for both the original and flip semantics during the babbling phase, and demonstrates the ability of the semantic intervention baseline to control for the ability of the probe to learn semantics. We conclude that a statistically significant portion of the observed semantic content from \Cref{sec:meaning} cannot be explained as the probe learning semantics, refuting \MH/.

\section{Related work}

\paragraph{LMs, semantics, and interpretability}
While many works have evaluated the external behavior of LMs on a range of semantically meaningful tasks~\citep{austin2021program,toshniwal2022chess,patel2022mapping,liu2023minds}, our work explores the internal state of the LM, falling under the broad umbrella of efforts toward LM interpretability. For instance, \citet{abdou2021can} find that pretrained LMs' representations of color terms are geometrically aligned with CIELAB, a color space based on human perception.
\citet{li2021implicit} fine-tune pretrained LMs on text that describes evolving situations, then probe whether the model states track entity states. Unlike this research, we study an LM trained from scratch, which (1) yields insights into how semantics emerge in the representations of LMs over time and (2) allows us to rigorously evaluate alternative explanations of the LM behavior.
\citet{li2023emergent} train a Transformer on transcripts of Othello, then use probes to intervene on the LM's internal representations; they find that the LM's subsequent generations are consistent with the edited version of the extracted board state. In contrast, our semantic probing interventions introduce a novel method of distinguishing the contributions of the LM and probe.
To the best of our knowledge, we are also the first to apply probing to find evidence that the LM encodes the meaning of text ahead of generation.

\vspace{-.5em}
\paragraph{Grounding programs from text}
The specific question of whether LMs of \emph{code} can ground programs from text has received prior attention in the literature. \citet{merrill2021provable} show that there exist programs whose semantics provably cannot be learned from text, albeit under strong assumptions not satisfied by our setting.
\citet{bender2020climbing} concede that meaning could be learned from programs paired with unit tests, but assert this requires a ``learner which has been equipped by its human developer with the ability to identify and interpret unit tests.'' Our research, in contrast, provides empirical evidence that LMs trained only to predict the next token can, in fact, learn aspects of the program semantics.

\vspace{-.5em}
\paragraph{Program synthesis with LMs}
There is a growing body of work on training large-scale, Transformer-based LMs for program synthesis \citep{austin2021program,chen2021evaluating,li2022competition,nijkamp2023codegen,fried2023incoder,rozière2023code}, as well as program synthesis as a benchmark for LMs \citep{hendrycks2021measuring,liang2022holistic}. Several of these works have observed that the BLEU score with respect to a reference solution is not a good predictor of the LM's competency, which complements our results regarding the LM's perplexity on the training corpus.

\vspace{-.5em}
\paragraph{Probing}
Probing \citep{shi-etal-2016-string,belinkov-glass-2019-analysis} is widely used as a technique to investigate the inner workings of LMs. A key challenge is controlling for what is learned by the probe rather than represented in the LM \citep{belinkov-2022-probing}. The standard methodology is to establish a baseline measurement on a task for which the model states are assumed to be meaningless.
\citet{hewitt-liang-2019-designing} develop \emph{control tasks} for word-level properties in the context of probing for parts of speech in LM representations. They compare against the performance of a probe that maps from the model states to a dataset with a \emph{random} part of speech assigned to each word. In our case, the control task approach would assign random features to each program state; however, this also destroys the syntactic structure of the program, and hence cannot be used to test the syntactic record hypothesis.
To address this, we introduce \emph{semantic} probing interventions, a control framework for probing that intervenes on the semantics of individual operations while preserving the overall syntax of programs. As our techniques specifically advance the study of semantics in LMs, 
we believe our contributions can be broadly applicable to future interpretability research, particularly for investigations involving meaning (rather just than syntactic structure).

\section{Conclusion}

This paper presents empirical evidence that \textbf{LMs of code can acquire the formal semantics of programs from next token prediction}. We find that, when training an LM to model text consisting of examples of input-output specifications followed by programs, the learning process of the LM appears to undergo 3 distinct phases, with the second half of training characterized by a strong, linear correlation between the emerging representations of the semantics and the ability of the LM to synthesize programs that correctly implement unseen specifications. We also find representations of \emph{future} semantics, suggesting a notion of intent during generation. Further explorations of these dynamics could yield deeper insights into the behavior of LMs.

We also present \textbf{semantic probing interventions}, a framework for the application of probes---a standard tool for interpreting the learned representations of, e.g., neural models---to understanding whether representations capture information related to the underlying semantics of a domain.
Specifically, we design experiments capable of distinguishing whether the probe's measurement is indicative of (1) the presence of semantic information intrinsic to the representations or (2) the ability of the probe to perform the task itself, with purely syntactic information encoded in the representations. 
This also allows us to justify the use of nonlinear probes that, absent our technique, are more likely yield false positives due to having more capacity to learn the task; we see moving beyond shallow probes as a way to progress toward understanding whether (and how) LMs represent more complex concepts.

More broadly, the question of what exactly LMs are learning from text has garnered considerable interest in recent years, driven by the increasingly impressive performance of frontier models. We believe the techniques and insights presented in this work can serve as a principled foundation for future studies of the capabilities and limitations of LMs.

\clearpage

\section*{Acknowledgements}

We would like to thank Jacob Andreas, Omar Costilla-Reyes, Kai Jia, Jason Kim, Armando Solar-Lezama, and Yichen Yang for their helpful comments and discussions on an earlier version of this paper. We also gratefully acknowledge support from DARPA Grants HR001120C0015, HR001120C0191, and N6600120C4025. The views expressed are those of the authors and do not reflect the official policy or position of the Department of Defense or the United States Government.

\section*{Impact statement}

This paper presents work whose goal is to advance the field of Machine Learning. There are many potential societal consequences of our work, none which we feel must be specifically highlighted here.

To aid in reproducibility, we open source all our code, including the code we use to generate the training data, train the LM, and conduct the probing experiments, at \url{https://github.com/charlesjin/emergent-semantics}.

\bibliography{references}
\bibliographystyle{icml2024}

\newpage
\appendix
\onecolumn

\appendix
\section{Experimental details}
\label{appendix:LM_details}

\subsection{Karel grammar specification}

We use the same grammar as \citet{devlin2017neural}, except with loops and conditionals removed.
\begin{align*}
\text{Prog } p &:=  \texttt{def run():} s \\
\text{Stmt } s &:=  s_1; s_2 \mid a \\
\text{Action } a &:=
     \texttt{move()}
    \mid \texttt{turnRight()}
    \mid \texttt{turnLeft()}
    \mid  \texttt{pickMarker()}
    \mid \texttt{putMarker()}
\end{align*}

\subsection{Language model and probe details}

We train the 350M-parameter variant of the CodeGen architecture \citep{nijkamp2023codegen} from the HuggingFace Transformers library \citep{huggingface}, implemented in PyTorch \citep{paszke2019pytorch}.
We use the Adam optimizer, %
a learning rate of 5e-5, a block size of 2048, and a batch size of 32768 tokens. We train for 2.5 billion tokens, which was close to 6 passes over our training corpus; we did not observe any instabilities with training (see the results in the main text). We use a warm up over roughly the first 3000 batches, then linearly decay the learning rate to 0 after 80000 batches (training runs over 76000 batches). On a single NVIDIA A100 GPU with 80GB of VRAM, training takes around 8 days.

The linear probe consists of a single linear layer.
The MLP probes consist of stacked linear, batch norm, and ReLU layers, in that order. The first linear layer in both MLP probes projects to a hidden dimension of 256, and the second linear layer (in the 2-layer MLP) projects to a hidden dimension of 1024.
Probes are trained on the first 100000 aligned traces in the training trace dataset. All probes are trained using the same recipe, which was tuned to saturate (or nearly saturate) their performance: we use the AdamW optimizer %
with weight decay of 1e-4 and a learning rate of 0.01 that decays by .1 at 75\% and 90\% of the way through training; and train for 10000000 steps using a batch size of 1024.

\section{Additional experimental results}
\label{appendix:additional_results}

\subsection{Plots for adversarial semantics}

\begin{figure}[tb]
\begin{subfigure}{.48\linewidth}
  \centering
  \includegraphics[height=5.5cm]{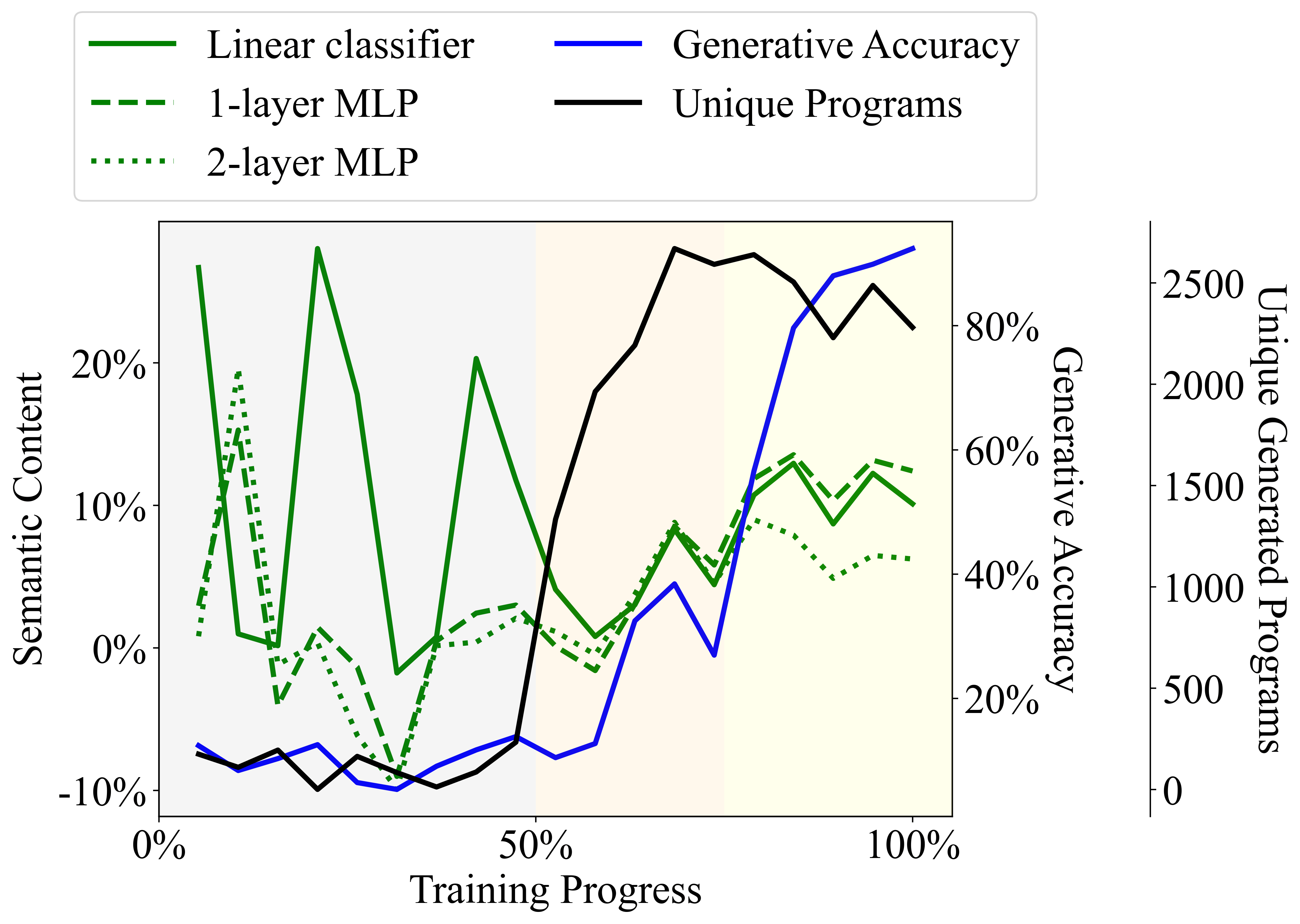}
  \caption{Excess of original over adversarial semantic content using different probing classifiers.}
  \label{app:fig:alt_adv_gen_vs_sem}
\end{subfigure}
\hspace{1em}
\begin{subfigure}{.48\linewidth}
  \centering
  \includegraphics[height=5.5cm]{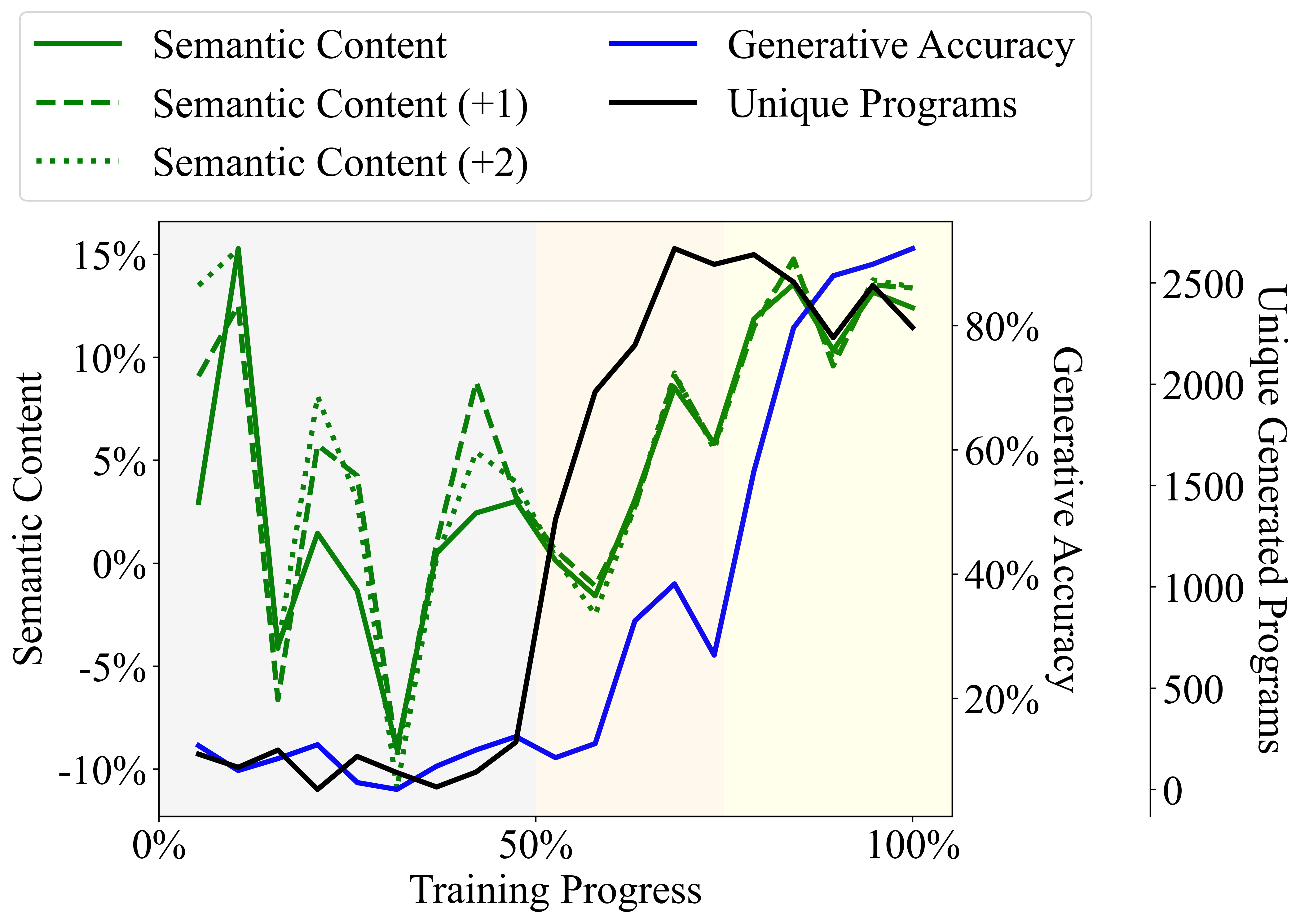}
  \caption{Excess of original over adversarial semantic content for current and future abstract states (1-layer MLP) .}
  \label{app:fig:alt_adv_intent_over_time}
\end{subfigure}
\caption{Excess of original over adversarial semantic content.}
\label{app:fig:alt_adv}

\end{figure}

\Cref{app:fig:alt_adv_gen_vs_sem,app:fig:alt_adv_intent_over_time} provide plots for the adversarial semantics that are analogous to those provided in the main text for the flip semantics (specifically \Cref{fig:alt_gen_vs_sem,fig:alt_intent_over_time}, respectively). We observe that, in contrast to the flip semantics (which achieve close to zero excess semantic content in the babbling phase), the excess of the original over adversarial semantics exhibit significant noise during the babbling phase of the LM training, which is often negative; we attribute this to weaker relationship between the adversarial and original semantics conjectured in \Cref{sec:semantic_intervention}: certain distributions of programs may be easier for the adversarial vs. original semantics, and vice versa. We also note that the excess semantic content is largely constant across the current and future states in \Cref{app:fig:alt_adv_intent_over_time}, particularly in the final phases of training, which is consistent with the probe learning the same map from the original abstract state to the adversarial abstract state (and thus having roughly the same absolute error rate, independent of how far into the future the abstract states are relative to the model state).

\FloatBarrier
\subsection{Generated programs become shorter than reference programs over training}

\begin{figure}[tb]
  \begin{center}
    \includegraphics[height=6cm]{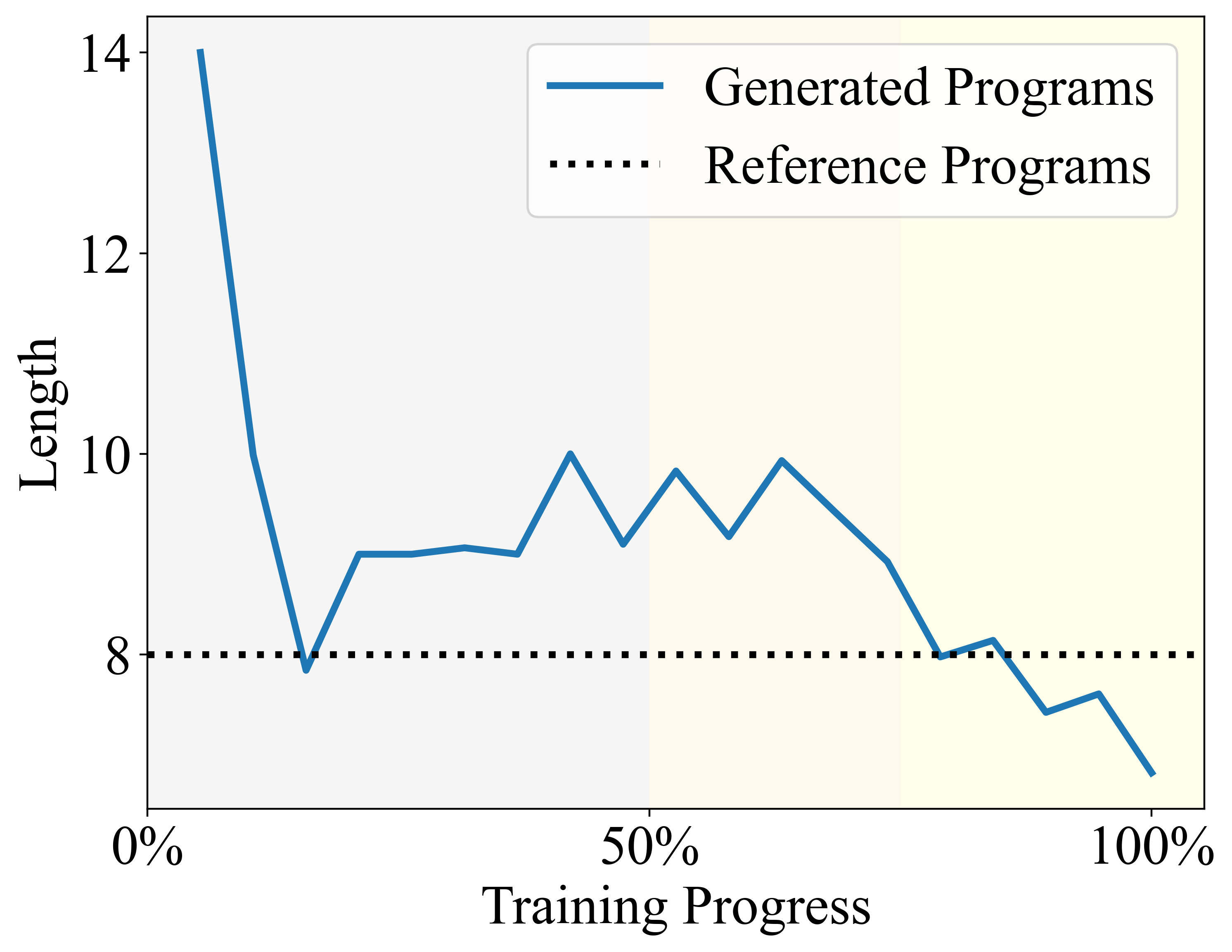}
  \end{center}
  \caption{The average length of generated programs over time generated from specifications sampled according to the training distribution and compared with the corresponding reference programs.}
  \label{fig:lengths_over_time}
\end{figure}

\Cref{fig:lengths_over_time} plots the average length of programs generated by the LMs over specifications sampled from the training distribution (i.e., where the reference programs are of length 6 to 10, inclusive). The results indicate that the LM learns to generate programs which are, on average, shorter than the reference program length. While this can be partially explained by the fact that we use greedy decoding to generate the outputs, we emphasize that the generated programs are also grow increasingly correct over the second half of training, i.e., the quality of the LM's generation continues to improve despite diverging from the distribution of the training corpus.

These results complements our findings in the main text, which show that the LM tends to underfit the distribution of tokens in the training data. Although prior work has explored the differences between an LM's output and its training corpus based on surface statistics \citep{meister2021language,lebrun2022evaluating}, we are, to the best of our knowledge, the first to present an account of how such syntactic divergence relate to the semantic properties of the LM generation and training data.

\FloatBarrier
\subsection{Semantic content of reference programs}
\label{appendix:reference_semantics}

This section explores a possible implication of the finding in \Cref{sec:meaning} that representations contain predictions of \emph{future} program states. In particular, these future program states are defined by the tokens produced by greedy decoding, i.e., the program is generated by taking the token that the LM models as most likely at each step in the autoregressive loop of \Cref{eq:trace_dataset}. However, it has been observed in other settings that sampling from the tokens according to a distribution derived from the logits of the LM head can improve the quality of the LM's outputs, e.g., with multinomial or nucleus sampling \citep{Holtzman2020The}. However, if the LM states do in fact encode the future program states according to greedy decoding, then any deviation could, in fact, force the LM states to visit ``unanticipated'' program states.

We perform a preliminary exploration of this hypothesis. Specifically, we measure the semantic content when decoding from the LM according to the \emph{reference} programs used to generate the specifications in our train and test datasets: we create new \textbf{reference trace datasets} by replacing the greedy decoding step of \Cref{eq:greedy_decode} with the tokens from the reference programs, i.e.,
\begin{align}
\label{eq:reference_trace_dataset}
(\text{state}_{LM})_i = {LM}(\text{input}, \text{output}, \{(\text{state}_{LM})_j\}_{j=1}^{i-1}) \\
\text{token}_i = \text{prog}_\text{ref}[i] \\
(\text{state}_{\text{prog}})_i = \text{exec}(\text{token}_i, (\text{state}_{\text{prog}})_{i-1})
\end{align}
where $\text{prog}_\text{ref}$ is the reference program and $\text{prog}_\text{ref}[i]$ is the $i^{th}$ program token in the reference program.
The idea is that, because the LM is specifically trained to minimize its perplexity on the reference programs, we might expect the LM representations to also be closely aligned with the semantics of the training corpus.

\begin{figure*}
\centering
\begin{subfigure}{.3\linewidth}
  \centering
  \includegraphics[height=.75\linewidth]{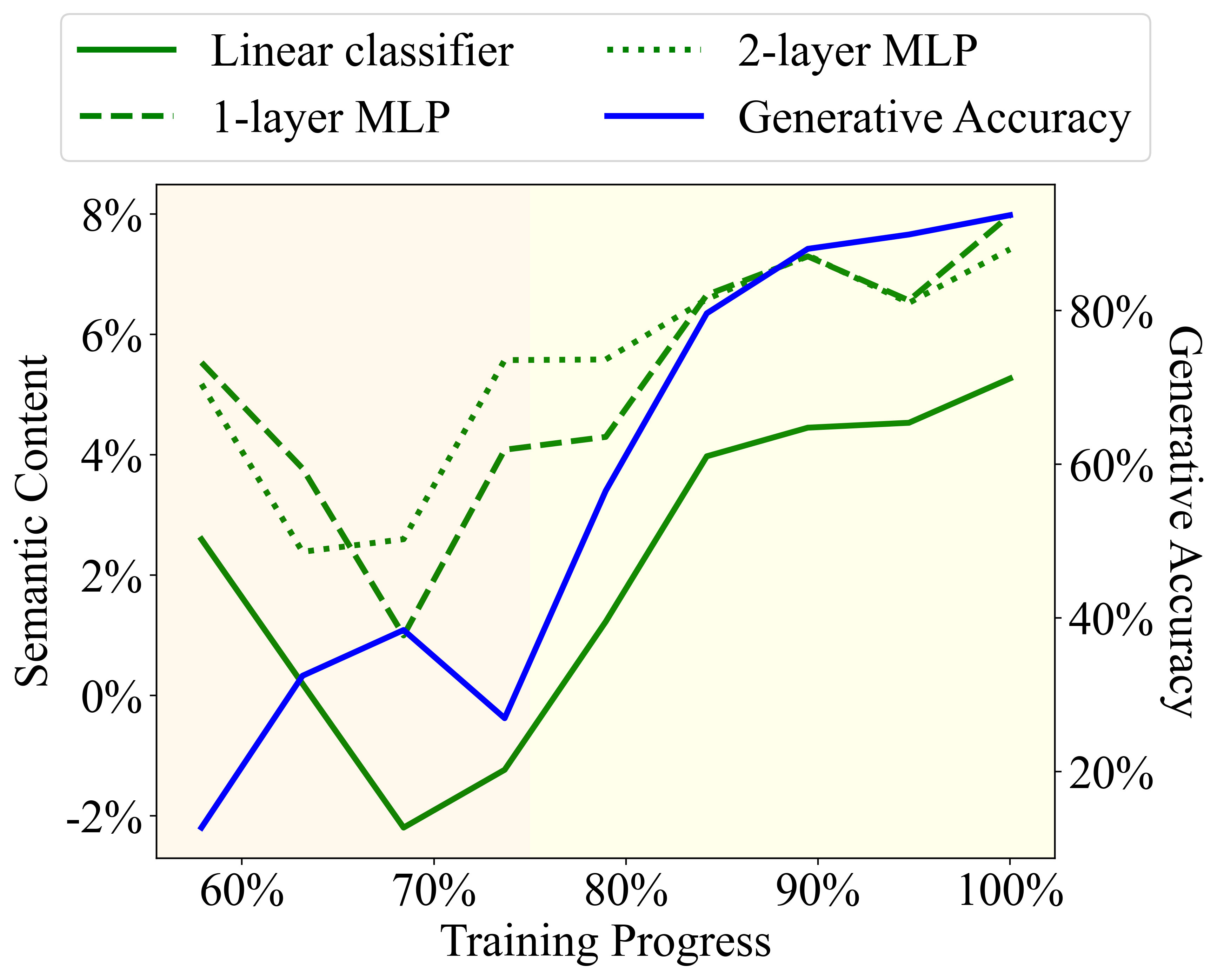}
  \caption{Probing 2 states into the past.}
  \label{fig:forced_sem:sub1}
\end{subfigure}
\hspace{1em}
\begin{subfigure}{.3\linewidth}
  \centering
  \includegraphics[height=.75\linewidth]{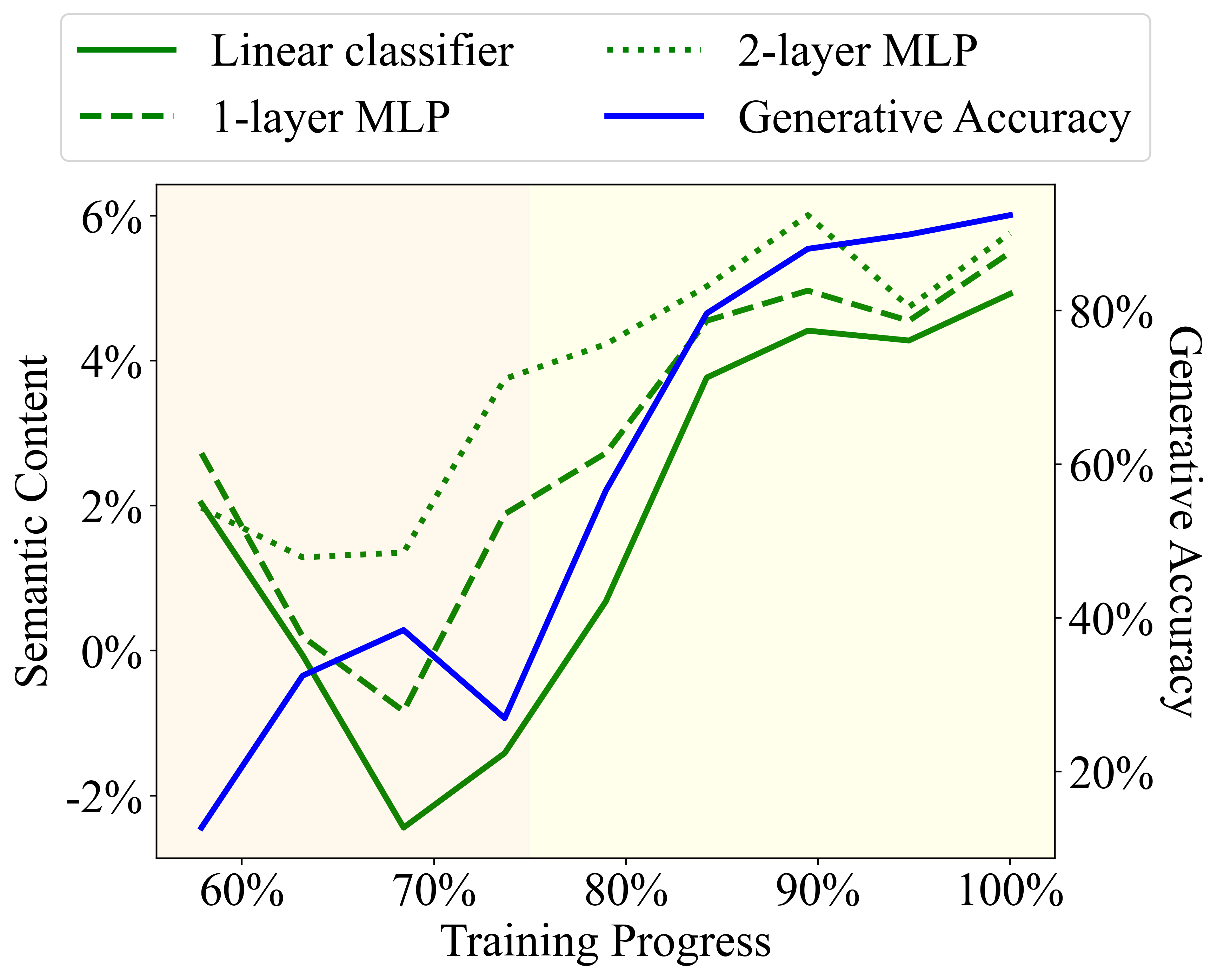}
  \caption{Probing 1 state into the past.}
  \label{fig:forced_sem:n1}
\end{subfigure}
\hspace{1em}
\begin{subfigure}{.3\linewidth}
  \centering
  \includegraphics[height=.75\linewidth]{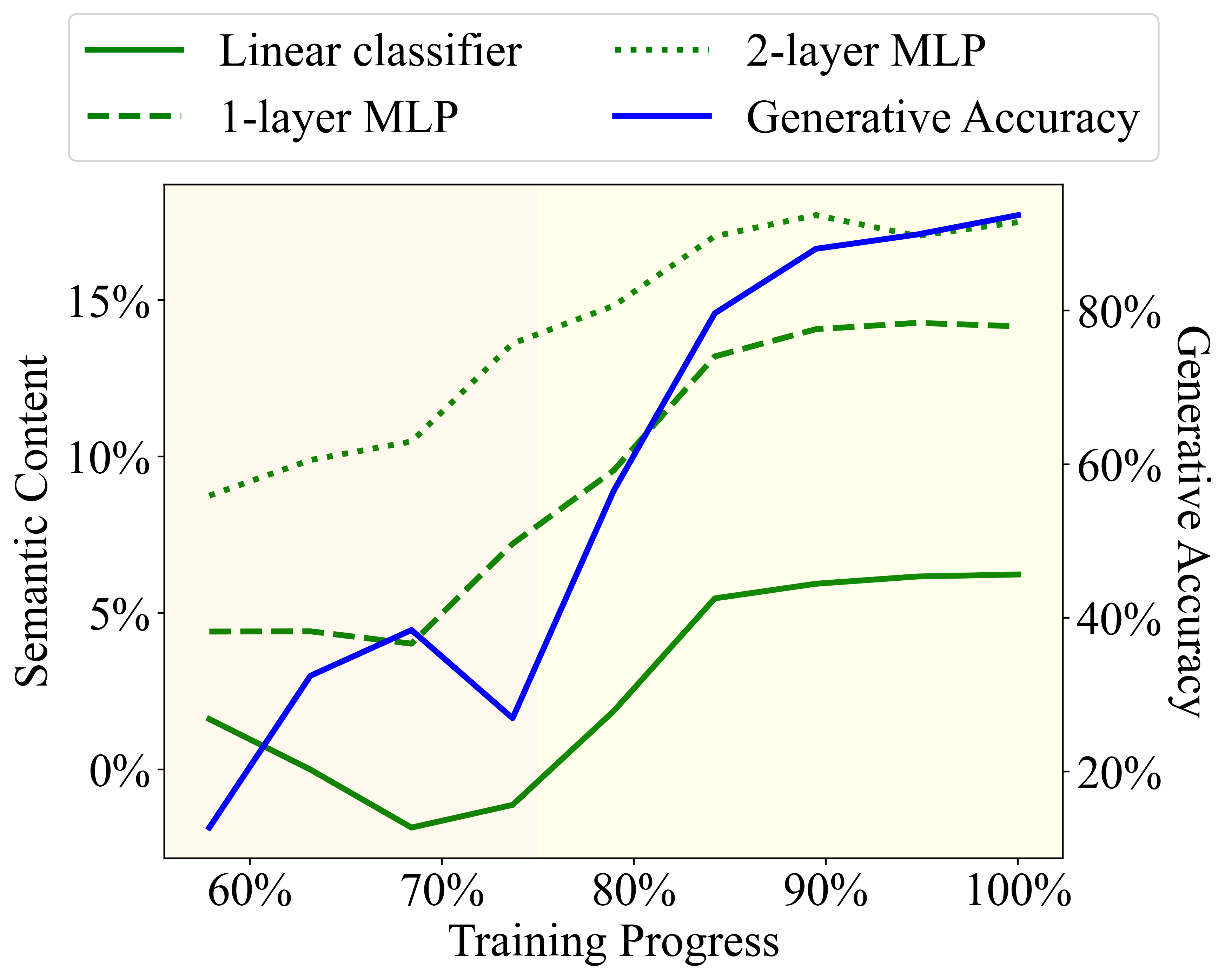}
  \caption{Probing the current state.}
  \label{fig:forced_sem:0}
\end{subfigure} \\ \vspace{1em}
\begin{subfigure}{.3\linewidth}
  \centering
  \includegraphics[height=.75\linewidth]{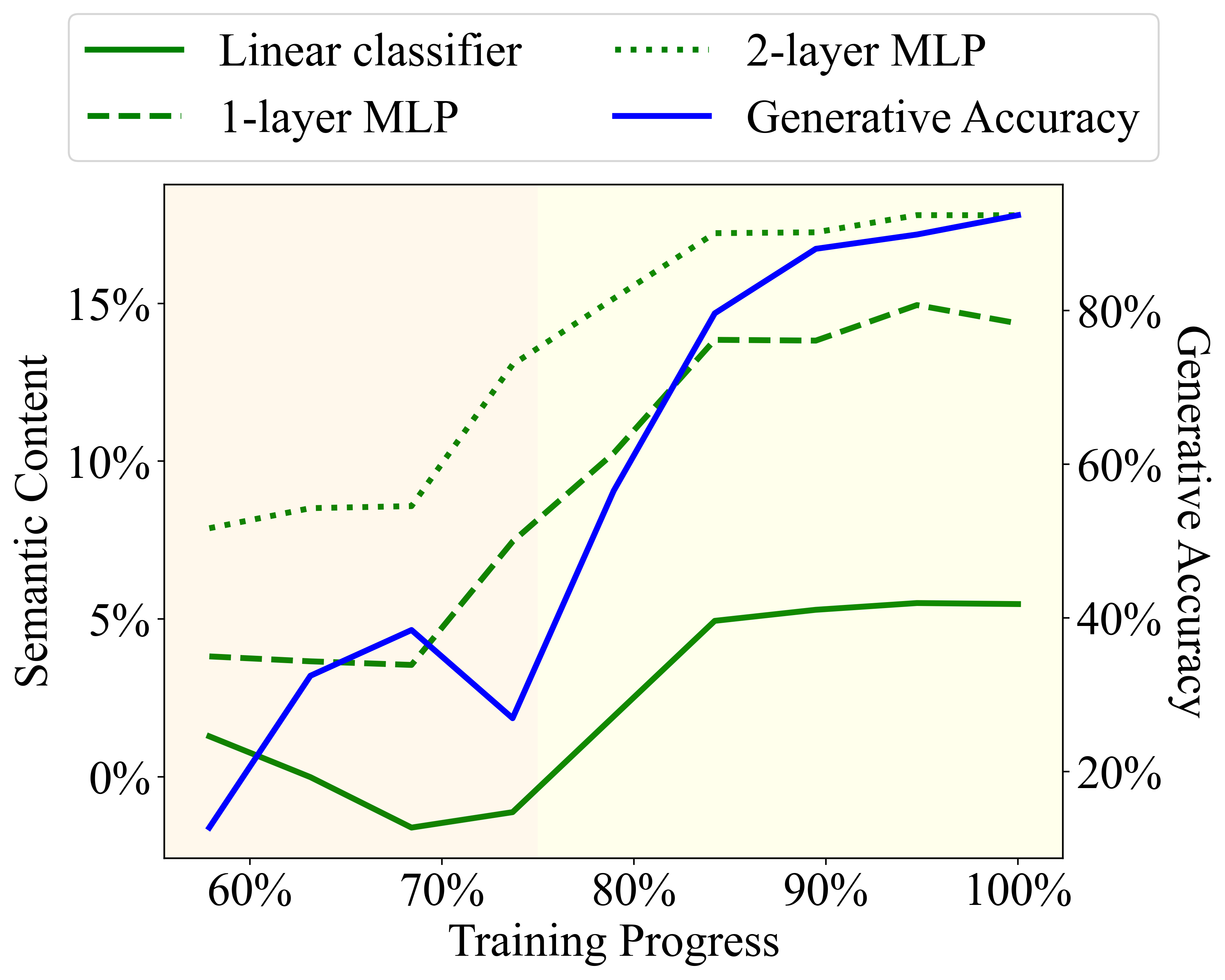}
  \caption{Probing 1 state into the future.}
  \label{fig:forced_sem:p1}
\end{subfigure}
\hspace{1em}
\begin{subfigure}{.3\linewidth}
  \centering
  \includegraphics[height=.75\linewidth]{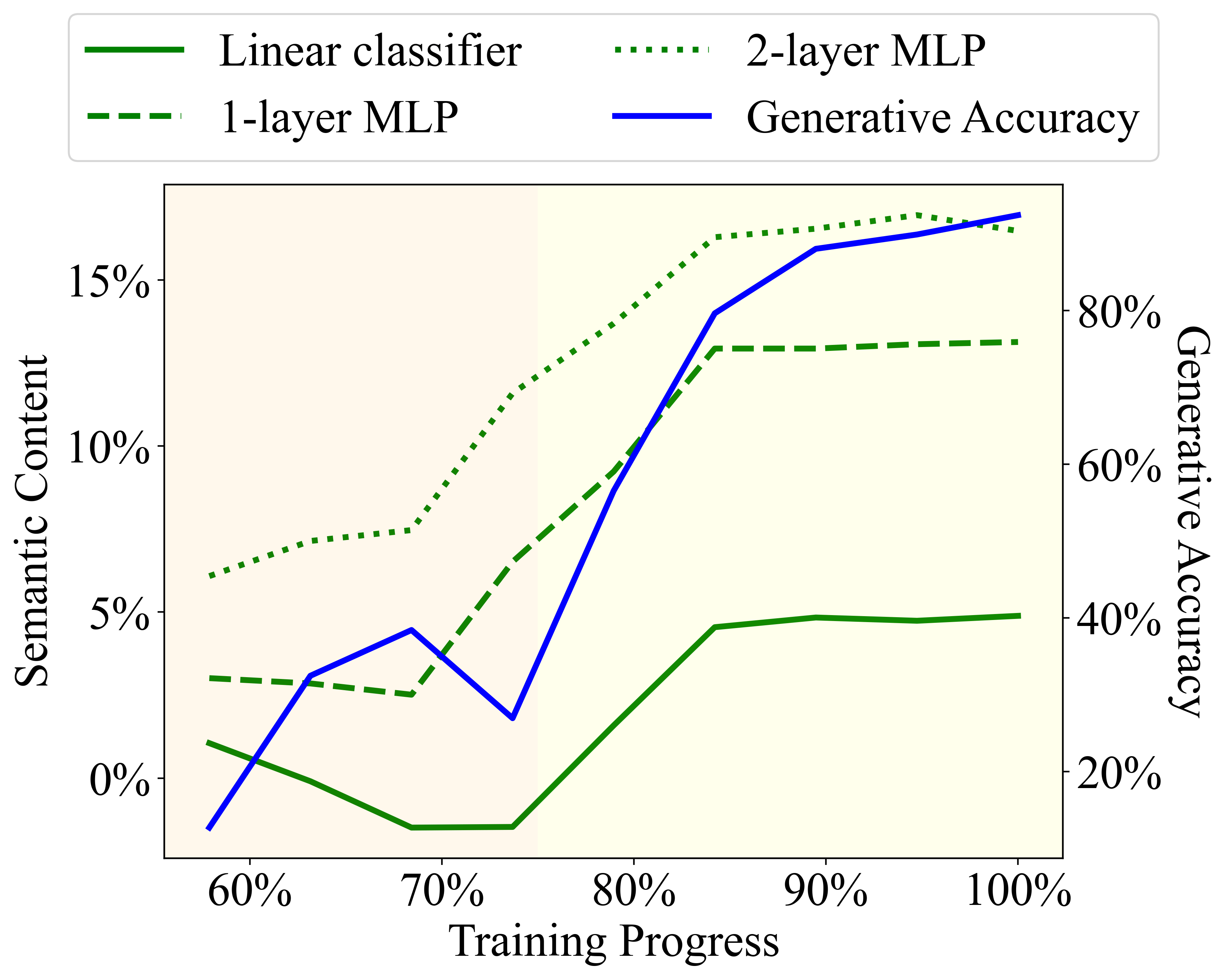}
  \caption{Probing 2 states into the future.}
  \label{fig:forced_sem:p2}
\end{subfigure}
\caption{Plotting the difference in the semantic content when decoding according to the reference programs vs. greedy decoding. A positive difference indicates that greedy decoding yields a higher semantic content.}
\label{fig:forced_sem}
\end{figure*}

\Cref{fig:forced_sem} plots the results of this experiment. For the linear probe, extracting both past and future semantic states does appear easier up until the beginning of the last phase of training. However, by the end of training, probing for the abstract states from greedy decoding all exhibit higher semantic content than the reference programs, which suggests that the LM representations are indeed better aligned with greedy decoding.
We also observe that the difference is particularly large for the present and future semantic states. This can be explained by the fact that the future program states are essentially a random process (due to the program itself being generated according to a random sampling procedure), and hence, there is a fundamental limit to how informative the LM's representation can be. Note that even the ``current'' abstract state is a random process, as the the current abstract state also results from a token that the LM has yet to see: the probe's task for the current abstract state is to predict $(\text{state}_{\text{prog}})_i$ from $(\text{state}_{LM})_i$ in \Cref{eq:reference_trace_dataset}.

Finally, we note that the difference (between the semantic contents with respect to the reference programs vs. generated programs) grows as the LM's generative accuracy improves; we conjecture that the LM's preference for greedy decoding may increase as it becomes better calibrated to the training data as a general principle.
We leave a more detailed exploration of how different decoding strategies affect the coherence of the LM's internal state to future work.

\FloatBarrier
\subsection{Semantics are inferred, not retrieved}
\label{appendix:retrieval}

\begin{table}
\caption{The semantic content at the end of training, separated by the depth of the program state and the 3 features in the abstract state. The LM only observes program states at depth 6 or greater in the training corpus. We display depths consisting of at least 1\% of the training set.}
\label{table:depth_main}
\begin{center}
\begin{tabular}{ cc|ccc|ccc|ccc}
\toprule
 & \multirow{2}{*}{depth} & \multicolumn{3}{c|}{linear} & \multicolumn{3}{c|}{MLP-1} & \multicolumn{3}{c}{MLP-2} \\
 & & direction & position & obstacle & direction & position & obstacle & direction & position & obstacle \\
 \midrule

\multirow{5}{*}{\rotatebox[origin=c]{90}{unseen}} & 1 & 46.7 & 85.0 & 73.7 & 76.7 & 92.9 & 78.6 & 89.1 & 93.3 & 79.5 \\
 & 2 & 47.9 & 71.3 & 73.0 & 76.4 & 85.2 & 77.7 & 87.3 & 85.7 & 78.2 \\
 & 3 & 49.8 & 64.0 & 73.5 & 76.7 & 80.1 & 78.2 & 86.8 & 80.3 & 78.5 \\
 & 4 & 50.5 & 61.4 & 74.8 & 78.8 & 76.9 & 79.1 & 87.3 & 77.3 & 79.4 \\
 & 5 & 51.5 & 60.1 & 76.0 & 79.7 & 74.9 & 80.2 & 87.1 & 74.9 & 80.4 \\
\midrule
\multirow{3}{*}{\rotatebox[origin=c]{90}{seen}} & 6 & 59.9 & 60.9 & 74.6 & 82.1 & 73.7 & 79.9 & 87.6 & 73.6 & 80.5 \\
 & 7 & 59.2 & 58.8 & 74.3 & 79.1 & 68.6 & 79.3 & 84.3 & 68.2 & 79.3 \\
 & 8 & 51.7 & 57.1 & 76.0 & 69.8 & 62.2 & 79.9 & 75.0 & 62.9 & 80.2 \\

 \bottomrule
\end{tabular}
\end{center}
\end{table}

\begin{figure*}
\centering
\begin{subfigure}{.3\linewidth}
  \centering
  \includegraphics[height=.6\linewidth]{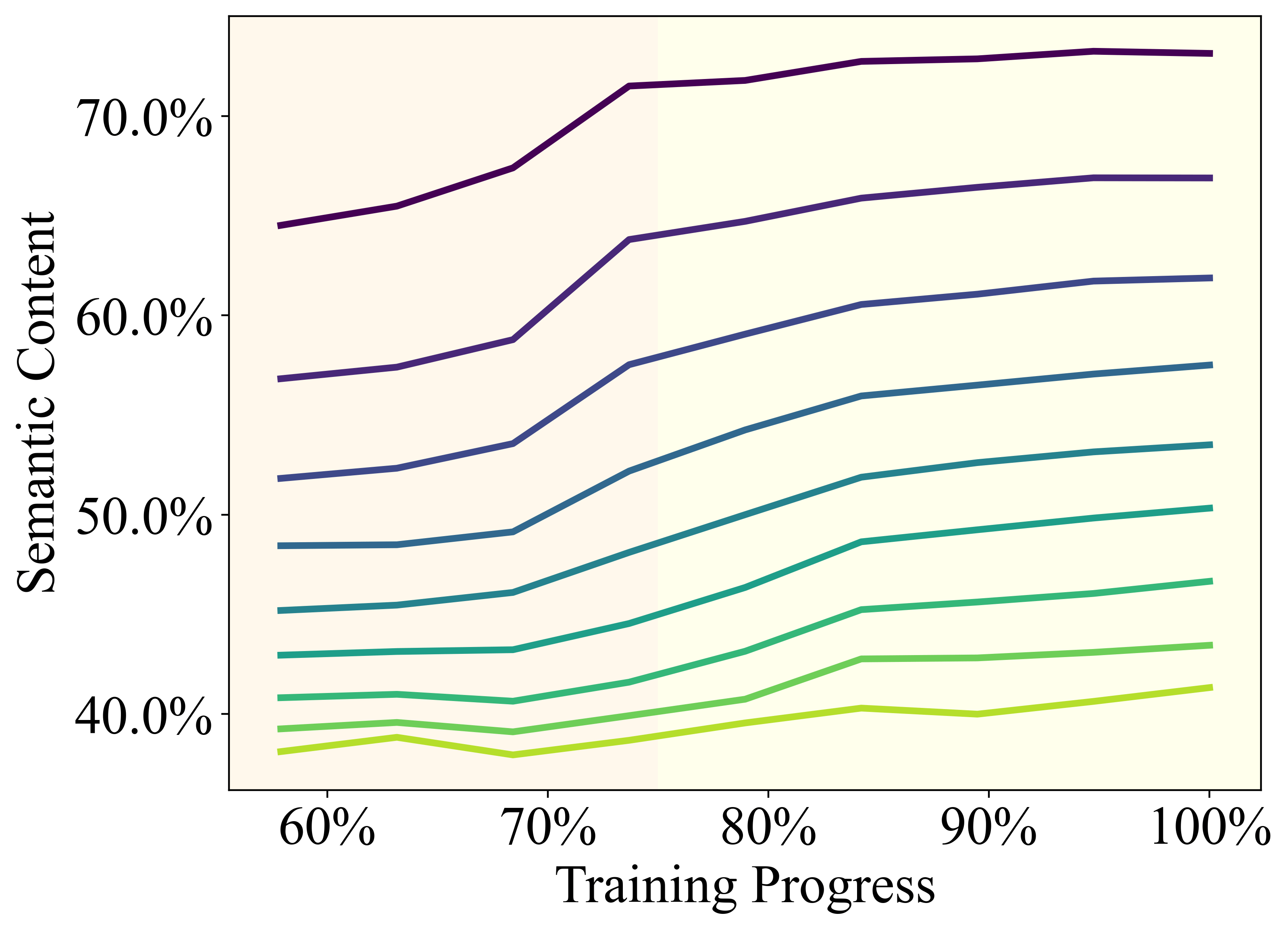}
  \caption{Probing with a linear classifier.}
  \label{fig:depth_sem:linear}
\end{subfigure}
\hspace{1em}
\begin{subfigure}{.3\linewidth}
  \centering
  \includegraphics[height=.6\linewidth]{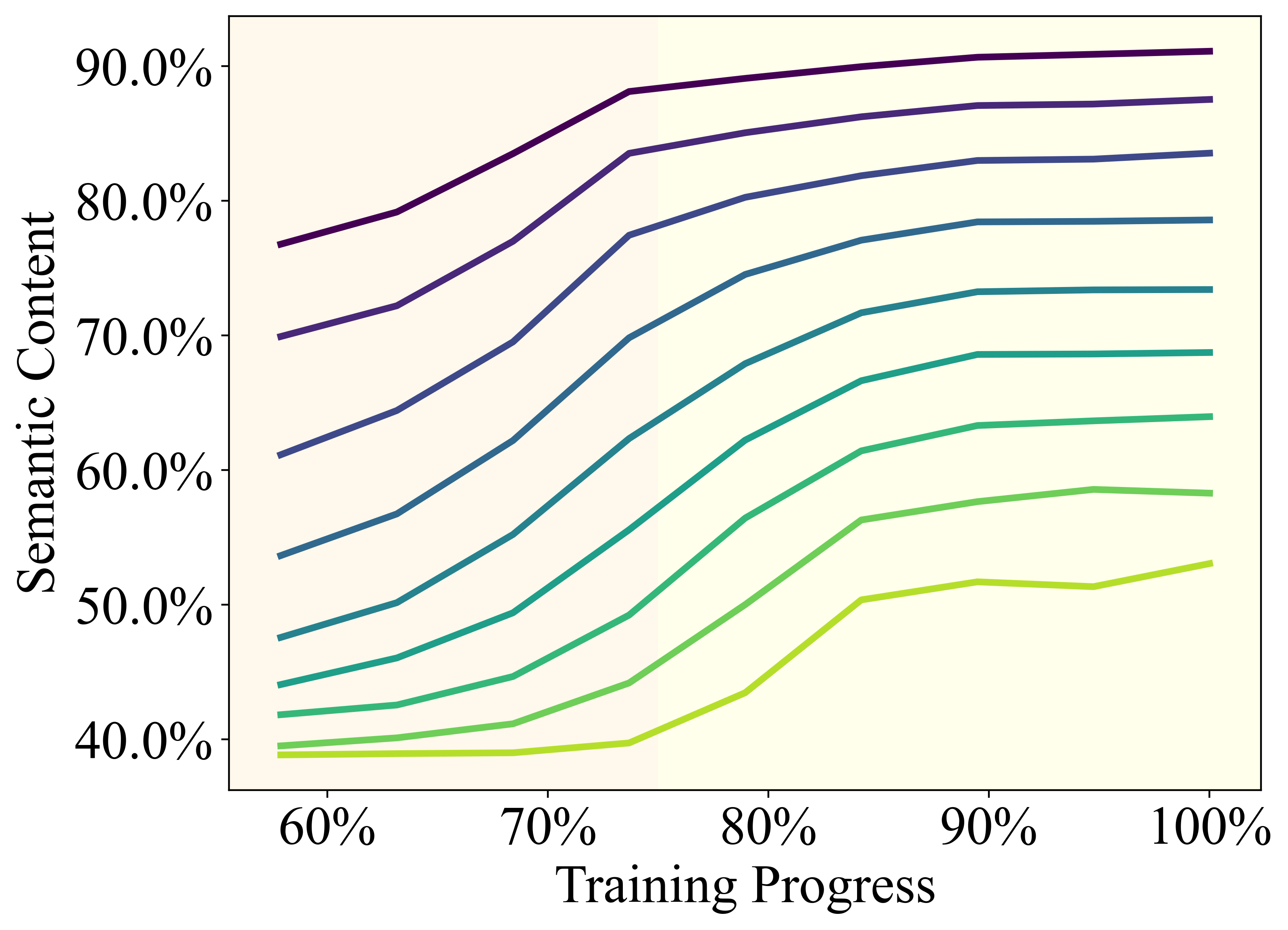}
  \caption{Probing with a 1-layer MLP.}
  \label{fig:depth_sem:mlp1}
\end{subfigure}
\hspace{1em}
\begin{subfigure}{.3\linewidth}
  \centering
  \includegraphics[height=.6\linewidth]{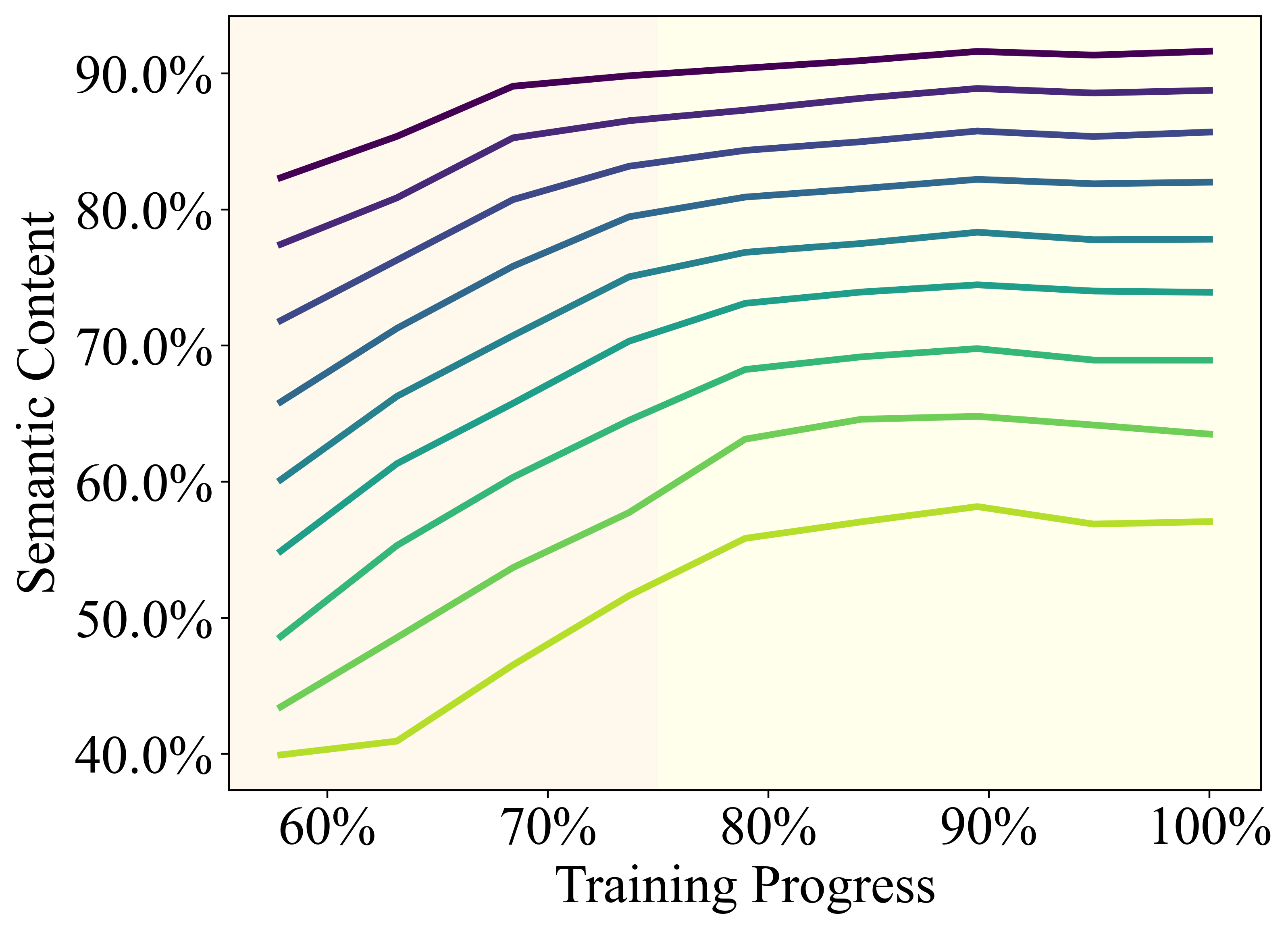}
  \caption{Probing with a 2-layer MLP.}
  \label{fig:depth_sem:mlp2}
\end{subfigure} \\ \vspace{1em}

\begin{subfigure}{.3\linewidth}
  \centering
  \includegraphics[height=.6\linewidth]{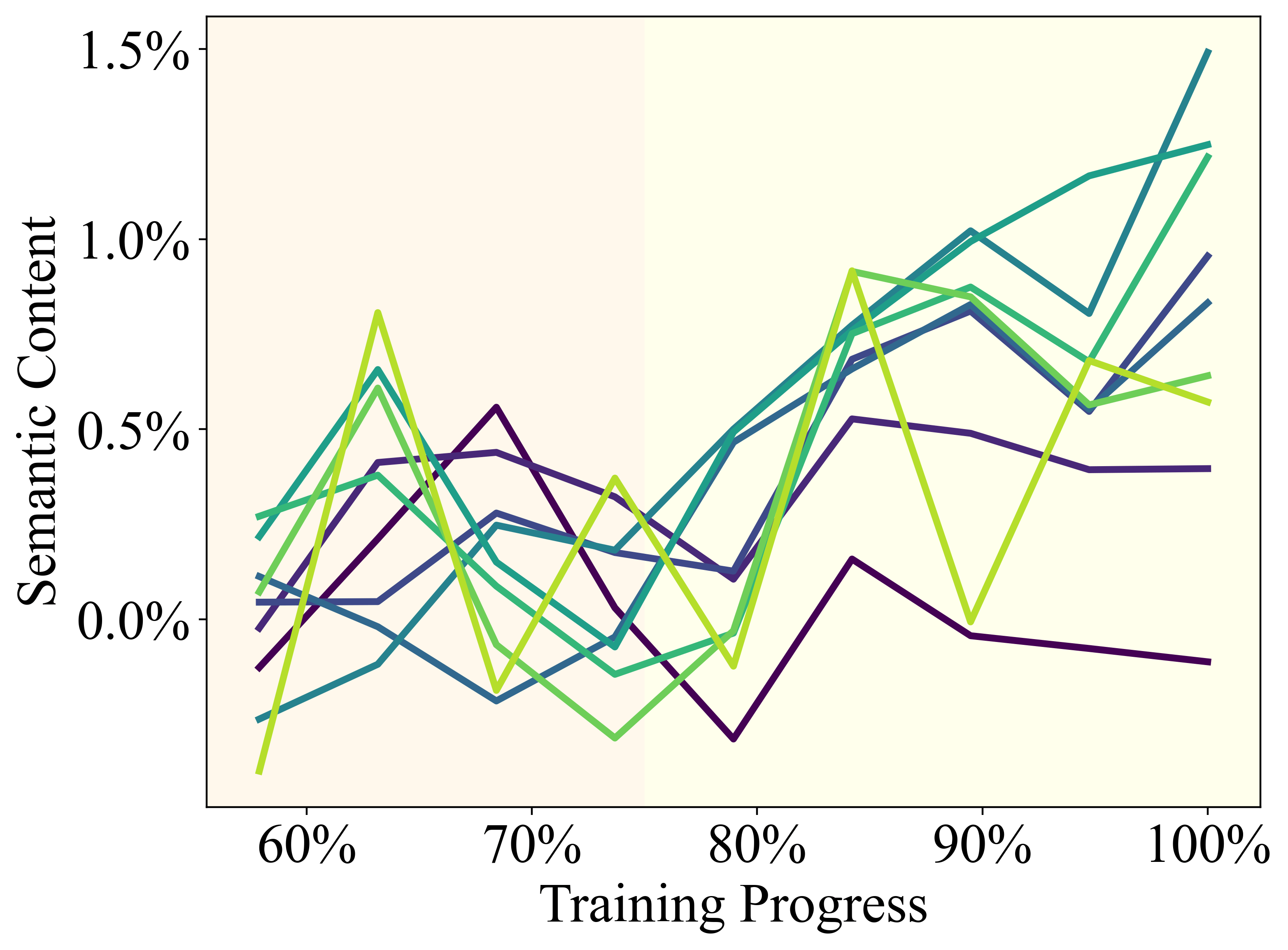}
  \caption{Probing with a linear classifier, excess over flip semantics.}
  \label{fig:depth_sem:linear:flip}
\end{subfigure}
\hspace{1em}
\begin{subfigure}{.3\linewidth}
  \centering
  \includegraphics[height=.6\linewidth]{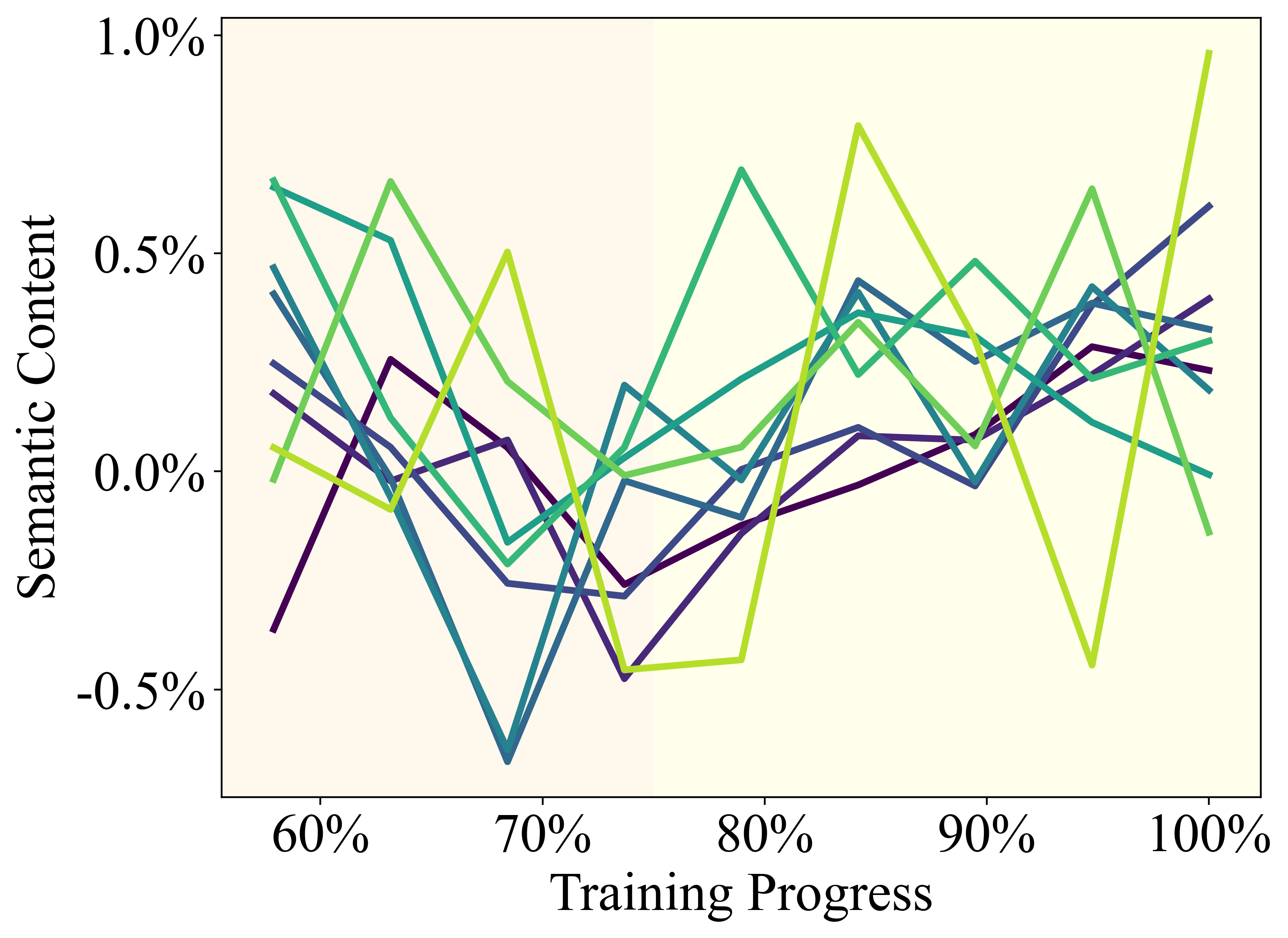}
  \caption{Probing with a 1-layer MLP, excess over flip semantics.}
  \label{fig:depth_sem:mlp1:flip}
\end{subfigure}
\hspace{1em}
\begin{subfigure}{.3\linewidth}
  \centering
  \includegraphics[height=.6\linewidth]{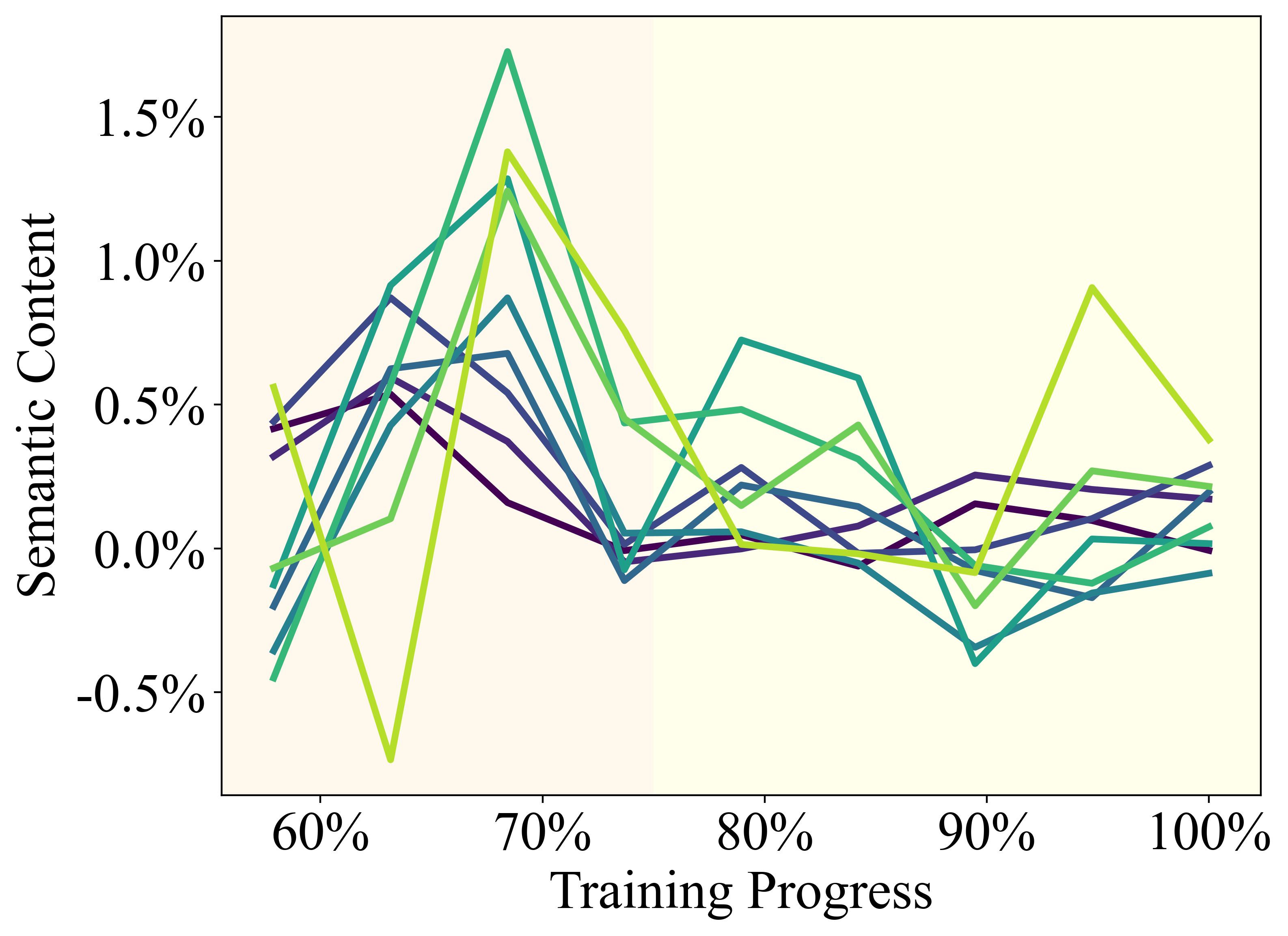}
  \caption{Probing with a 2-layer MLP, excess over flip semantics.}
  \label{fig:depth_sem:mlp2:flip}
\end{subfigure} \\ \vspace{1em}

\begin{subfigure}{.3\linewidth}
  \centering
  \includegraphics[height=.6\linewidth]{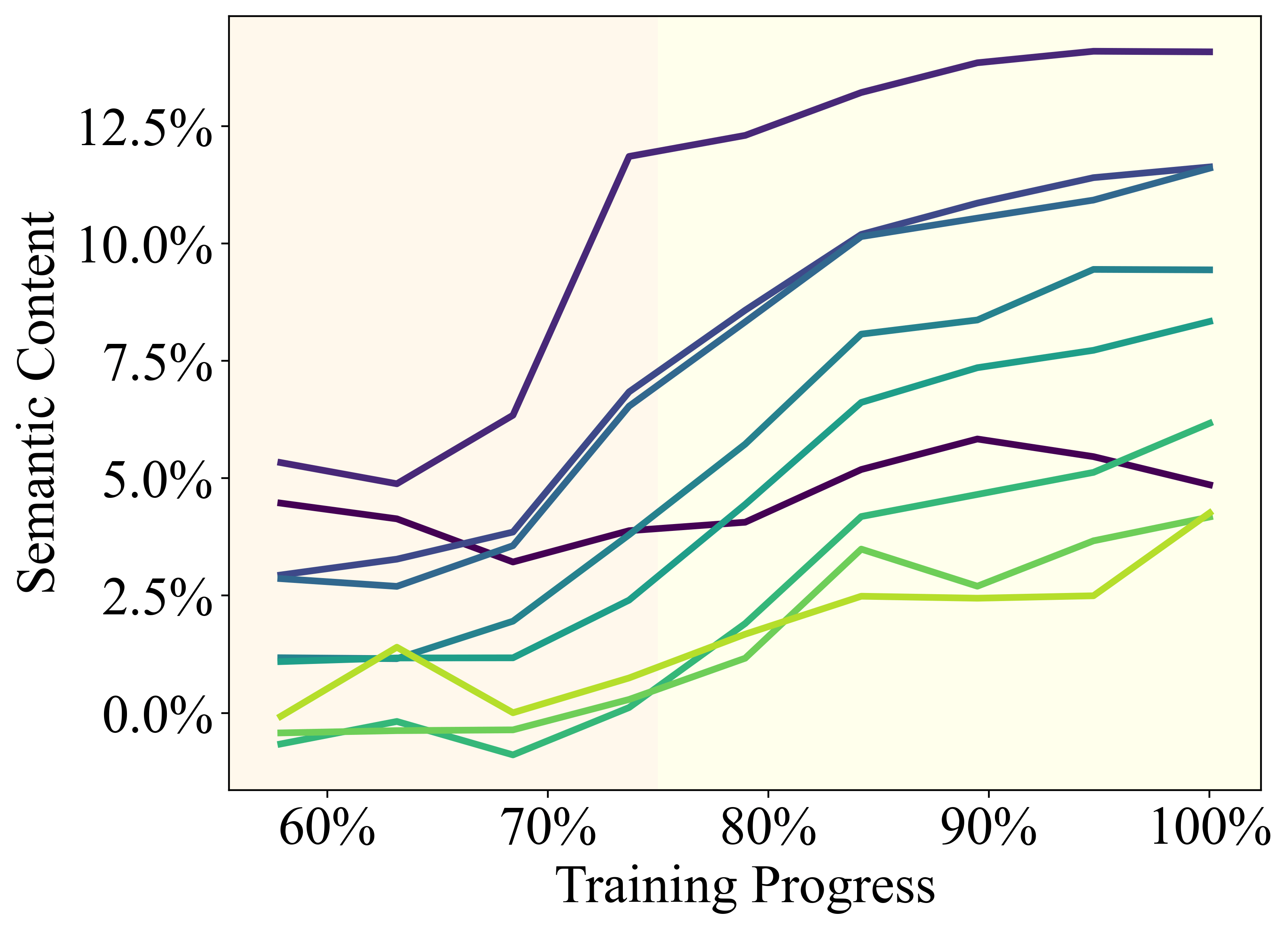}
  \caption{Probing with linear classifier, excess over adversarial semantics.}
  \label{fig:depth_sem:linear:adv}
\end{subfigure}
\hspace{1em}
\begin{subfigure}{.3\linewidth}
  \centering
  \includegraphics[height=.6\linewidth]{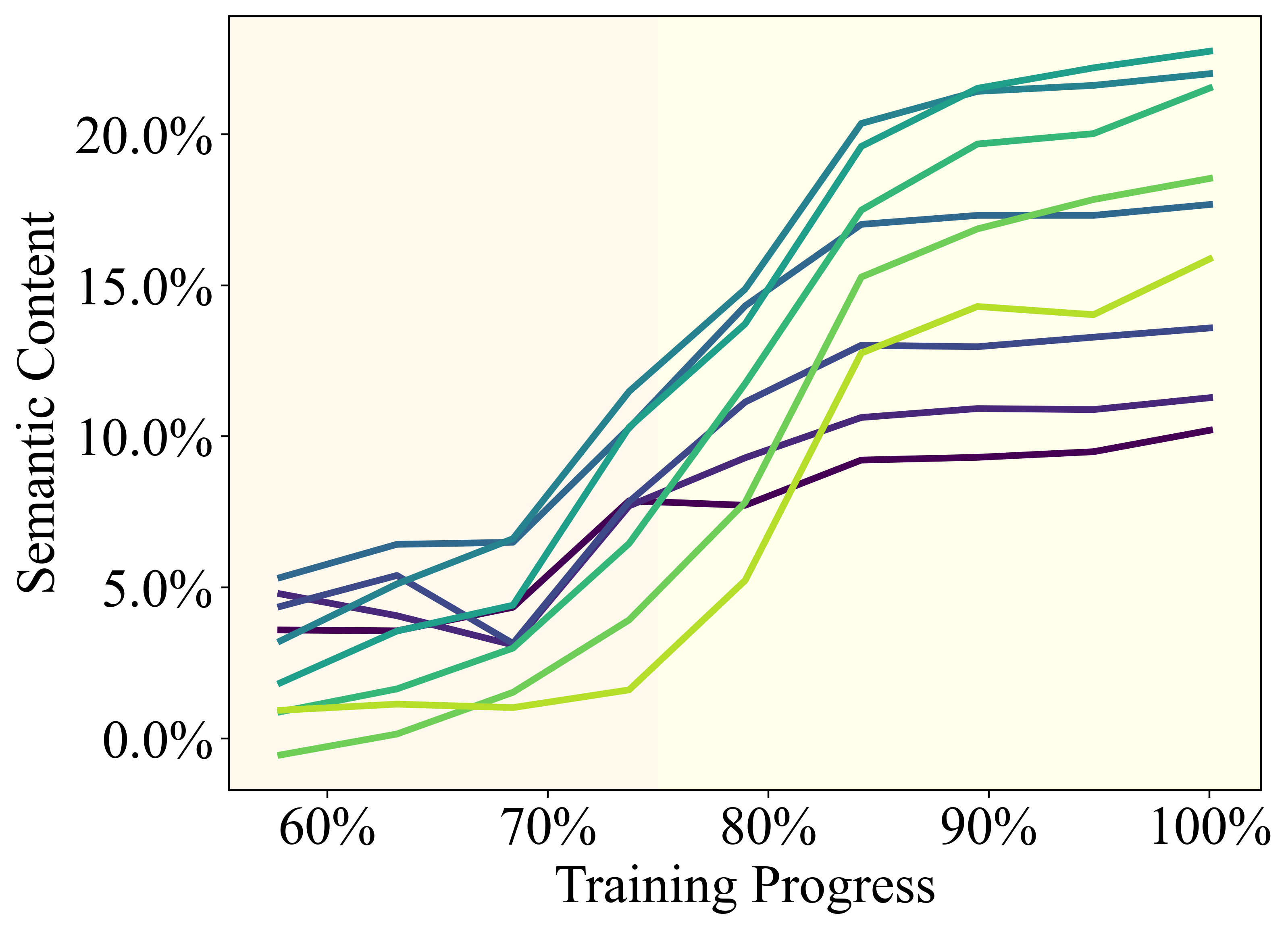}
  \caption{Probing with a 1-layer MLP, excess over adversarial semantics.}
  \label{fig:depth_sem:mlp1:adv}
\end{subfigure}
\hspace{1em}
\begin{subfigure}{.3\linewidth}
  \centering
  \includegraphics[height=.6\linewidth]{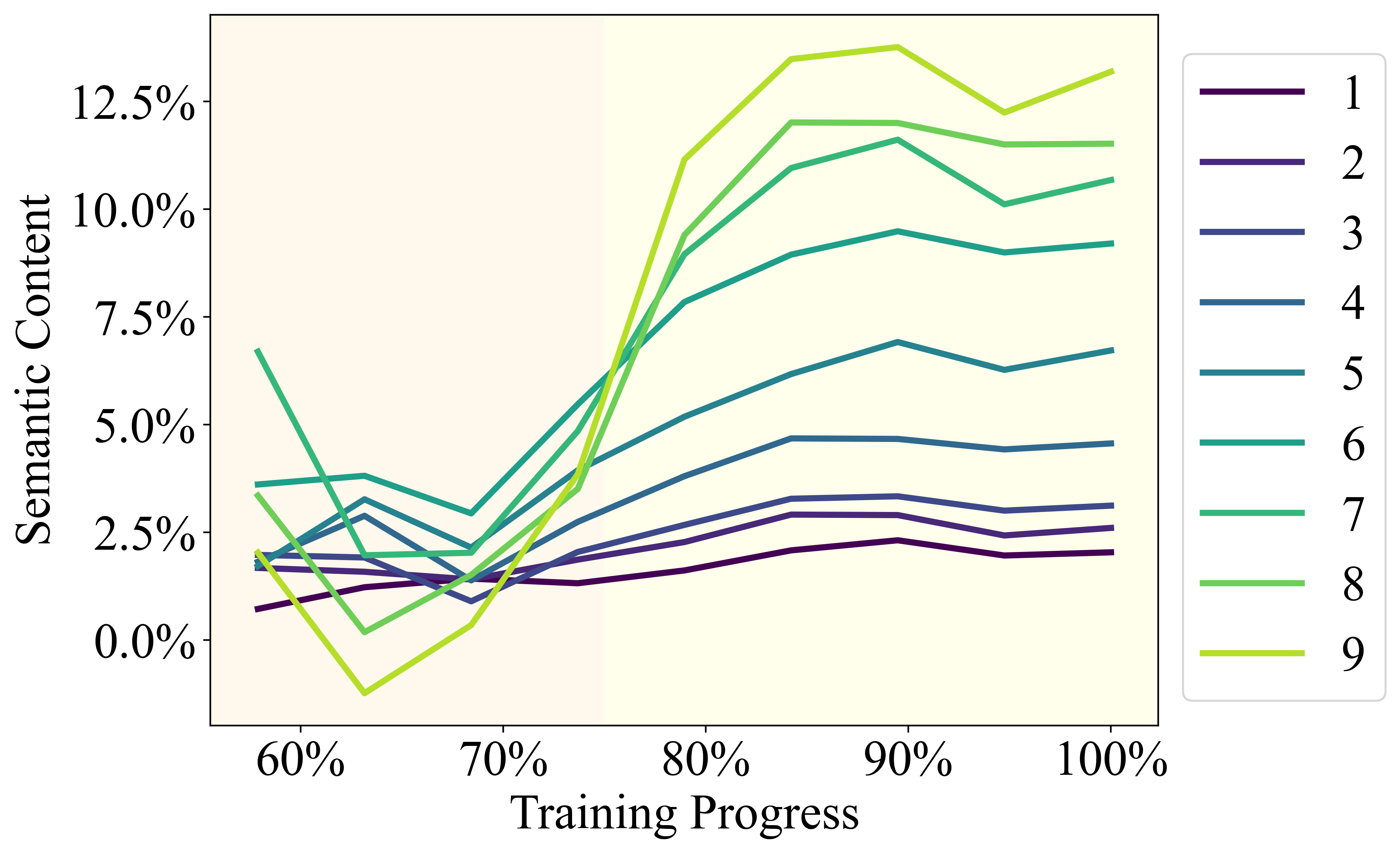}
  \caption{Probing with a 2-layer MLP, excess over adversarial semantics.}
  \label{fig:depth_sem:mlp2:adv}
\end{subfigure}

\caption{Plotting the semantic content separated by depth over time. Note that the excess over the two alternative semantics starts at zero during the middle phase of training, then becomes significantly positive around the final phase for all depths, including those corresponding to unseen programs.}
\label{fig:depth_sem}
\end{figure*}

In this section, we evaluate the \textbf{retrieval hypothesis}: the semantic content can be attributed entirely to the LM recalling previously seen training data. Similar to the syntactic record hypothesis, this hypothesis offers another potential explanation for the results in \Cref{sec:meaning} which is consistent with \MH/.
For instance, if we test the LM on (A) ``\texttt{turnRight}, \texttt{move}, \texttt{turnLeft}'' and the LM was trained on (B) ``\texttt{turnRight}, \texttt{move}'', then the LM may have seen the second entry in the trace of 
(A) as the output of (B).

We design our experimental setting specifically to test this hypothesis: as the training corpus contains only programs of length 6 or greater, while the test set contains programs of length 1 to 10, it is impossible for the LM to ``retrieve'' program states corresponding to the first 5 entries in any trace. Hence, we argue that any representations of unseen program states must be \emph{inferred} by the LM according to the semantics of the programming language.

\Cref{table:depth_main} displays the accuracy of the probes at the end of training; as the probe achieves non-trivial accuracy when tracing \emph{unseen} programs (of length 5 or less), we argue that the observed semantic content cannot be fully attributed to a retrieval-like process, and instead requires the LM to perform some degree of generalization over the semantics. We also remark that \Cref{appendix:reference_semantics} offers another piece of evidence in support of this claim, as the semantic content is lower when using the reference programs (whereas we would expect the ability of the LM to perform retrieval to \emph{increase} when following the training distribution more closely).

We make 2 additional observations. First, the effect of retrieval differs by feature, and appears more pronounced the ``simpler'' the feature is to compute. For instance, with the direction feature, there is a local optimum in the accuracy of the linear and 1-layer MLP probes at depth 6, which is indicative that some amount of retrieval is occurring in the representations of the LM: as the direction of the robot depends only on the initial direction and program, the LM has likely seen the answer previously in its training data. Conversely, the feature corresponding to the position of the robot decreases monotonically with depth for all 3 probes (with a very small exception at depth 6 for the linear probe), which is consistent with execution (deeper program states are harder to trace without a mistake due to compounding errors). Second, deeper probes exhibit lower sensitivity to retrieval across all 3 features. As deeper probes can extract more complex representations, this suggests that (1) the results of retrieval are represented more shallowly in the LM states, while (2) benefits of the shallow retrieval representations are less important once the probe is able to extract the deeper semantic representations.

\Cref{fig:depth_sem} also plots, for all 3 probes and separated by depth across all training steps, (1) the semantic content for the original semantics and (2) the excess semantic content with respect to the flip and adversarial semantics. The results show that the excess over the flip and adversarial semantics begins around zero in the middle phase (for all depths), then becomes positive by the end of training. This behavior is consistent with the results in the main text that suggest that the LM learns semantics over the latter half of training. We also point out the excess semantic content for the flip semantics as shown in \Cref{fig:depth_sem:linear:flip,fig:depth_sem:mlp1:flip,fig:depth_sem:mlp2:flip} increase as the depth increases from 1 to 5 (starting from close to 0 at depth 1), which we attribute to the high degree of correlation between the flip and original semantics (which degrades as the program length increases); beyond depth 5, the difference actually decreases, whereas we would expect the difference to be greater if the probe is simply relying on representations of retrieved program states in the original semantics. To summarize, while we do find evidence that the LM is performing some retrieval process, our results indicate that this process coexists with semantics in the representations of the LM.

\FloatBarrier
\subsection{Interpreting programs without outputs}

We conduct another series of experiments for explore whether the LM is capable of interpreting programs when only inputs are provided in the specification, and the outputs in the specification are obscured. Specifically, for every reference program and specification in the original trace datasets, we generate an \textbf{input-only trace dataset} according to the following loop (cf. \Cref{eq:trace_dataset}):
\begin{align}
\label{eq:no_input_trace_dataset}
(\text{state}_{LM})_i = {LM}(\text{input}, \text{empty}, \{(\text{state}_{LM})_j\}_{j=1}^{i-1}) \\
\text{token}_i = \text{prog}_\text{ref}[i] \\
(\text{state}_{\text{prog}})_i = \text{exec}(\text{token}_i, (\text{state}_{\text{prog}})_{i-1})
\end{align}
where $\text{prog}_\text{ref}$ is the reference program and $\text{prog}_\text{ref}[i]$ is the $i^{th}$ program token in the reference program; $\text{input} = (\text{state}_{\text{prog}})_0$ is the input state from the specification; and we replace every output in the specification with an \emph{empty} Karel grid. In other words, we force the LM to generate the reference program given \emph{only inputs and no outputs}. 

We emphasize that this task is firmly outside of the distribution of the training data, and there is no guarantee that the LM states will even be coherent. Hence, any ability to interpret programs in this setting could be considered \emph{emergent} behavior on the part of the LM.
Additionally, as the text still contains the necessary information for \emph{interpreting} the program (as the output is not necessary for this task), we also expect to find that probing for the alternative semantics yields accuracies that are no higher than the original semantics.

\begin{table*}[tbp]
\caption{The results of our input-only probing experiments. For each of the original, flip, and adversarial semantics, we report the semantic content (SC) at the end of training for 2 abstract states into the past (-2, -1), the current state (0), and 2 abstract states into the future (+1, +2), using linear, 1-layer MLP, and 2-layer MLP probes. We also regress the SC against the generative accuracy over the second half of training ($R^2 (p)$). For each of the alternative semantics, we additionally compute the difference with respect to the original semantics ($\Delta$) and regress the difference against the generative accuracy over the second half of training as ($R^2 (p)$ of $\Delta$). Highlighted cells are statistically significant at a level of $p<0.05$ with an $R^2$ of at least 50\%; all such correlations are positive.}
\label{table:obscured}
\begin{center}
\tabcolsep=0.15cm
\begin{tabular}{ c c | c c | c c | c c | c c | c c }
\toprule
 & & \multicolumn{2}{c|}{original} & \multicolumn{4}{c|}{flip} &  \multicolumn{4}{c}{adversarial} \\
 & & SC & $R^2 (p)$ & SC & \multicolumn{1}{c}{$R^2 (p)$} & $\Delta$ & $R^2 (p)$ of $\Delta$ & SC & \multicolumn{1}{c}{$R^2 (p)$} & $\Delta$ & $R^2 (p)$ of $\Delta$ \\
 \midrule

\multirow{5}{*}{\rotatebox[origin=c]{90}{linear}} & -2 & \cellcolor{yellow!25} 57.7 & \cellcolor{yellow!25} 89.6 (\textless.001) & \cellcolor{yellow!25} 57.5 & \cellcolor{yellow!25} 88.2 (\textless.001) & \cellcolor{yellow!25} 0.3 & \cellcolor{yellow!25} 55.8 (0.021) & \cellcolor{yellow!25} 50.5 & \cellcolor{yellow!25} 68.6 (0.006) & \cellcolor{yellow!25} 7.2 & \cellcolor{yellow!25} 92.9 (\textless.001) \\
 & -1 & \cellcolor{yellow!25} 60.5 & \cellcolor{yellow!25} 83.8 (\textless.001) & \cellcolor{yellow!25} 59.8 & \cellcolor{yellow!25} 80.4 (0.001) & \cellcolor{yellow!25} 0.7 & \cellcolor{yellow!25} 84.1 (\textless.001) & \cellcolor{yellow!25} 51.5 & \cellcolor{yellow!25} 55.5 (0.021) & \cellcolor{yellow!25} 9.1 & \cellcolor{yellow!25} 87.4 (\textless.001) \\
 & 0 & \cellcolor{yellow!25} 54.6 & \cellcolor{yellow!25} 85.9 (\textless.001) & \cellcolor{yellow!25} 54.2 & \cellcolor{yellow!25} 82.7 (\textless.001) & \cellcolor{yellow!25} 0.4 & \cellcolor{yellow!25} 92.9 (\textless.001) & 47.6 & 6.3 (0.514) & \cellcolor{yellow!25} 7.0 & \cellcolor{yellow!25} 86.7 (\textless.001) \\
 & 1 & \cellcolor{yellow!25} 52.4 & \cellcolor{yellow!25} 81.1 (\textless.001) & \cellcolor{yellow!25} 52.2 & \cellcolor{yellow!25} 80.7 (0.001) & \cellcolor{yellow!25} 0.2 & \cellcolor{yellow!25} 75.7 (0.002) & \cellcolor{yellow!25} 47.5 & \cellcolor{yellow!25} 79.0 (0.001) & \cellcolor{yellow!25} 4.9 & \cellcolor{yellow!25} 84.8 (\textless.001) \\
 & 2 & \cellcolor{yellow!25} 51.7 & \cellcolor{yellow!25} 70.3 (0.005) & \cellcolor{yellow!25} 51.6 & \cellcolor{yellow!25} 69.1 (0.006) & \cellcolor{yellow!25} 0.1 & \cellcolor{yellow!25} 62.1 (0.012) & 47.6 & 12.0 (0.362) & \cellcolor{yellow!25} 4.1 & \cellcolor{yellow!25} 74.8 (0.003) \\
\midrule
\multirow{5}{*}{\rotatebox[origin=c]{90}{MLP-1}} & -2 & \cellcolor{yellow!25} 77.3 & \cellcolor{yellow!25} 81.2 (\textless.001) & \cellcolor{yellow!25} 77.2 & \cellcolor{yellow!25} 81.8 (\textless.001) & 0.0 & 0.8 (0.824) & \cellcolor{yellow!25} 62.6 & \cellcolor{yellow!25} 54.7 (0.023) & \cellcolor{yellow!25} 14.7 & \cellcolor{yellow!25} 89.8 (\textless.001) \\
 & -1 & \cellcolor{yellow!25} 79.8 & \cellcolor{yellow!25} 81.7 (\textless.001) & \cellcolor{yellow!25} 79.5 & \cellcolor{yellow!25} 80.4 (0.001) & 0.3 & 38.3 (0.076) & \cellcolor{yellow!25} 64.7 & \cellcolor{yellow!25} 56.1 (0.020) & \cellcolor{yellow!25} 15.1 & \cellcolor{yellow!25} 88.4 (\textless.001) \\
 & 0 & \cellcolor{yellow!25} 62.3 & \cellcolor{yellow!25} 82.4 (\textless.001) & \cellcolor{yellow!25} 62.3 & \cellcolor{yellow!25} 81.4 (\textless.001) & 0.1 & 5.1 (0.558) & \cellcolor{yellow!25} 50.2 & \cellcolor{yellow!25} 70.6 (0.005) & \cellcolor{yellow!25} 12.2 & \cellcolor{yellow!25} 83.6 (\textless.001) \\
 & 1 & \cellcolor{yellow!25} 55.3 & \cellcolor{yellow!25} 80.8 (\textless.001) & \cellcolor{yellow!25} 55.3 & \cellcolor{yellow!25} 80.1 (0.001) & 0.0 & 22.8 (0.194) & 49.1 & 39.3 (0.071) & \cellcolor{yellow!25} 6.2 & \cellcolor{yellow!25} 85.7 (\textless.001) \\
 & 2 & \cellcolor{yellow!25} 52.9 & \cellcolor{yellow!25} 70.7 (0.005) & \cellcolor{yellow!25} 52.7 & \cellcolor{yellow!25} 73.0 (0.003) & 0.1 & 0.0 (0.974) & \cellcolor{yellow!25} 48.5 & \cellcolor{yellow!25} 55.7 (0.021) & \cellcolor{yellow!25} 4.4 & \cellcolor{yellow!25} 76.3 (0.002) \\
\midrule
\multirow{5}{*}{\rotatebox[origin=c]{90}{MLP-2}} & -2 & \cellcolor{yellow!25} 81.6 & \cellcolor{yellow!25} 68.5 (0.006) & \cellcolor{yellow!25} 81.7 & \cellcolor{yellow!25} 69.5 (0.005) & -0.1 & 2.1 (0.710) & 76.1 & 43.1 (0.055) & 5.5 & 48.3 (0.038) \\
 & -1 & \cellcolor{yellow!25} 82.5 & \cellcolor{yellow!25} 70.1 (0.005) & \cellcolor{yellow!25} 82.4 & \cellcolor{yellow!25} 70.5 (0.005) & 0.1 & 18.7 (0.245) & \cellcolor{yellow!25} 77.7 & \cellcolor{yellow!25} 52.3 (0.028) & \cellcolor{yellow!25} 4.8 & \cellcolor{yellow!25} 84.8 (\textless.001) \\
 & 0 & \cellcolor{yellow!25} 63.8 & \cellcolor{yellow!25} 67.6 (0.006) & \cellcolor{yellow!25} 63.7 & \cellcolor{yellow!25} 69.6 (0.005) & 0.2 & 28.3 (0.140) & \cellcolor{yellow!25} 51.3 & \cellcolor{yellow!25} 59.2 (0.015) & \cellcolor{yellow!25} 12.5 & \cellcolor{yellow!25} 69.0 (0.006) \\
 & 1 & \cellcolor{yellow!25} 56.7 & \cellcolor{yellow!25} 76.1 (0.002) & \cellcolor{yellow!25} 56.8 & \cellcolor{yellow!25} 69.1 (0.005) & -0.1 & 0.4 (0.872) & 49.3 & 47.8 (0.039) & \cellcolor{yellow!25} 7.4 & \cellcolor{yellow!25} 80.5 (0.001) \\
 & 2 & \cellcolor{yellow!25} 53.9 & \cellcolor{yellow!25} 59.4 (0.015) & \cellcolor{yellow!25} 53.7 & \cellcolor{yellow!25} 59.7 (0.015) & 0.2 & 4.6 (0.580) & 48.8 & 49.4 (0.035) & \cellcolor{yellow!25} 5.1 & \cellcolor{yellow!25} 60.8 (0.013) \\

 \bottomrule
\end{tabular}
\end{center}
\end{table*}

\Cref{table:obscured} displays the results. 
As expected, the semantic content for the original abstract states is significantly degraded (compared to \Cref{table:main}), especially when probing into the future, which we attribute mainly to obscuring the outputs.
We also see that there the performance of the 3 probes is highly compressed when probing into the future, i.e., deeper probes do not perform much better than shallow probes. This can be explained by the same reasoning as in \Cref{appendix:reference_semantics}, i.e., the present and future program states are subject to an inherent degree of randomness (recall that the probe's task for the ``current'' abstract state is to predict $(\text{state}_{\text{prog}})_i$ from $(\text{state}_{LM})_i$ in \Cref{eq:no_input_trace_dataset}).
Indeed, we note that the semantic content is maximized at 1 abstract state into the past across all settings.

Nonetheless, our results suggest the LM still maintains the ability to interpret programs, even in this challenging setting; moreover, the interventional baseline offers strong support that the representations of the LM are aligned with the original semantics (rather than being learned by the probe). In particular, we see statistically significant correlations for all three probes and all 5 abstract states when comparing the original semantic content against the adversarial semantic content. Note that, even though predicting future states is difficult, the ability to predict future states is correlated with knowing the current state, so that it is not surprisingly that the adversarial semantic content for future states would be lower, despite the lack of information about future states.

We also observe a statistically significant positive correlation when regressing the excess of the original semantic content over the flip semantic content over the second half of training, when using the linear probe, though this effect disappears as we move to deeper probes. We emphasize that the flip semantics presents a very strong baseline as the semantics can often be inferred directly from the result of the original semantics (i.e., by reflecting the robot across an axis).

\begin{table}[tbp]
\caption{The excess semantic content at the end of training when comparing the original and alternative semantics using a linear probe on the input-only trace datasets. We separate the semantic content by the depth of the program state and the 3 features in the abstract state. The LM only observes program states at depth 6 or greater in the training corpus. We display depths consisting of at least 1\% of the training set. A positive value indicates that the original semantic content is greater than the alternative semantic content. }
\label{table:depth_space}
\begin{center}
\begin{tabular}{ cc|cccc| cccc}
\toprule
 &\multirow{2}{*}{depth} & \multicolumn{4}{c|}{flip} & \multicolumn{4}{c}{adversarial} \\
 & & direction & position & obstacle & all & direction & position & obstacle & all \\
 \midrule

\multirow{5}{*}{\rotatebox[origin=c]{90}{unseen}} & 1 & -0.51 & 0.15 & 0.17 & -0.11 & 9.30 & 1.19 & 1.63 & 4.86 \\
 & 2 & 0.36 & 0.31 & 0.50 & 0.40 & 25.35 & 3.58 & 0.97 & 14.08 \\
 & 3 & 1.53 & 0.39 & 0.65 & 0.96 & 20.97 & 4.25 & -0.12 & 11.63 \\
 & 4 & 1.92 & -0.14 & 0.17 & 0.83 & 20.09 & 4.80 & -0.17 & 11.60 \\
 & 5 & 3.12 & 0.32 & -0.00 & 1.49 & 16.07 & 4.52 & -0.09 & 9.44 \\
\midrule
\multirow{4}{*}{\rotatebox[origin=c]{90}{seen}} & 6 & 2.33 & 0.55 & 0.00 & 1.25 & 12.66 & 5.78 & 0.40 & 8.34 \\
 & 7 & 2.01 & 0.39 & 0.71 & 1.22 & 9.37 & 4.30 & -0.04 & 6.17 \\
 & 8 & 1.46 & -0.17 & 0.15 & 0.64 & 5.96 & 3.47 & -0.33 & 4.17 \\
 & 9 & 1.04 & -0.04 & 0.50 & 0.57 & 4.32 & 4.06 & 2.64 & 4.26 \\

 \bottomrule
\end{tabular}
\end{center}
\end{table}

\begin{figure*}[tbp]
\centering
\begin{subfigure}{.3\linewidth}
  \centering
  \includegraphics[height=.6\linewidth]{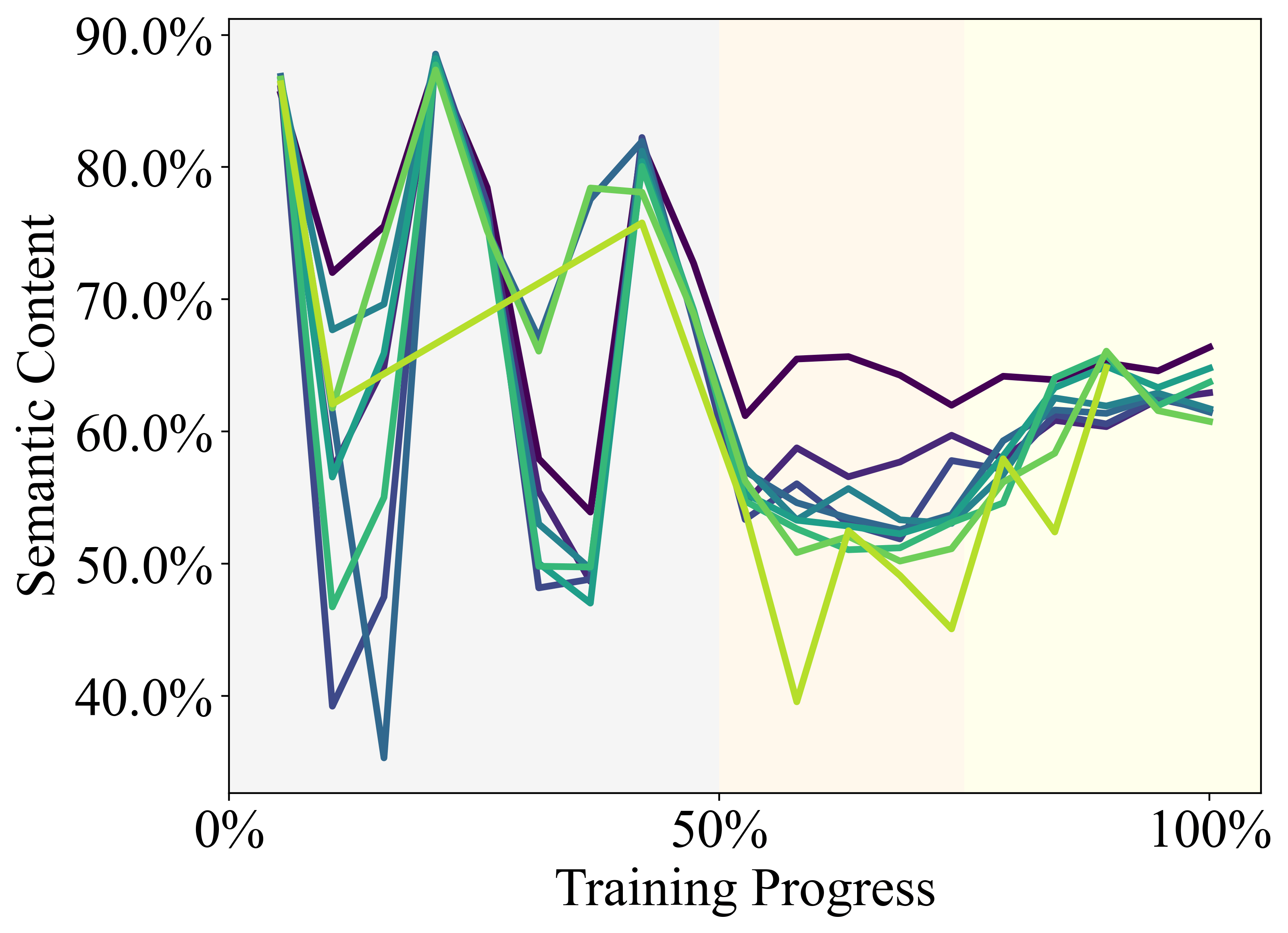}
  \caption{Probing with a linear classifier.}
  \label{fig:depth_sem_space:linear}
\end{subfigure}
\hspace{1em}
\begin{subfigure}{.3\linewidth}
  \centering
  \includegraphics[height=.6\linewidth]{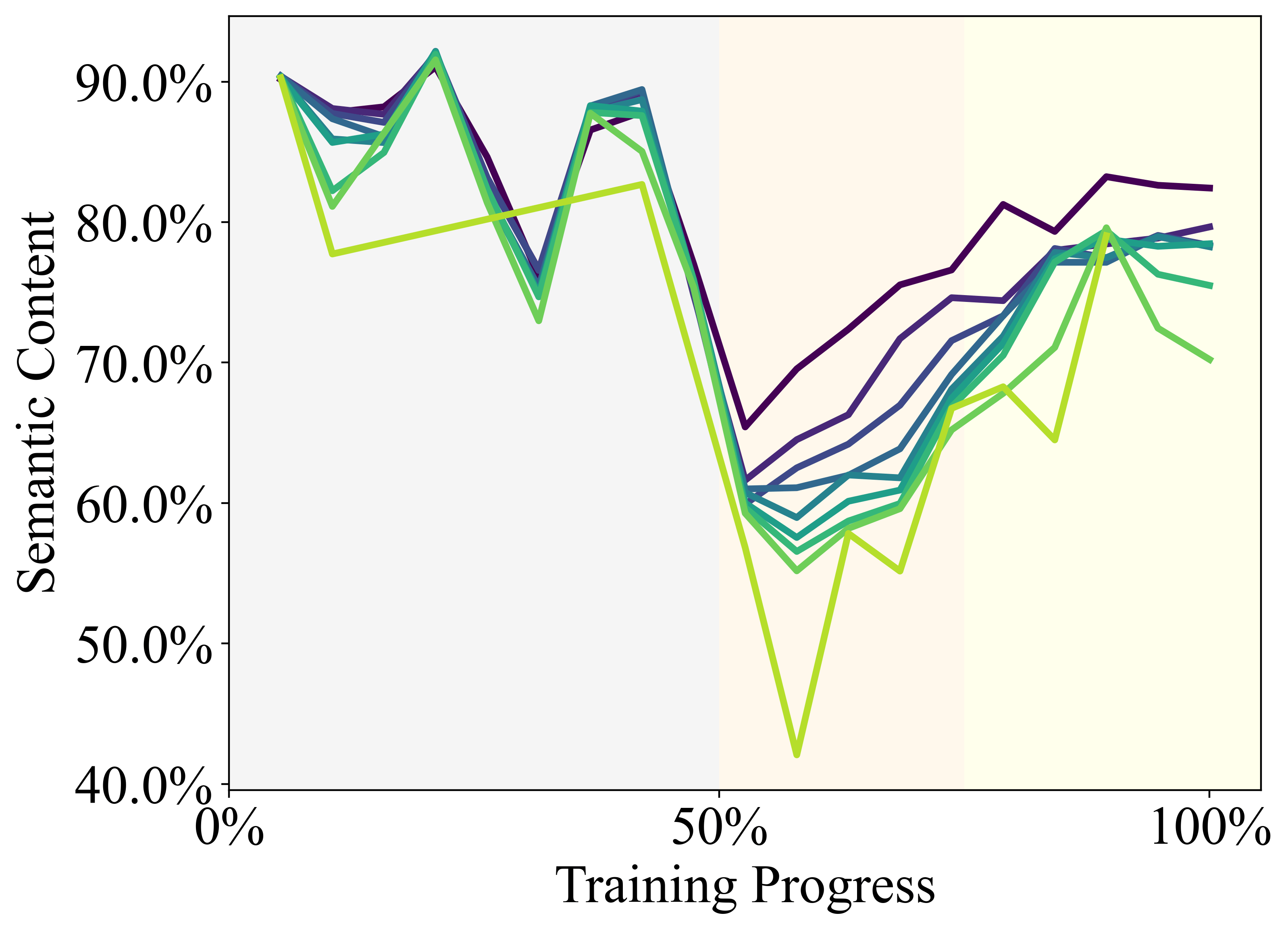}
  \caption{Probing with a 1-layer MLP.}
  \label{fig:depth_sem_space:mlp1}
\end{subfigure}
\hspace{1em}
\begin{subfigure}{.3\linewidth}
  \centering
  \includegraphics[height=.6\linewidth]{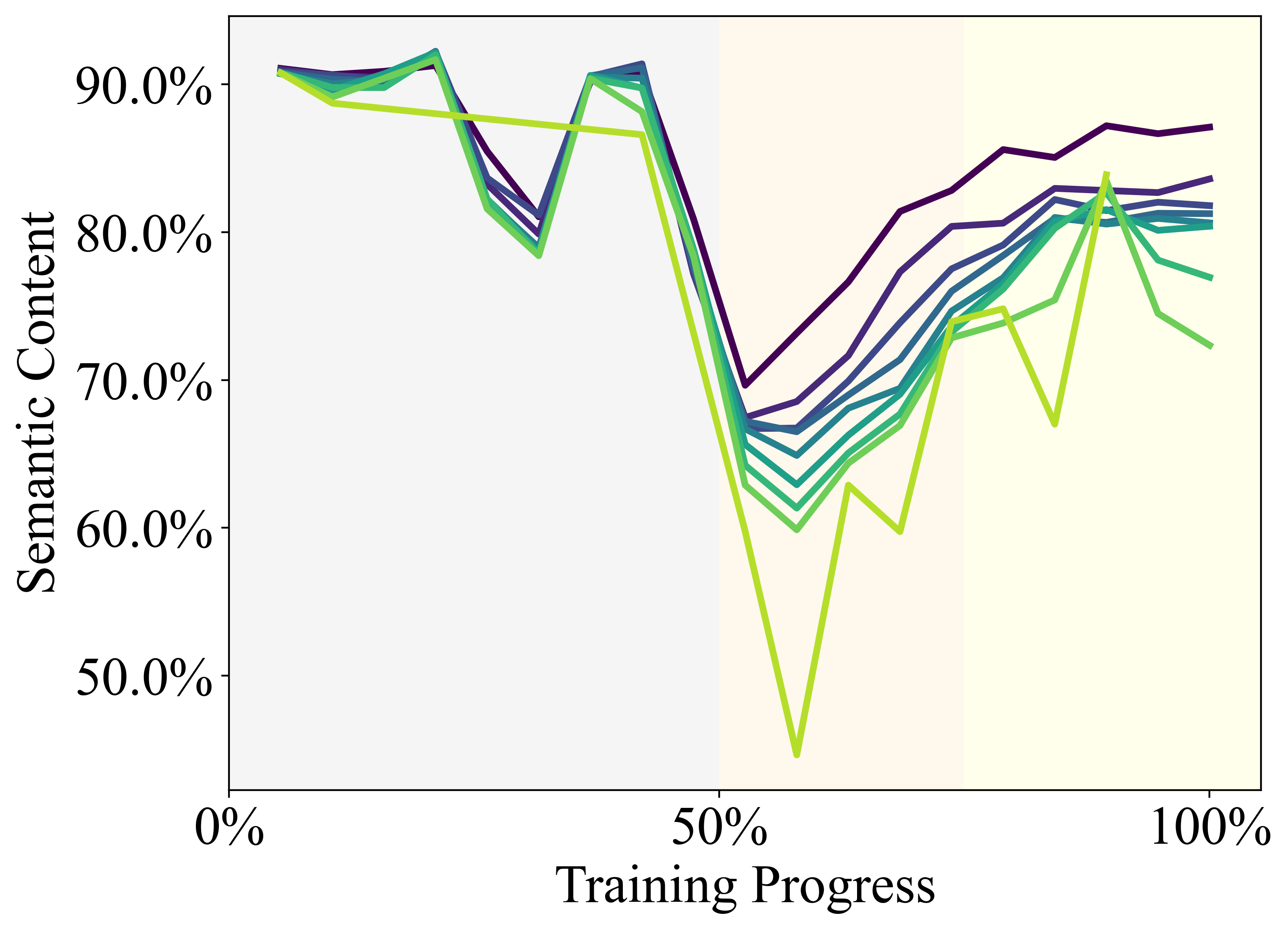}
  \caption{Probing with a 2-layer MLP.}
  \label{fig:depth_sem_space:mlp2}
\end{subfigure}
\\ \vspace{1em}

\begin{subfigure}{.3\linewidth}
  \centering
  \includegraphics[height=.6\linewidth]{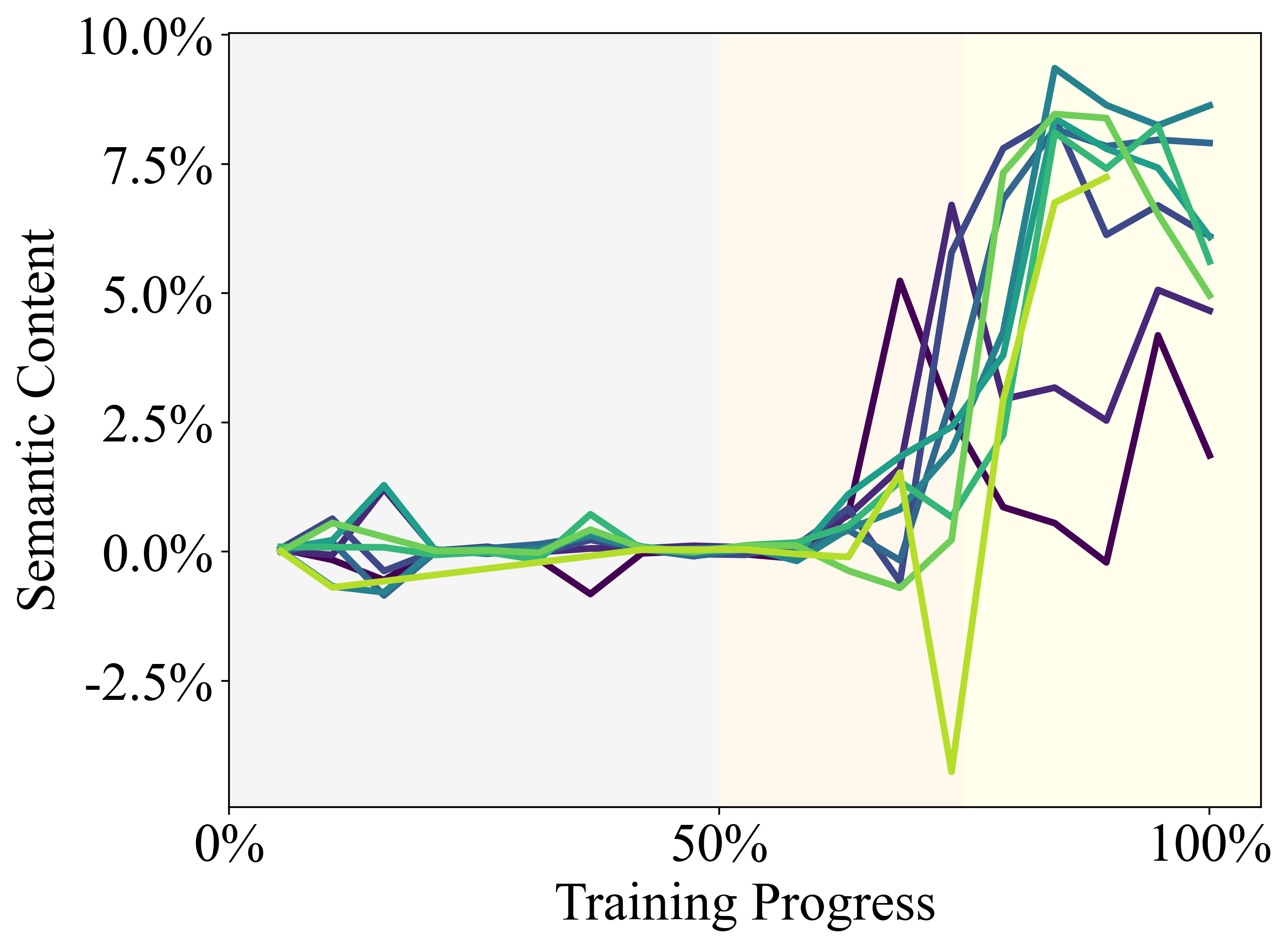}
  \caption{Probing with a linear classifier, excess over flip semantics.}
  \label{fig:depth_sem_space:linear:flip}
\end{subfigure}
\hspace{1em}
\begin{subfigure}{.3\linewidth}
  \centering
  \includegraphics[height=.6\linewidth]{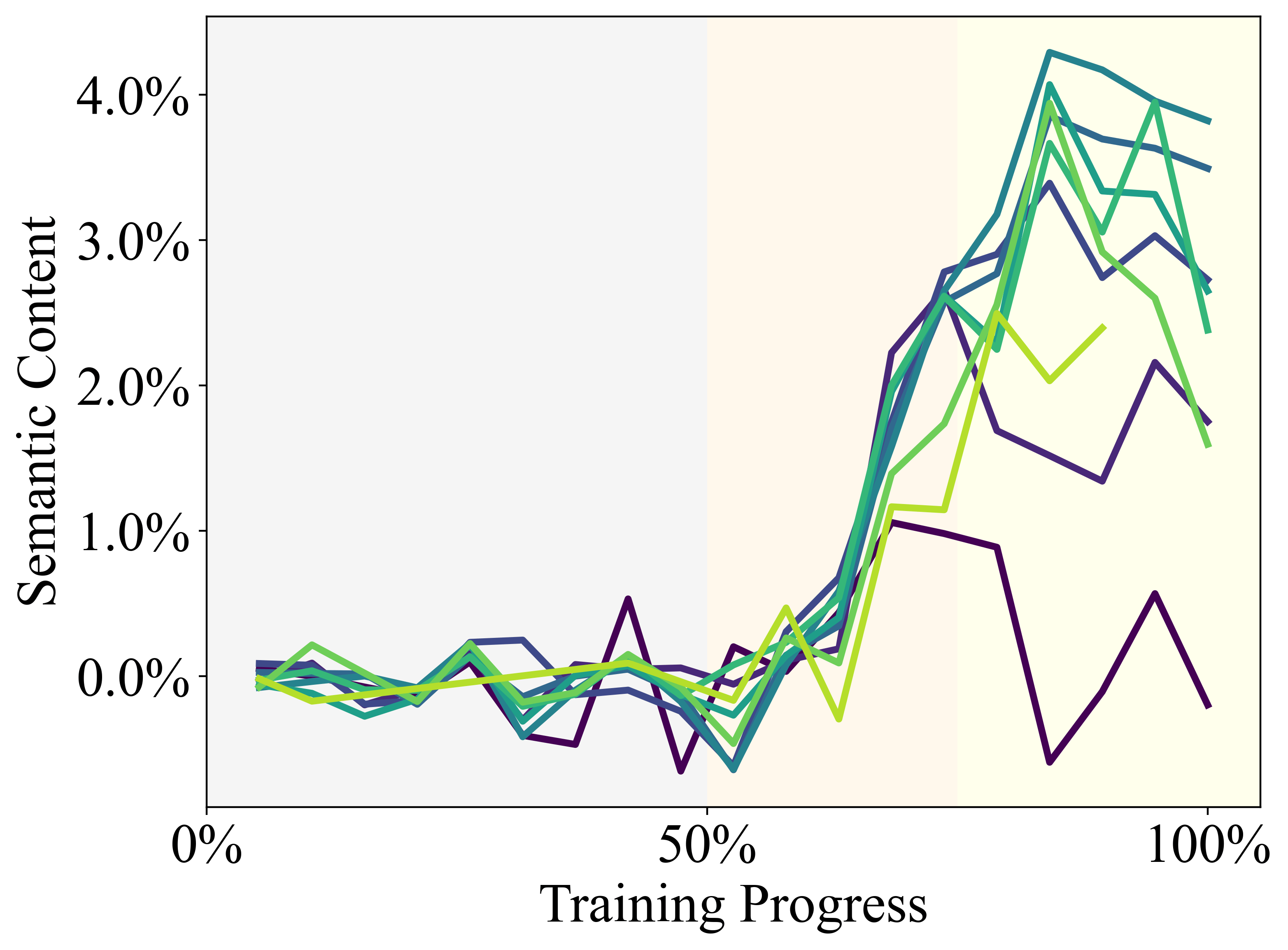}
  \caption{Probing with a 1-layer MLP, excess over flip semantics.}
  \label{fig:depth_sem_space:mlp1:flip}
\end{subfigure}
\hspace{1em}
\begin{subfigure}{.3\linewidth}
  \centering
  \includegraphics[height=.6\linewidth]{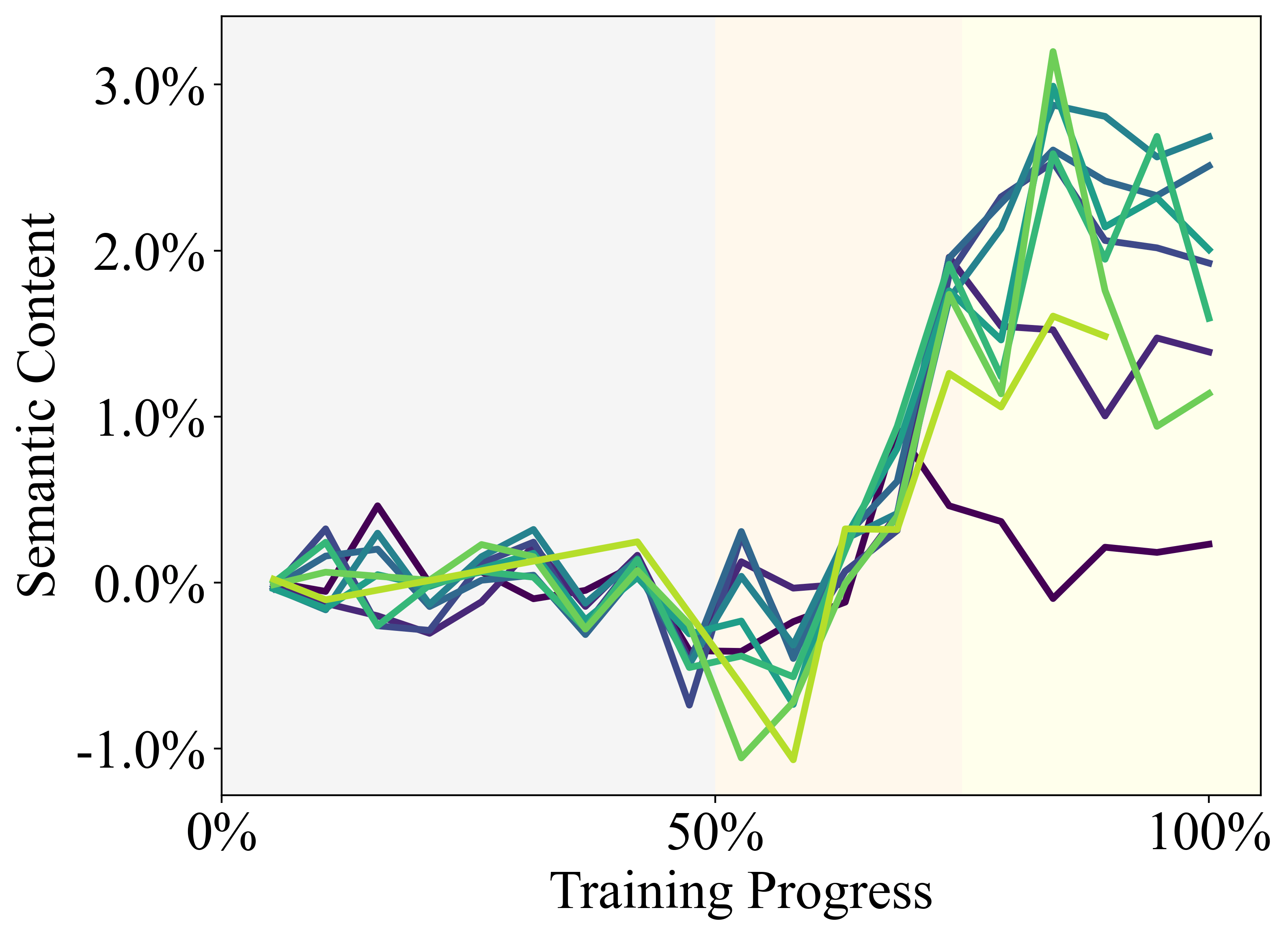}
  \caption{Probing with a 2-layer MLP, excess over flip semantics.}
  \label{fig:depth_sem_space:mlp2:flip}
\end{subfigure}
\\ \vspace{1em}

\begin{subfigure}{.3\linewidth}
  \centering
  \includegraphics[height=.6\linewidth]{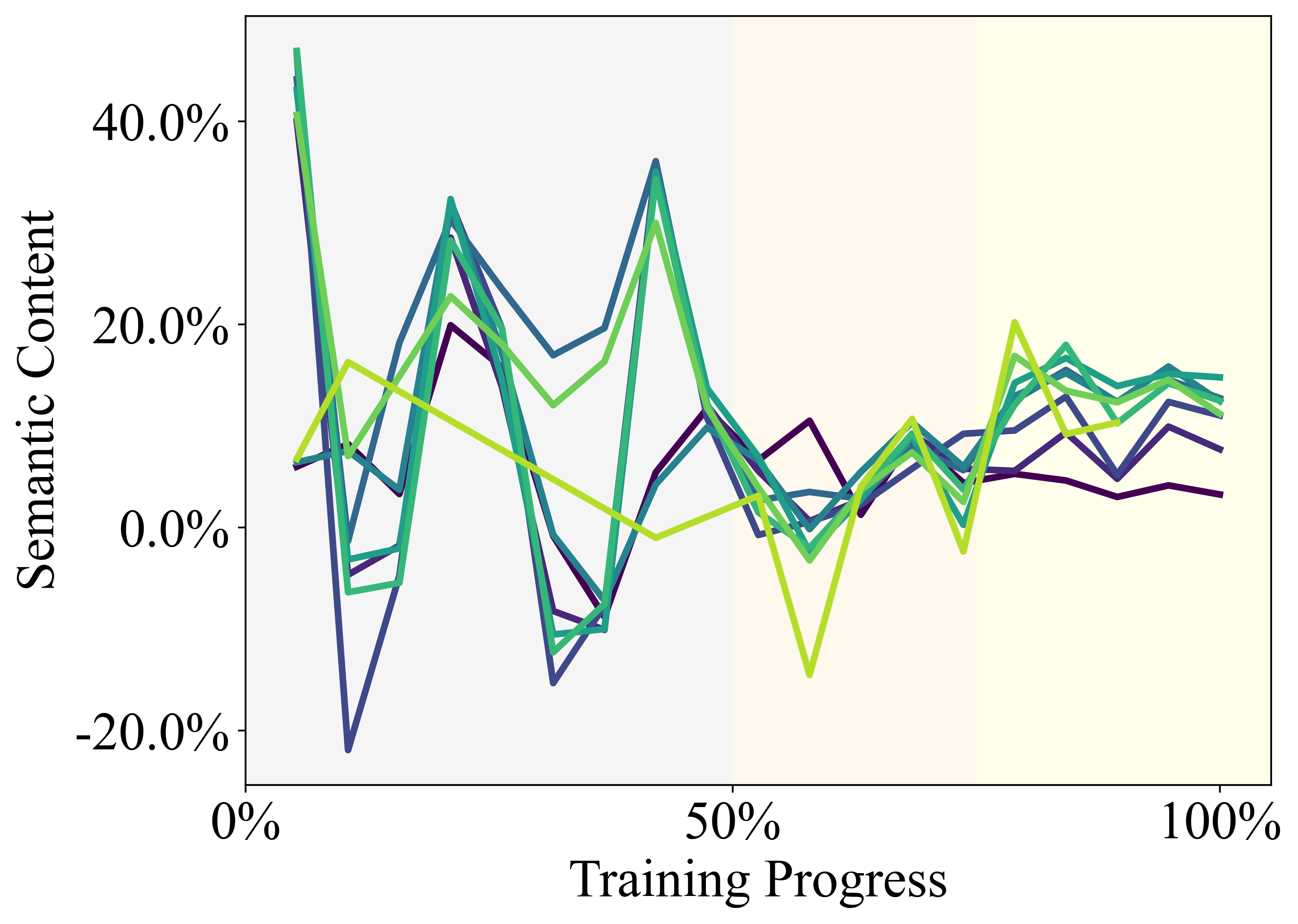}
  \caption{Probing with linear classifier, excess over adversarial semantics.}
  \label{fig:depth_sem_space:linear:adv}
\end{subfigure}
\hspace{1em}
\begin{subfigure}{.3\linewidth}
  \centering
  \includegraphics[height=.6\linewidth]{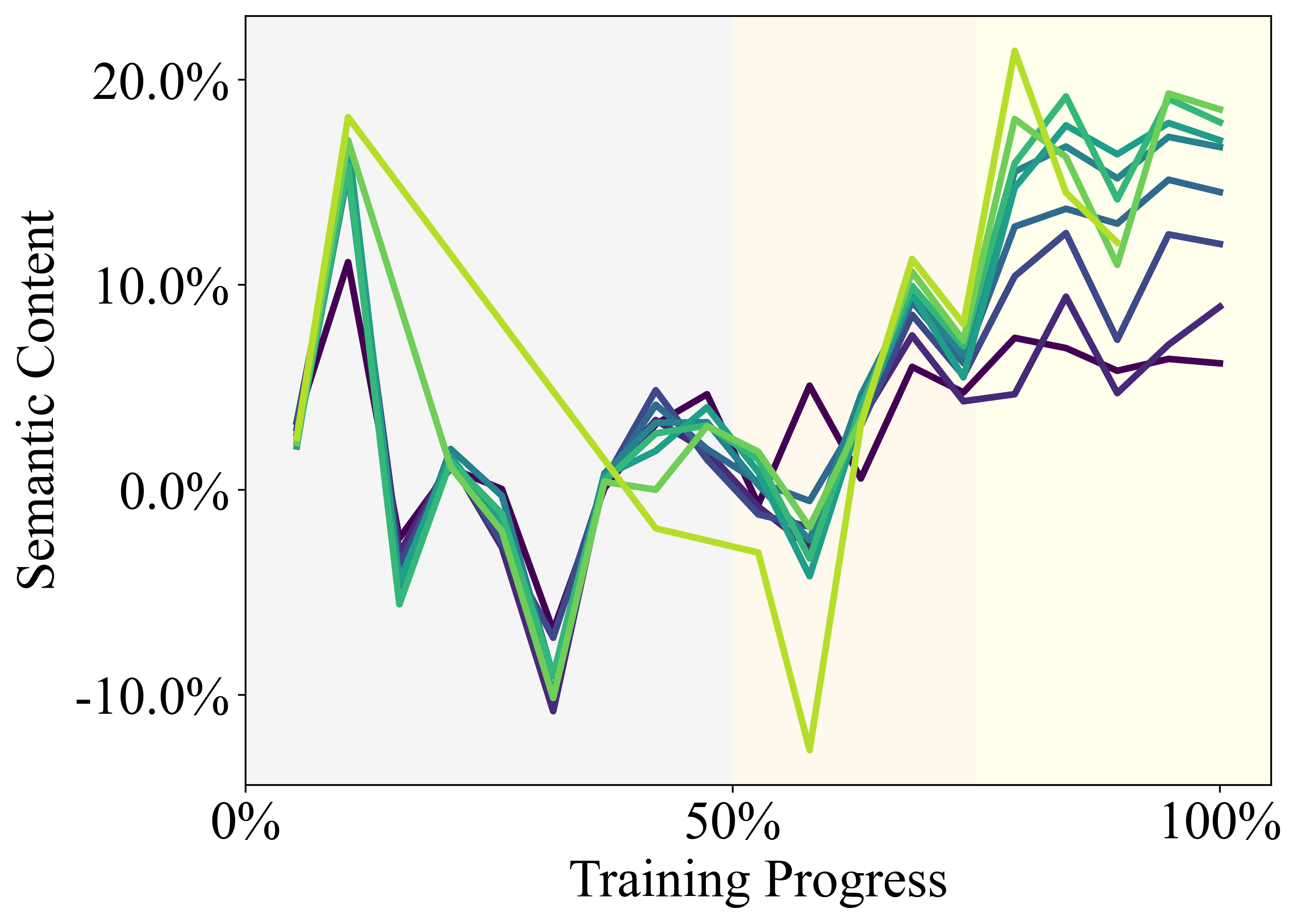}
  \caption{Probing with a 1-layer MLP, excess over adversarial semantics.}
  \label{fig:depth_sem_space:mlp1:adv}
\end{subfigure}
\hspace{1em}
\begin{subfigure}{.3\linewidth}
  \centering
  \includegraphics[height=.6\linewidth]{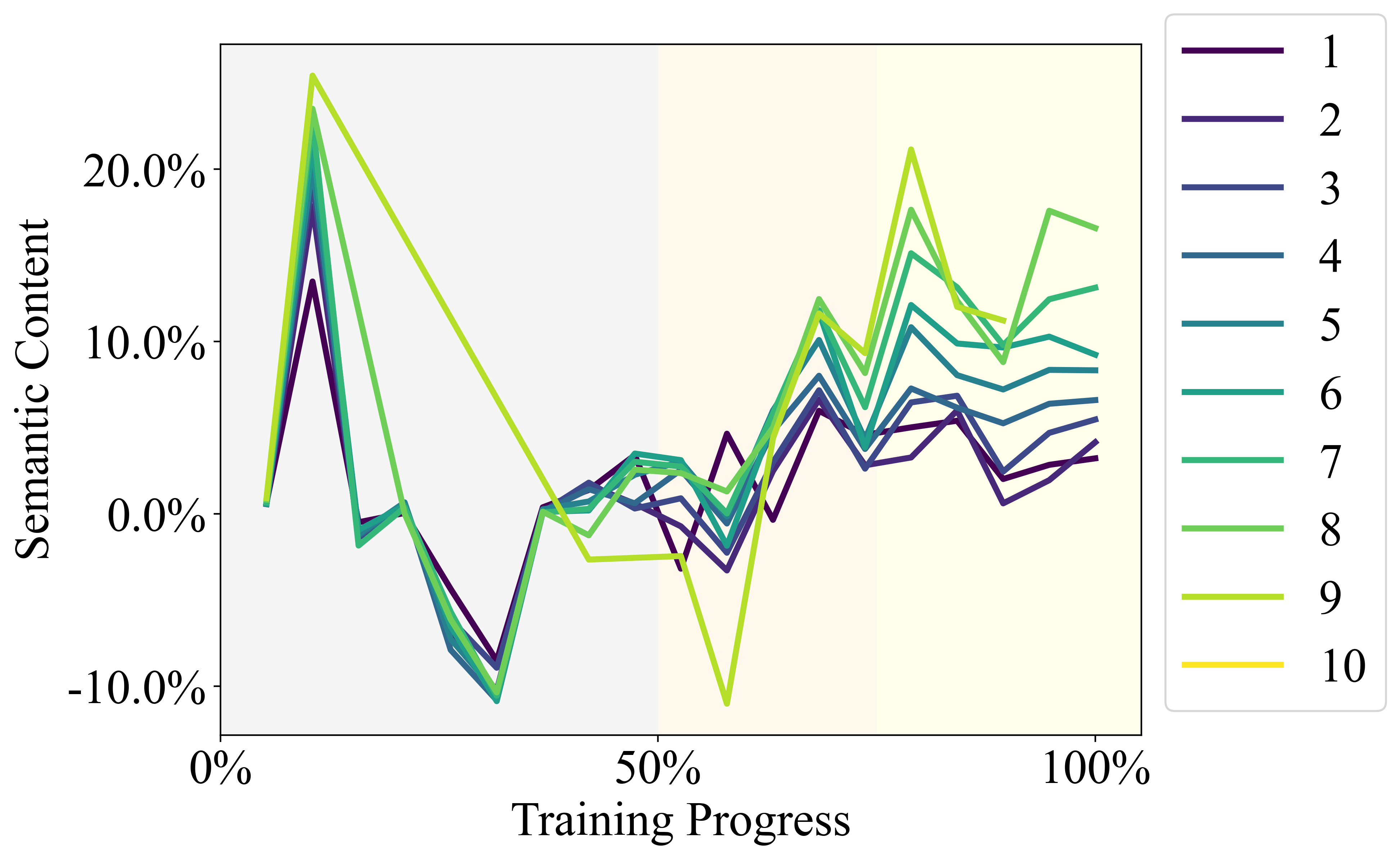}
  \caption{Probing with a 2-layer MLP, excess over adversarial semantics.}
  \label{fig:depth_sem_space:mlp2:adv}
\end{subfigure}
\caption{The semantic content on the input-only trace datasets, separated by depth over the full LM training run. Note that all plots show trends which, when aggregated over depth, exhibit a statistically significant linear correlation with the generate accuracy, except for \Cref{fig:depth_sem_space:mlp1:flip,fig:depth_sem_space:mlp2:flip}.}
\label{fig:depth_sem_space}
\end{figure*}

Could these results could be due to retrieval? To test this hypothesis, we also reproduce the experiments from \Cref{appendix:retrieval} on the input-only trace dataset. We focus on 1 abstract state into the past, as it is where the semantic content is maximized across all settings; and the linear probe, as the original semantic content and excess over the alternative semantics are all correlated with the generative accuracy to a statistically significant degree over the course of training.
\Cref{table:depth_space} displays the excess semantic content of the original over the flip and adversarial semantics, respectively, using a linear probe on the input-only trace datasets when probing for 1 abstract state into the past. We see that almost every semantic content of \emph{unseen} features (and overall abstract state) is greater when probing for the original semantics compared to the alternative semantics, confirming that the observed difference in semantic content cannot be entirely attributed to the LM performing retrieval. We thus conclude that the LM is able to abstractly interpret programs even without seeing the final outputs of the program, which is emergent behavior on out-of-distribution text. 
For completeness, \Cref{fig:depth_sem_space} plots results over the entire LM training.

\subsection{Selected regression and residual plots}

\begin{figure*}[tbp]
\centering
\begin{subfigure}{.48\linewidth}
  \centering
  \includegraphics[scale=0.3]{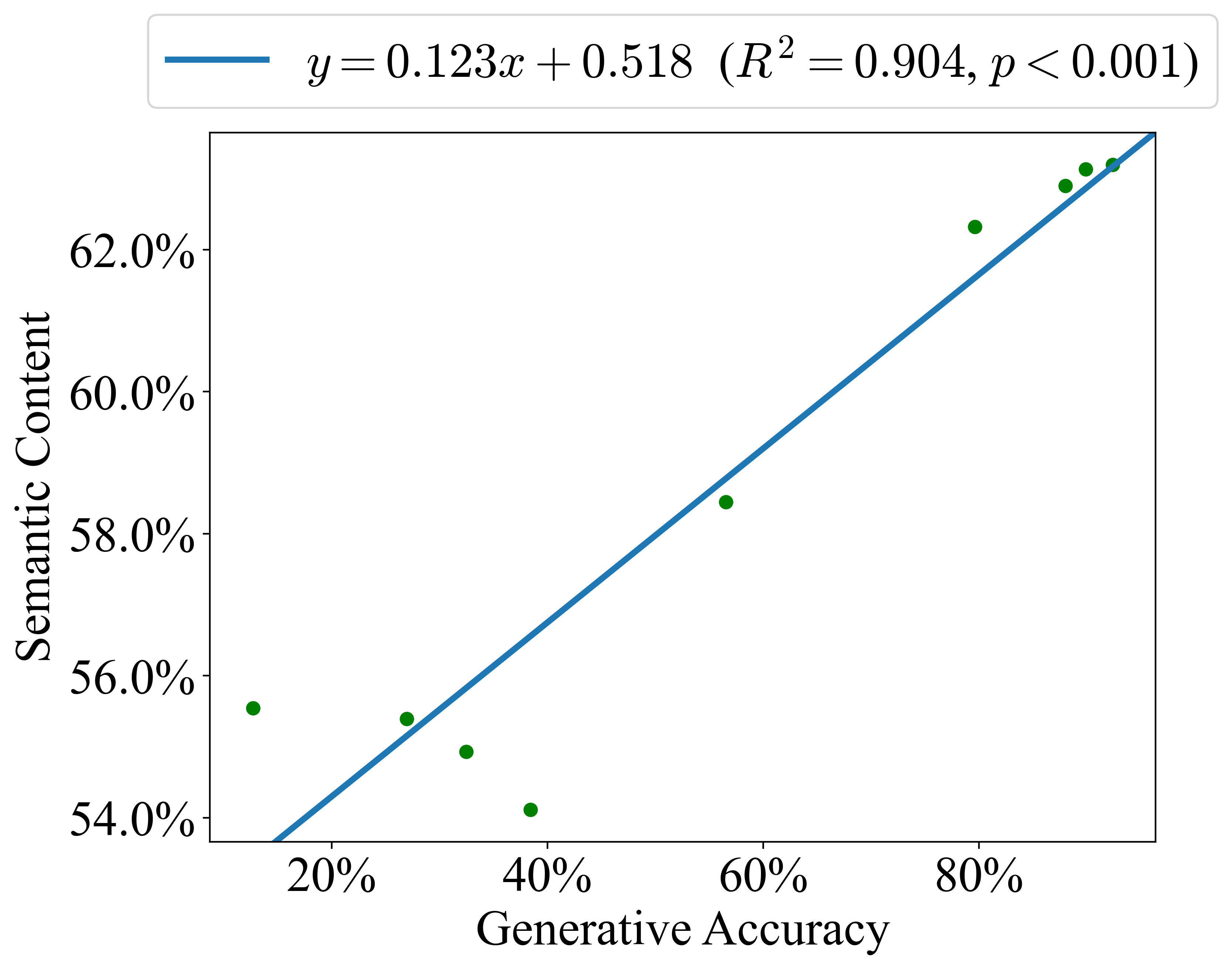}
  \caption{Regressing semantic content vs. generative accuracy.}
  \label{fig:regression:reg}
\end{subfigure}
\hspace{1em}
\begin{subfigure}{.48\linewidth}
  \centering
  \includegraphics[scale=0.3]{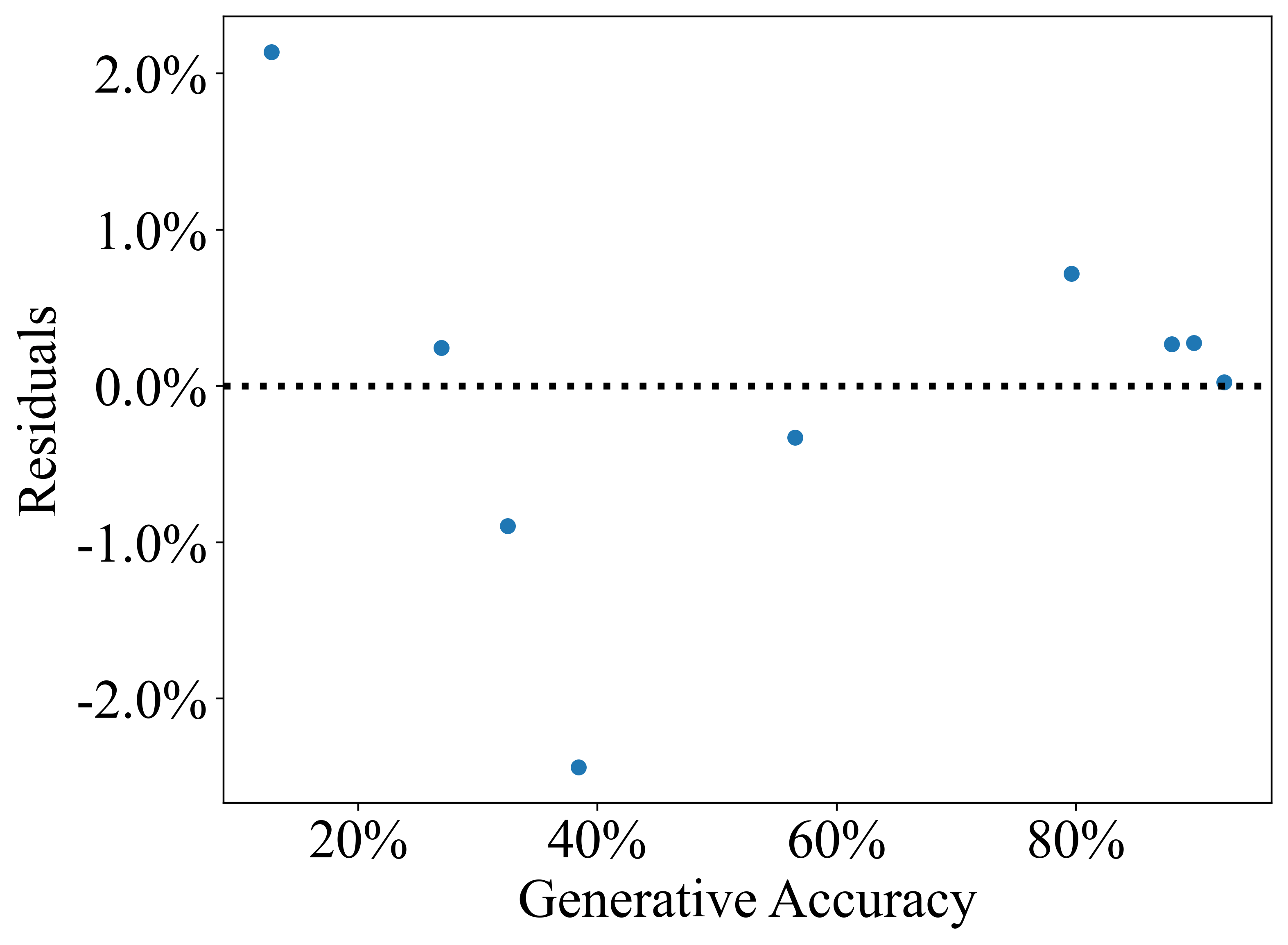}
  \caption{Residuals for semantic content vs. generative accuracy.}
  \label{fig:regression:res}
\end{subfigure} \\ \vspace{1em}

\begin{subfigure}{.48\linewidth}
  \centering
  \includegraphics[scale=0.3]{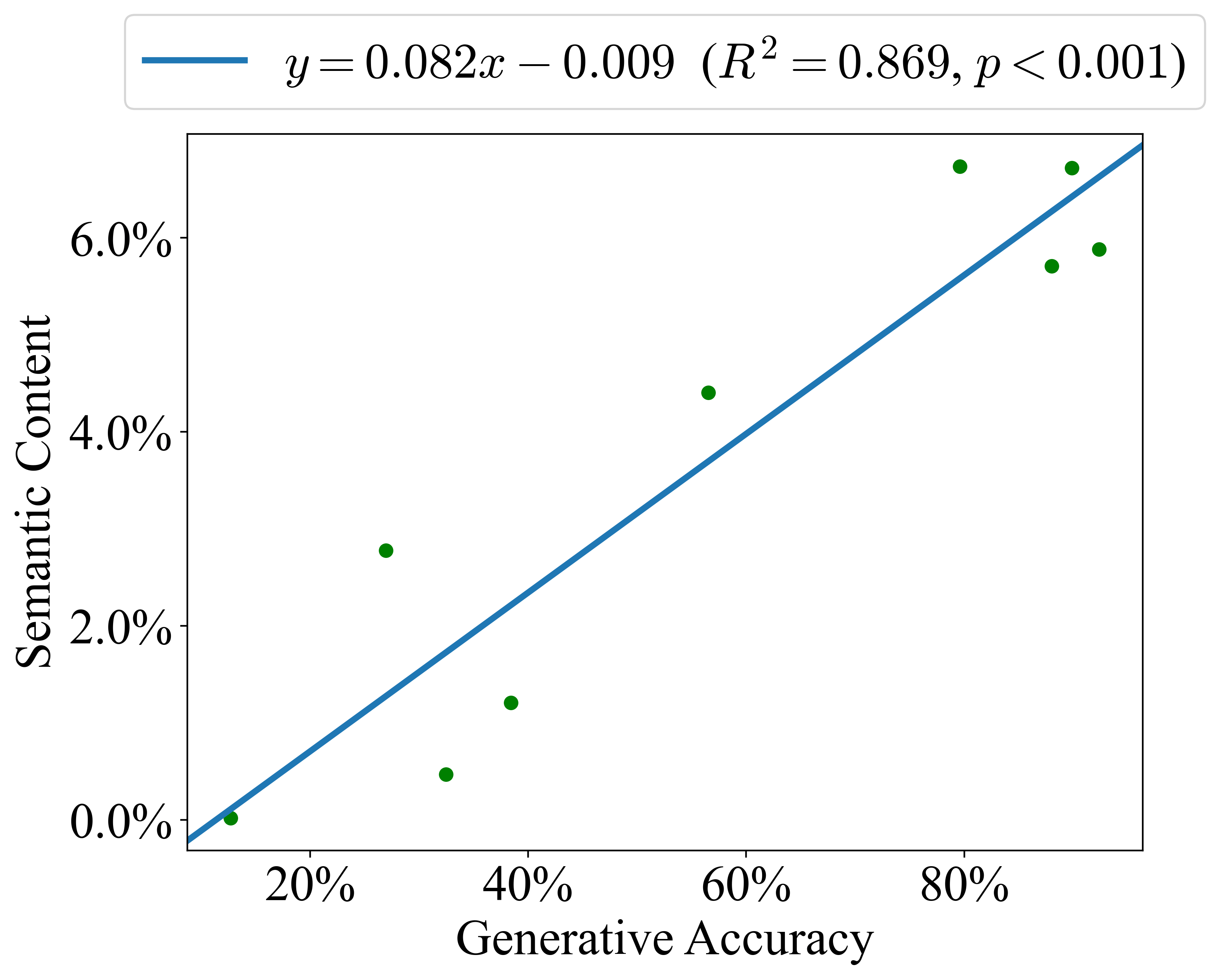}
  \caption{Regressing excess of original over flip semantic content vs. generative accuracy.}
  \label{fig:regression:flip:reg}
\end{subfigure}
\hspace{1em}
\begin{subfigure}{.48\linewidth}
  \centering
  \includegraphics[scale=0.3]{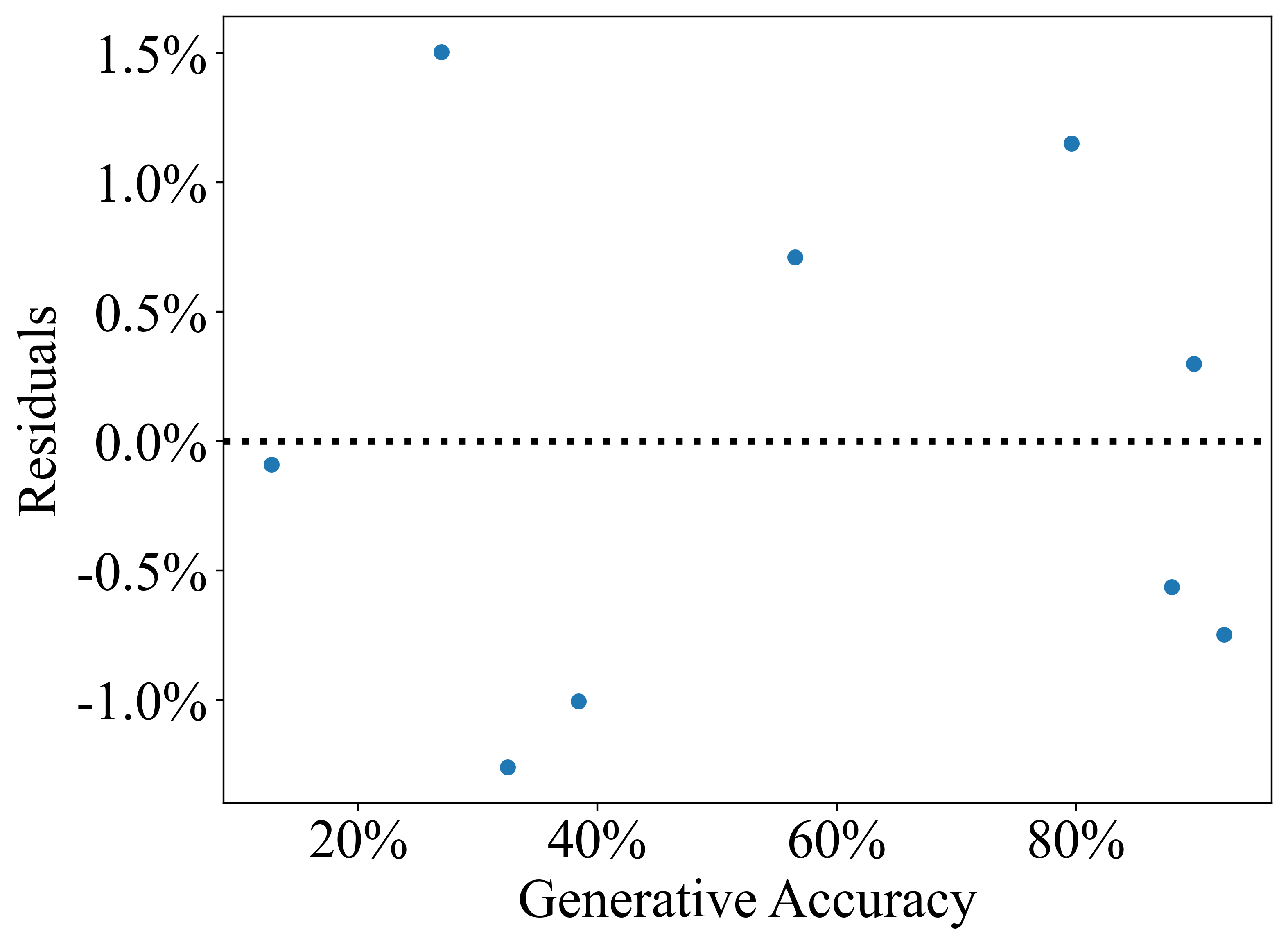}
  \caption{Residuals for excess of original over flip semantic content vs. generative accuracy.}
  \label{fig:regression:flip:res}
\end{subfigure} \\ \vspace{1em}

\begin{subfigure}{.48\linewidth}
  \centering
  \includegraphics[scale=0.3]{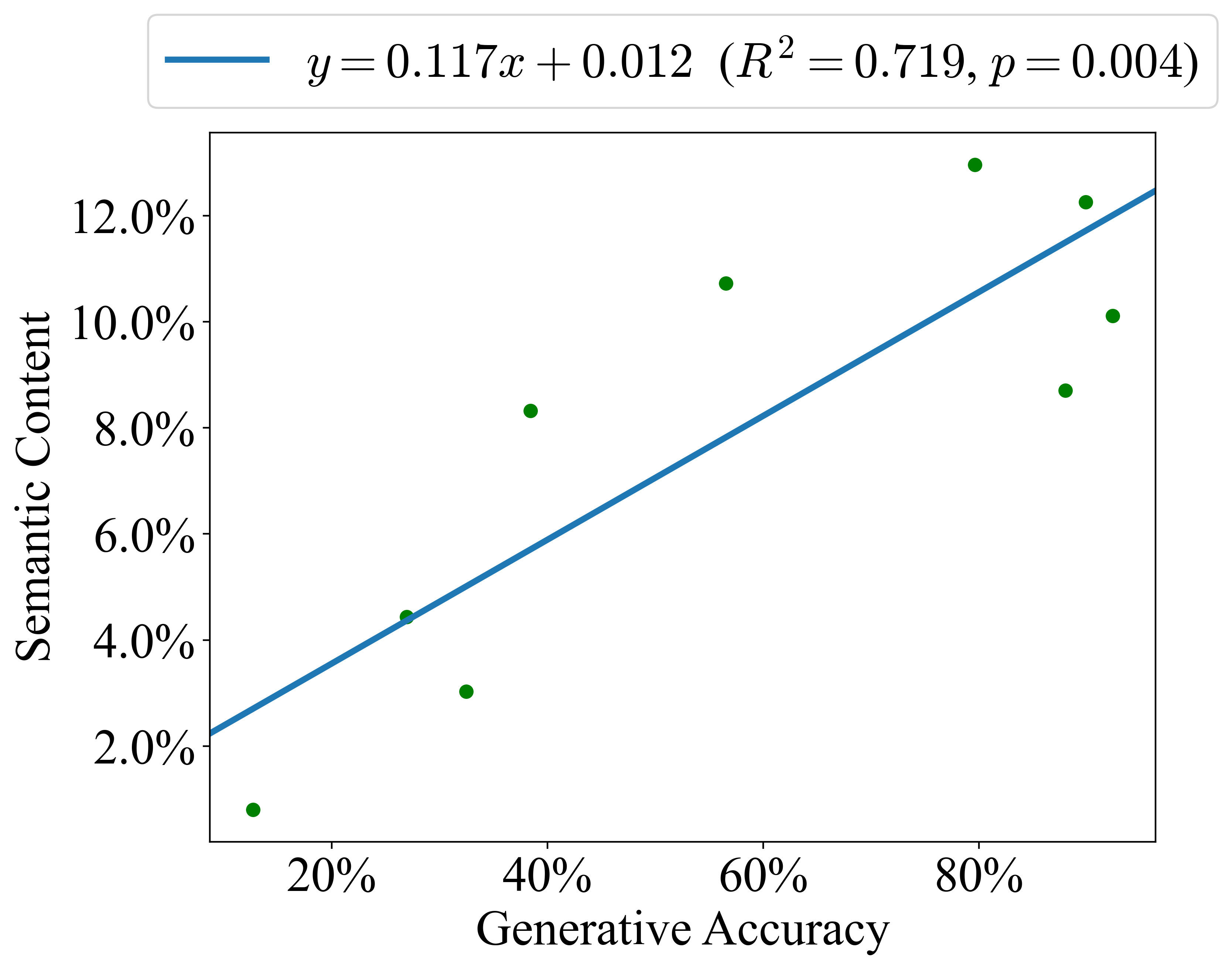}
  \caption{Regressing excess of original over adversarial semantic content vs. generative accuracy.}
  \label{fig:regression:adv:reg}
\end{subfigure}
\hspace{1em}
\begin{subfigure}{.48\linewidth}
  \centering
  \includegraphics[scale=0.3]{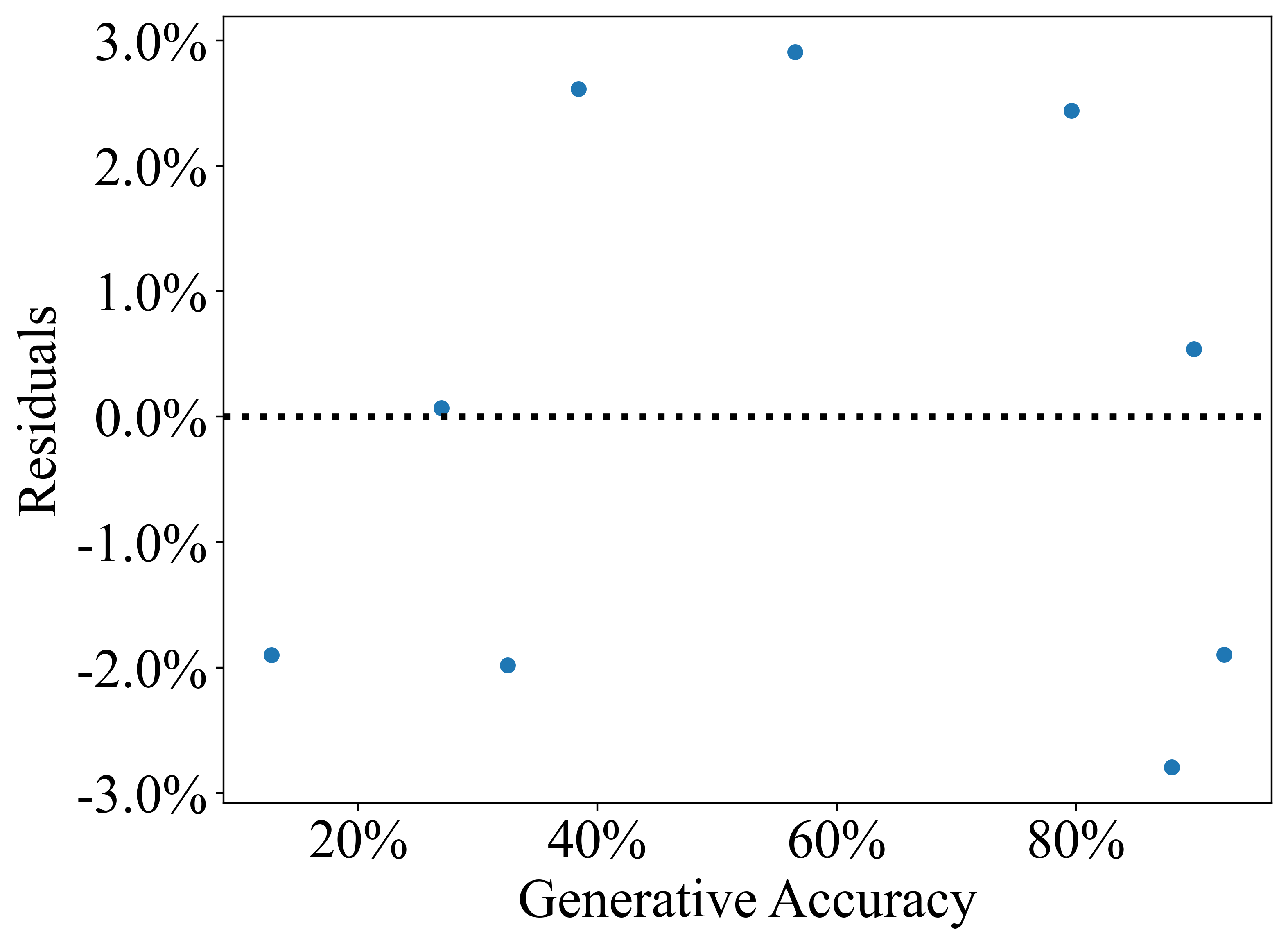}
  \caption{Residuals for excess of original over adversarial semantic content vs. generative accuracy.}
  \label{fig:regression:adv:res}
\end{subfigure}
\caption{Regression and residual plots for semantic content (measured by a linear classifier) vs. generative accuracy over the second half of training.}
\label{fig:regression}
\end{figure*}

\begin{figure*}[tbp]
\centering
\begin{subfigure}{.48\linewidth}
  \centering
  \includegraphics[scale=0.3]{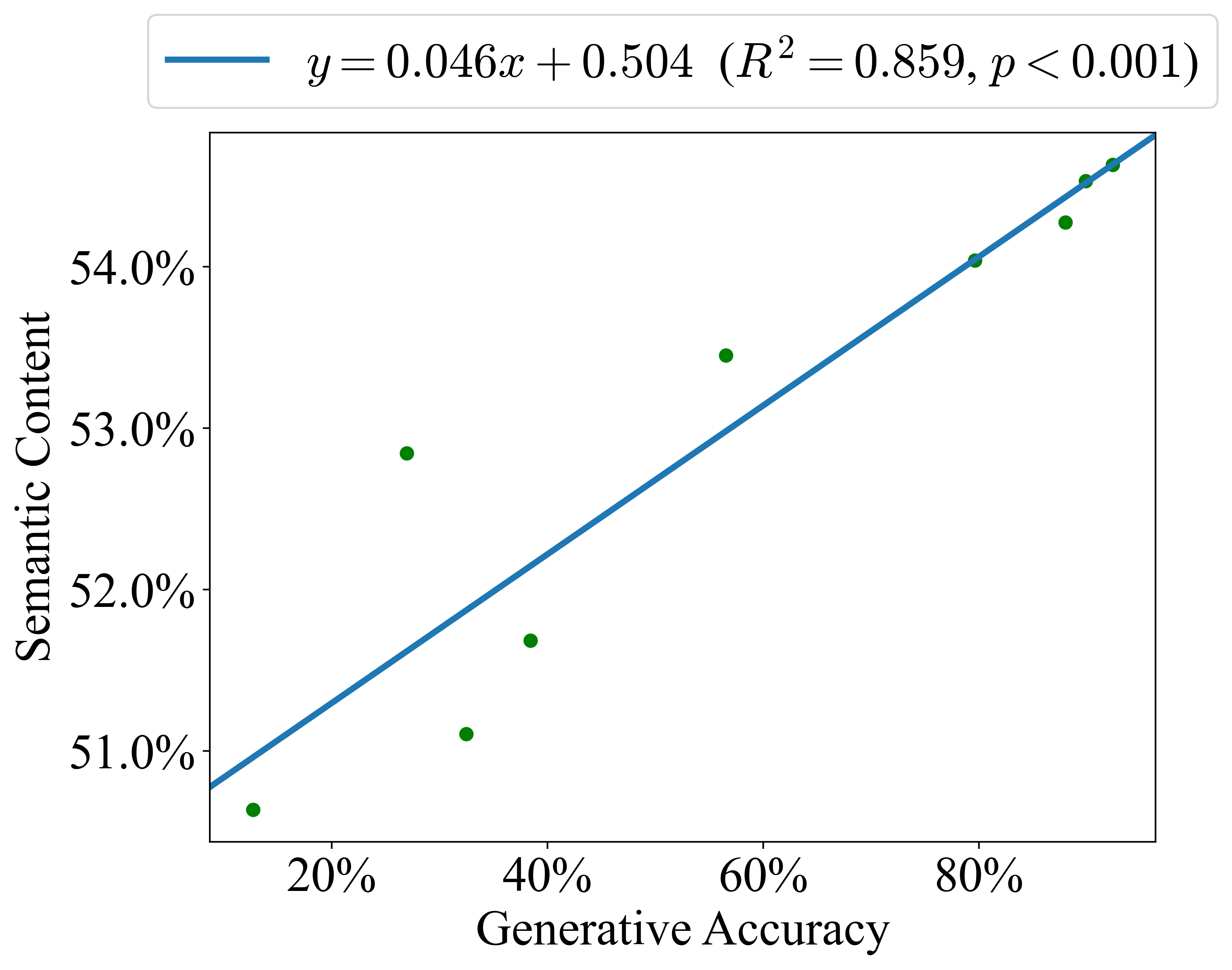}
  \caption{Regressing semantic content vs. generative accuracy.}
  \label{fig:regression_space:reg}
\end{subfigure}
\hspace{1em}
\begin{subfigure}{.48\linewidth}
  \centering
  \includegraphics[scale=0.3]{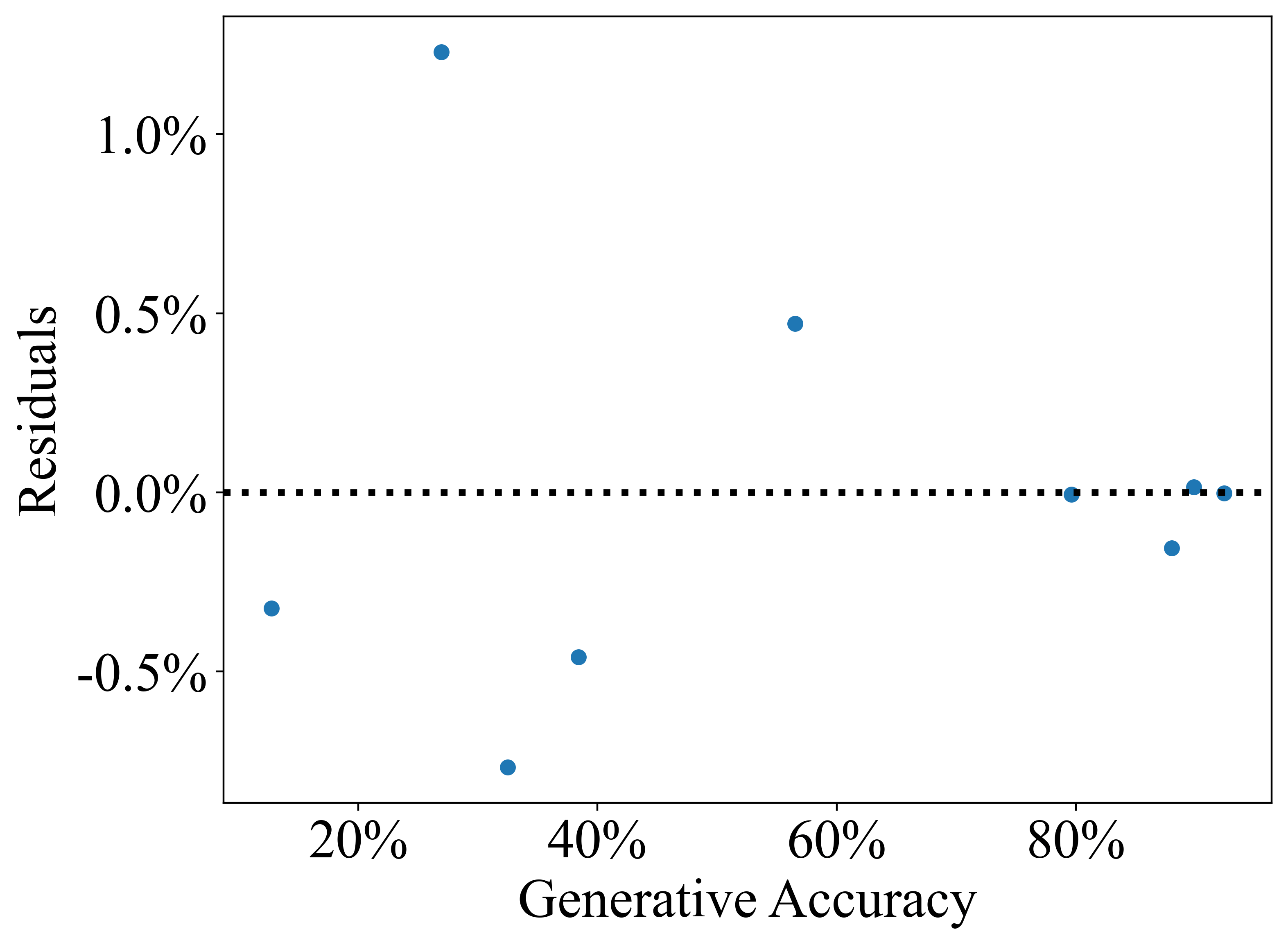}
  \caption{Residuals for semantic content vs. generative accuracy.}
  \label{fig:regression_space:res}
\end{subfigure} \\ \vspace{1em}

\begin{subfigure}{.48\linewidth}
  \centering
  \includegraphics[scale=0.3]{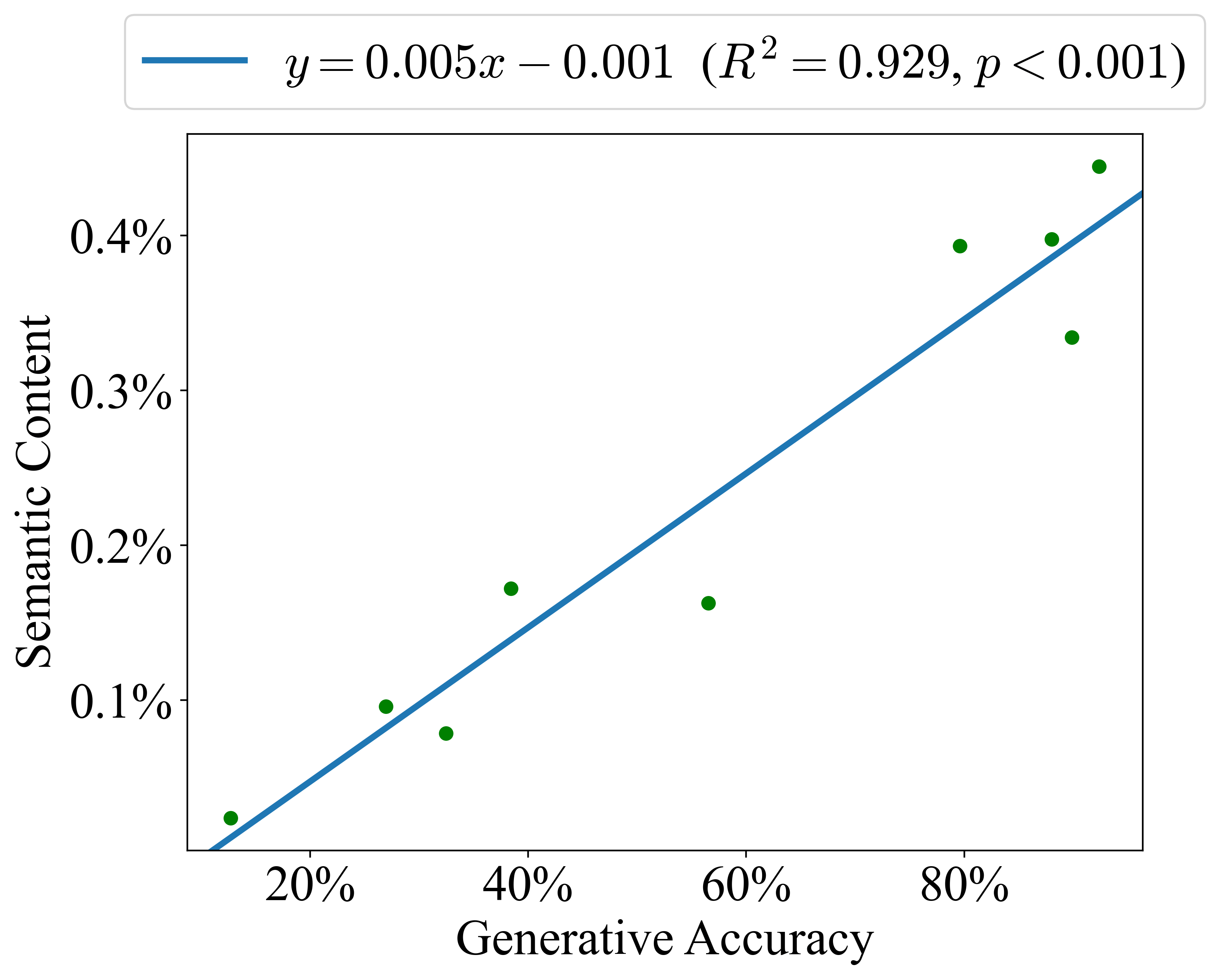}
  \caption{Regressing excess of original over flip semantic content vs. generative accuracy.}
  \label{fig:regression_space:flip:reg}
\end{subfigure}
\hspace{1em}
\begin{subfigure}{.48\linewidth}
  \centering
  \includegraphics[scale=0.3]{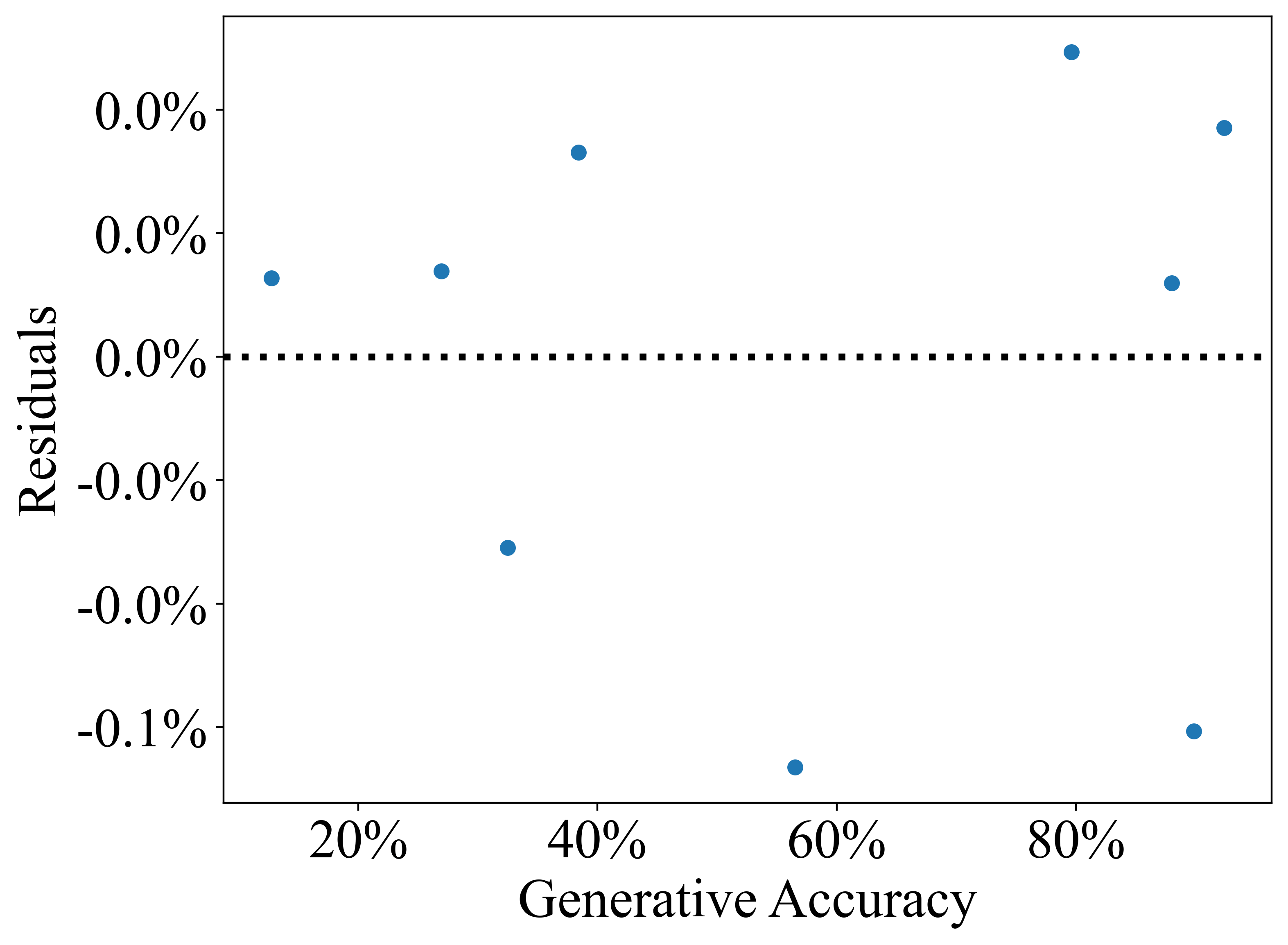}
  \caption{Residuals for excess of original over flip semantic content vs. generative accuracy.}
  \label{fig:regression_space:flip:res}
\end{subfigure} \\ \vspace{1em}

\begin{subfigure}{.48\linewidth}
  \centering
  \includegraphics[scale=0.3]{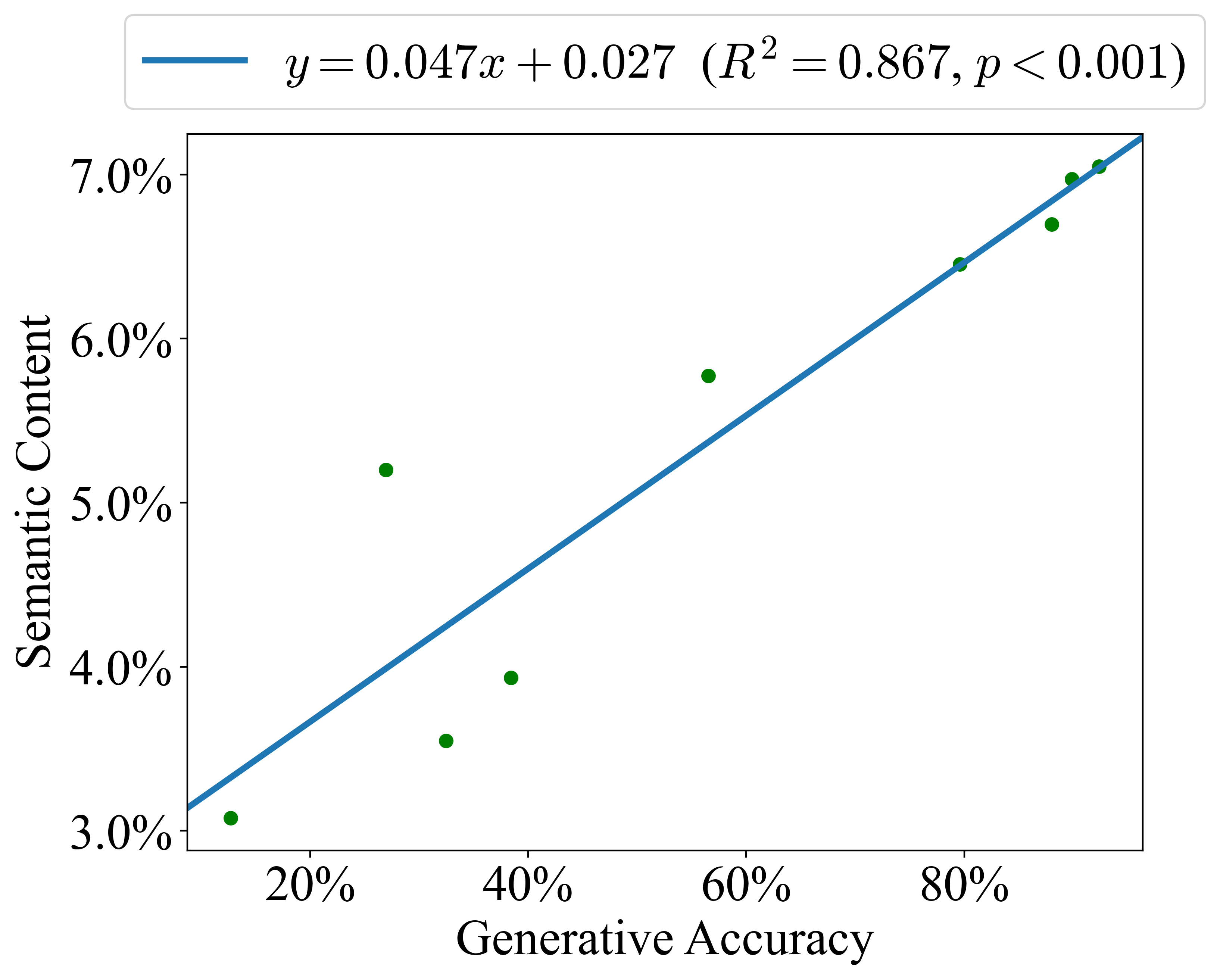}
  \caption{Regressing excess of original over adversarial semantic content vs. generative accuracy.}
  \label{fig:regression_space:adv:reg}
\end{subfigure}
\hspace{1em}
\begin{subfigure}{.48\linewidth}
  \centering
  \includegraphics[scale=0.3]{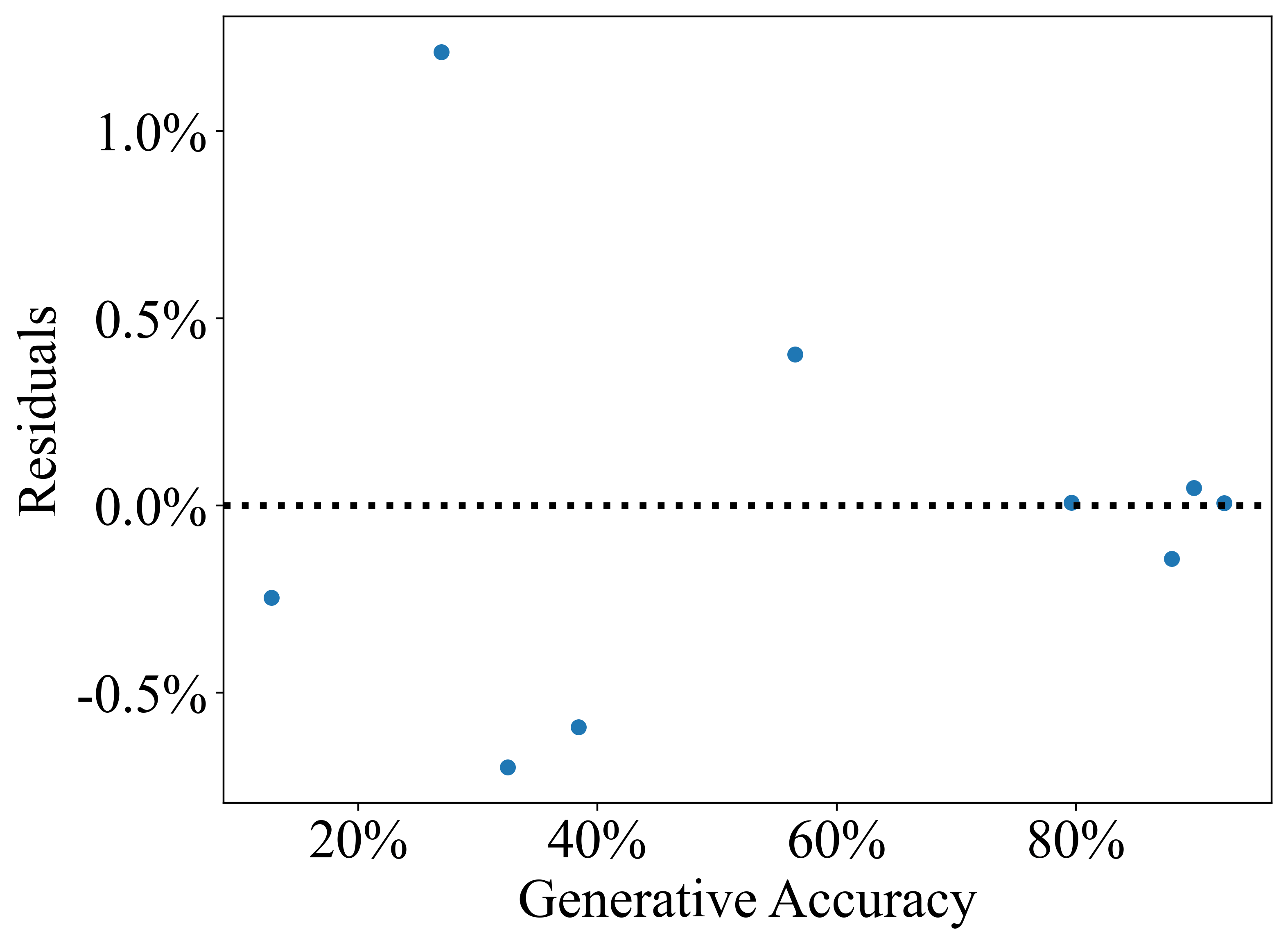}
  \caption{Residuals for excess of original over adversarial semantic content vs. generative accuracy.}
  \label{fig:regression_space:adv:res}
\end{subfigure}
\caption{Regression and residual plots for semantic content (measured by a linear classifier on the input-only trace datasets) vs. generative accuracy over the second half of training.}
\label{fig:regression_space}
\end{figure*}

\Cref{fig:regression,fig:regression_space} display selected regression and residual plots from \Cref{table:main,table:obscured}, respectively. Specifically, we provide the regression and residual plots corresponding to the semantic content of the current state as measured by a linear classifier. In all cases (including those  omitted for brevity and not explicitly shown in the figures), the residual plots confirm a linear relationship between the semantic content and generative accuracy.

\end{document}